\documentclass[10pt,twocolumn,letterpaper]{article}

\usepackage[pagenumbers]{cvpr} 
\makeatletter
\@namedef{ver@everyshi.sty}{}
\makeatother
\usepackage{tikz}
\usepackage{graphicx}
\usepackage{amsmath}
\usepackage{amssymb}
\usepackage{booktabs}
\usepackage{subcaption}
\usepackage{xcolor}
\usepackage{multicol}
\usepackage{multirow}
\usepackage{tabularx}
\usepackage{caption}
\usepackage{dcolumn}
\usepackage{tabularx}
\usepackage{gensymb}
\usepackage{hyperref}
\usepackage{pifont}
\usepackage{arydshln}
\usepackage{colortbl}
\usepackage{enumitem}

\usepackage{svg}
\usepackage{graphicx}
\usepackage{amsmath}
\usepackage{amssymb}
\usepackage{booktabs}
\usepackage{multirow}
\usepackage{pdflscape}
\graphicspath{ {images/} }
 \renewcommand{\paragraph}[1]{
    \vspace{0.5mm}
     \noindent\textbf{#1.} 
 }

\usepackage{rotating}
\usepackage{xcolor}
\usepackage{multicol}
\usepackage{multirow}
\usepackage{tabularx}
\usepackage{caption}
\usepackage{dcolumn}
\usepackage{tabularx}
\usepackage{gensymb}

\usepackage{pifont}
\usepackage{arydshln}
\usepackage{colortbl}
\usepackage{enumitem}
\definecolor{darkgreen}{rgb}{0, 0.5, 0}
\definecolor{myblue}{RGB}{47, 114, 193}
\definecolor{brickred}{rgb}{0.8, 0.25, 0.33}
\definecolor{brandeisblue}{rgb}{0.0, 0.44, 1.0}
\definecolor{blueish}{rgb}{0.0, 0.3, .6}
\definecolor{pink}{rgb}{1, 0, 1}

\hypersetup{
    colorlinks=true,
    linkcolor=cyan ,
    urlcolor=cyan ,
    linktoc=all
}

\usepackage[capitalize]{cleveref}
\crefname{section}{Sec.}{Secs.}
\Crefname{section}{Section}{Sections}
\Crefname{table}{Table}{Tables}
\crefname{table}{Tab.}{Tabs.}

\makeatletter
\@namedef{ver@everyshi.sty}{}
\makeatother

\begin{document}

\title{V2V4Real: A Real-world Large-scale Dataset for Vehicle-to-Vehicle Cooperative Perception}

\author{
Runsheng Xu$^{1}$, Xin Xia$^{1}$\thanks{Corresponding Author, email address: x35xia@ucla.edu}
, Jinlong Li$^{2}$, Hanzhao Li$^{1}$, Shuo Zhang$^{1}$, Zhengzhong Tu$^{3}$, Zonglin Meng$^{1}$, \\ Hao Xiang$^{1}$, Xiaoyu Dong$^{4}$, Rui Song$^{5,6}$, Hongkai Yu$^{2}$, Bolei Zhou$^{1}$,  Jiaqi Ma$^{1}$ \vspace{2mm}
\\ University of California, Los Angeles$^{1}$   \hspace{2mm} Cleveland State University$^{2}$ \hspace{2mm}  University of Texas at Austin$^{3}$ 
\\ Northwestern University$^{4}$  \hspace{2mm} Technical University of Munich$^{5}$  \hspace{2mm} Fraunhofer Institute$^{6}$
}
\maketitle

\begin{abstract}
Modern perception systems of autonomous vehicles are known to be sensitive to occlusions and lack the capability of long perceiving range. It has been one of the key bottlenecks that prevents Level 5 autonomy. Recent research has demonstrated that the Vehicle-to-Vehicle (V2V) cooperative perception system has great potential to revolutionize the autonomous driving industry. However, the lack of a real-world dataset hinders the progress of this field. To facilitate the development of cooperative perception, we present V2V4Real, the first large-scale real-world multi-modal dataset for V2V perception. The data is collected by two vehicles equipped with multi-modal sensors driving together through diverse scenarios.  Our V2V4Real dataset covers a driving area of 410 $km$, comprising 20K LiDAR frames, 40K RGB frames, 240K annotated 3D bounding boxes for 5 classes, and HDMaps that cover all the driving routes. V2V4Real introduces three perception tasks, including cooperative 3D object detection, cooperative 3D object tracking, and Sim2Real domain adaptation for cooperative perception. We provide comprehensive benchmarks of recent cooperative perception algorithms on three tasks. The V2V4Real dataset and codebase can be found at \href{https://research.seas.ucla.edu/mobility-lab/v2v4real}{research.seas.ucla.edu/mobility-lab/v2v4real}.

\end{abstract}

\section{Introduction}
\label{sec:intro}

Perception is critical in autonomous driving (AV) for accurate navigation and safe planning.
The recent development of deep learning brings significant breakthroughs in various perception tasks such as 3D object detection~\cite{wang2022detr3d,rukhovich2022imvoxelnet,li2023voxformer}, object tracking~\cite{weng20203d,zhao2022tracking}, and semantic segmentation~\cite{zhou2022cross,xu2022cobevt}. However, single-vehicle vision systems still suffer from many real-world challenges, such as occlusions and short-range perceiving capability~\cite {wang2020v2vnet,xu2022v2xvit,han2023collaborative}, which can cause catastrophic accidents.
The shortcomings stem mainly from the limited field-of-view of the individual vehicle, leading to an incomplete understanding of the surrounding traffic. 

\begin{figure}[!t]
\centering
\subfloat[Aggregated LiDAR data]{%
  \includegraphics[width=.99\columnwidth]{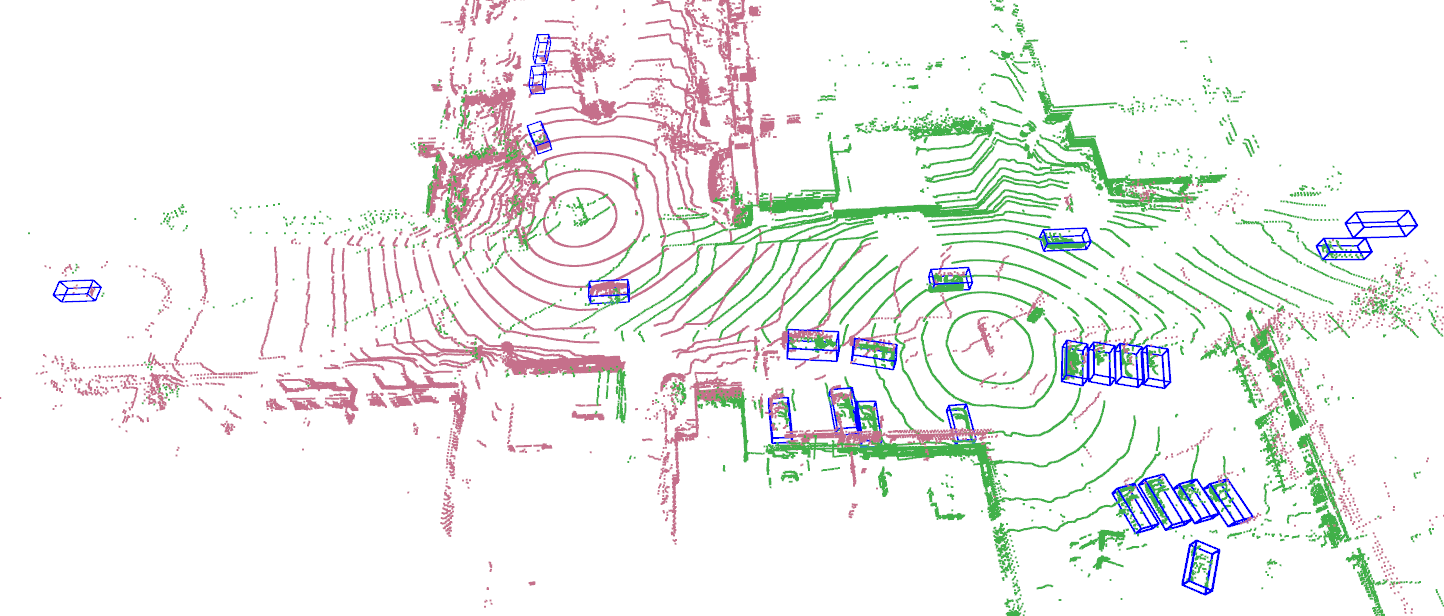}%
}
\hfil
\subfloat[HD map]{%
  \includegraphics[width=.49\columnwidth,height=.26\columnwidth]{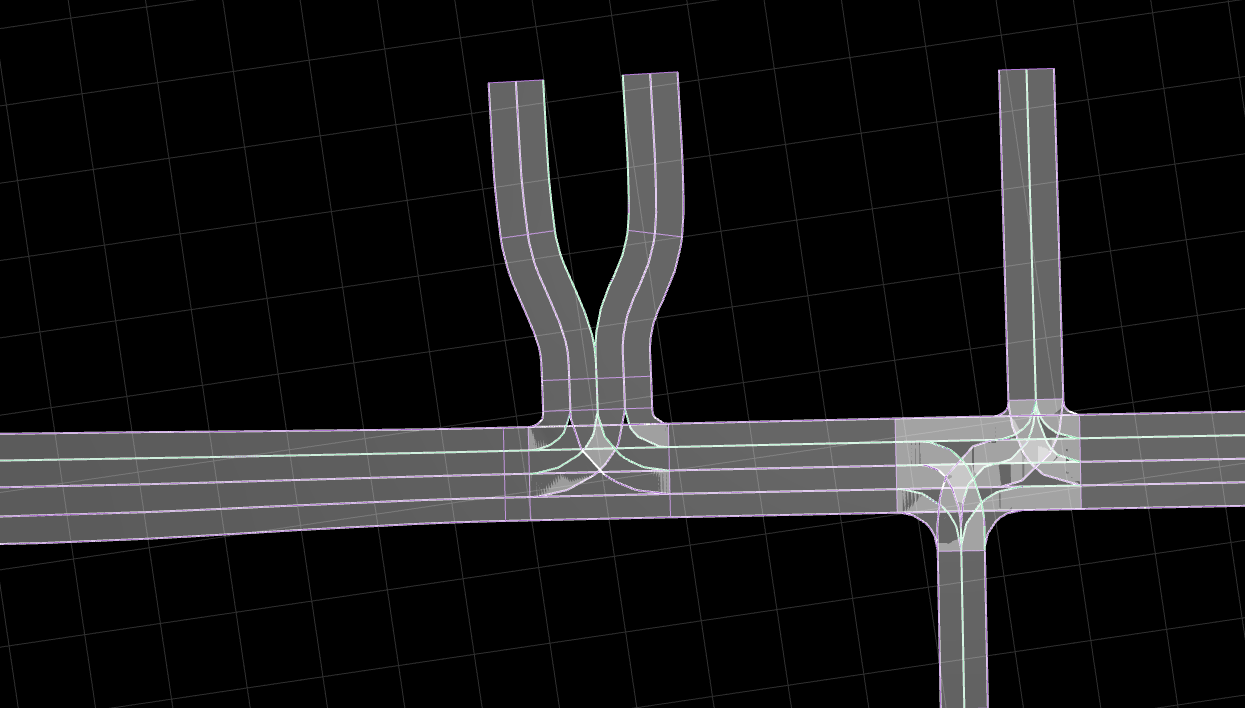}%
}
\hfil
\subfloat[Satallite Map]{%
  \includegraphics[width=.49\columnwidth,height=.26\columnwidth]{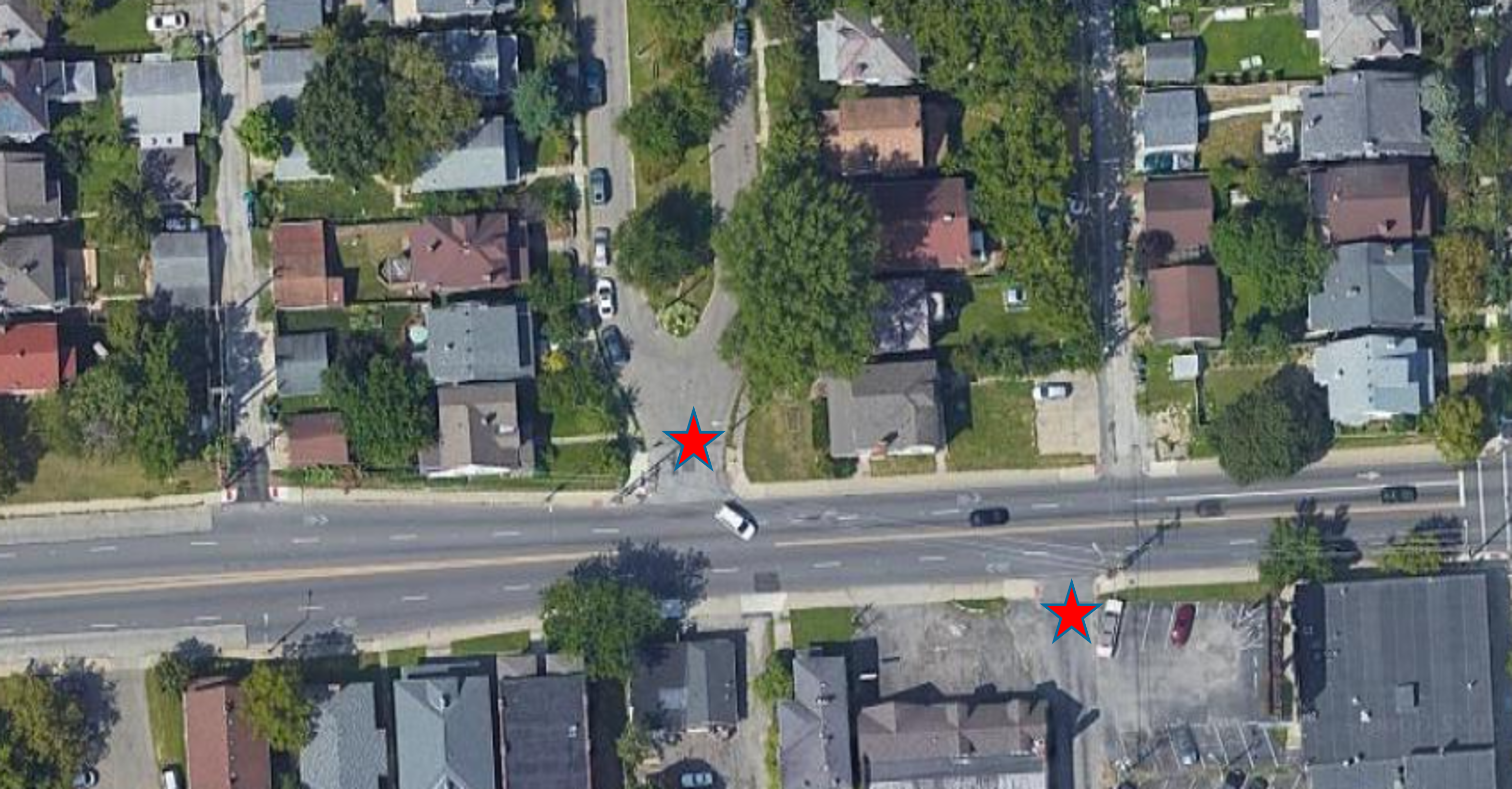}%
}
\caption{\textbf{A data frame sampled from V2V4Real}: (a) aggregated LiDAR data, (b) HD map, and (c) satellite map to indicate the collective position. More qualitative examples of V2V4Real can be found in the supplementary materials.}
\label{fig:data_sample}
\end{figure}

\begin{table*}[!t]
\setlength{\tabcolsep}{5pt}
\centering
\begin{tabular}{l|l|c|c|c|c|c|c|c|c|c}
\hline
Dataset &
  Year &
  \multicolumn{1}{c|}{\begin{tabular}[c]{@{}c@{}}Real/\\ Sim\end{tabular}} &
  V2X &
  \multicolumn{1}{c|}{\begin{tabular}[c]{@{}c@{}} Size\\ (km)\end{tabular}} &
  \begin{tabular}[c]{@{}l@{}}RGB\\ images\end{tabular} &
  \multicolumn{1}{c|}{LiDAR} &
  Maps &
  \multicolumn{1}{c|}{\begin{tabular}[c]{@{}c@{}}3D\\ boxes\end{tabular}} &
  \multicolumn{1}{c|}{Classes} &
  Locations \\ \hline
Kitti~\cite{geiger2012we} &
  2012 &
  Real &
  No &
  - &
  15k &
  15k &
  No &
  200k &
  8 &
  Karlsruhe \\
nuScenes~\cite{caesar2020nuscenes} &
  2019 &
  Real &
  No &
  33 &
  1.4M &
  400k &
  Yes &
  1.4M &
  23 &
  Boston, SG \\
Argo~\cite{chang2019argoverse} &
  2019 &
  Real &
  No &
  290 &
  107k &
  22k &
  Yes &
  993k &
  15 &
  \multicolumn{1}{c}{2x USA} \\
Waymo Open~\cite{sun2020scalability} &
  2019 &
  Real &
  No &
  - &
  1M &
  200k &
  Yes &
  12M &
  4 &
  3x USA \\
OPV2V~\cite{xu2022opv2v} &
  2022 &
  Sim &
  V2V &
  - &
  44k &
  11k &
  Yes &
  230k &
  1 &
  \begin{tabular}[c]{@{}c@{}}CARLA \end{tabular} \\
V2X-Sim~\cite{li2022v2x} &
  2022 &
  Sim &
  \begin{tabular}[c]{@{}l@{}}V2V\&I\ \end{tabular} &
  - &
  60K &
  10k &
  Yes &
  26.6k &
  1 &
  \begin{tabular}[c]{@{}c@{}}CARLA\end{tabular} \\
V2XSet~\cite{xu2022v2xvit} &
  2022 &
  Sim &
  \begin{tabular}[c]{@{}l@{}} V2V\&I\end{tabular} &
  - &
  44K &
  11k &
  Yes &
  230k &
  1 &
  \begin{tabular}[c]{@{}c@{}}CARLA\end{tabular} \\
DAIR-V2X~\cite{yu2022dair} &
  2022 &
  Real &
  V2I &
  20 &
  39K &
  39K &
  No &
  464K &
  10 &
  Beijing, CN \\ \hline
V2V4Real (ours) &
  2022 &
  Real &
  V2V &
  410 &
  40K &
  20K &
  Yes &
  240K &
  5 &
  \begin{tabular}[c]{@{}c@{}} Ohio, USA
  \end{tabular}
  \\\hline
\end{tabular}
\vspace{-2mm}
\caption{\textbf{Comparison of the proposed dataset and existing representative autonomous driving datasets.}}
\label{tab:datasets}
\vspace{-3mm}
\end{table*}

A growing interest and recent advancement in cooperative perception systems have enabled a new paradigm that can potentially overcome the limitation of single-vehicle perception.
By leveraging vehicle-to-vehicle (V2V) technologies, multiple connected and automated vehicles (CAVs) can communicate and share captured sensor information simultaneously.
As shown in a complex intersection in \cref{fig:data_sample}, for example, the ego vehicle (red liDAR) struggles to perceive the upcoming objects located across the way due to occlusions.
Incorporating the LiDAR features from the nearby CAV (green scans) can largely broaden the sensing range of the vehicle and make it even see across the occluded corner.

Despite the great promise, however, it remains challenging to validate V2V perception in real-world scenarios due to the lack of public benchmarks.
Most of the existing V2V datasets, including OPV2V~\cite{xu2022opv2v}, V2X-Sim~\cite{li2022v2x}, and V2XSet~\cite{xu2022v2xvit}, rely on open-source simulators like CARLA~\cite{dosovitskiy2017carla} to generate synthetic road scenes and traffic dynamics with simulated connected vehicles. However, it is well known that there exists a clear domain gap between synthetic data and real-world data, as the traffic behavior and sensor rendering in simulators are often not realistic enough~\cite{suo2021trafficsim,manivasagam2020lidarsim}. Hence, models trained on these benchmarks may not generalize well to realistic driving situations.

To further advance innovative research on V2V cooperative perception, we present a large-scale multimodal and multitask V2V autonomous driving dataset, which covers  410 $km$ road and contains $20K$ LiDAR frames with more than $240K$ 3D bounding box annotations.
Compared to the only existing real-world cooperative dataset DAIR-V2X~\cite{yu2022dair}, our proposed V2V4Real dataset shows several strengths: (1) DAIR-V2X focuses on Vehicle-to-Infrasctrure~(V2I) applications without supporting V2V perception. Compared to V2I, V2V does not require the pre-installed sensors restricted in a certain area, which is more flexible and scalable. Our dataset fills the gap by focusing on the important V2V cooperation. (2) V2V4Real includes four diverse road types, including intersection, highway entrance ramp, highway straight road, and city straight road, covering broader driving areas and greater mileage. (3) We also provide high-definition (HD) maps that can be used for road topology prediction and semantic bird's-eye-view (BEV) map understanding. (4) We construct several benchmarks that can train and evaluate recent autonomous perception algorithms, including 3D object detection, object tracking, and Sim2Real domain adaption, while DAIR-V2X only has a single track. 5) We have provided 8 state-of-the-art cooperative perception algorithms for benchmarking, whereas DAIR-V2X only implements 3 baseline methods. Unlike DAIR-V2X, which can be only accessed within China\footnote{https://thudair.baai.ac.cn/index}, we will make all the data, benchmarks, and models publically available across the globe. 

Our contributions can be summarized as follows:
\begin{itemize}
 \setlength{\parskip}{0pt}
  \setlength{\itemsep}{0pt plus 0pt}
    \item We build the V2V4Real, 
    a large real-world dataset dedicated to V2V cooperative autonomous perception. All the frames are captured by multi-modal sensor readings from real-world diverse scenarios in Columbus, Ohio, in the USA.
    \item We provide more than $240K$ annotated 3D bounding boxes for 5 vehicle classes, as well as corresponding HDMaps along the driving routes, which enables us to train and test cooperative perception models in real-world scenarios. 
    \item We introduce three cooperative perception tasks, including 3D object detection, object tracking, and Sim2Real, providing comprehensive benchmarks with several SOTA models. The results show the effectiveness of V2V cooperation in multiple tasks.
\end{itemize}

\section{Relaed Work}
\label{sec:related-work}

\subsection{Autonomous Driving Datasets.}

Public datasets have contributed to the rapid progress of autonomous driving technologies in recent years. \cref{tab:datasets} summarizes the recent autonomous driving datasets.
The earlier datasets mainly focus on 2D annotations (boxes, masks) for RGB camera images, such as Cityscapes~\cite{cordts2016cityscapes}, Synthia~\cite{ros2016synthia}, BDD100K~\cite{yu2020bdd100k}, to name a few.
However, achieving human-level autonomous driving requires accurate perception and localization in the 3D real world, whereas learning the range or depth information from pure 2D images is an ill-posed problem.

To enable robust perception in 3D or map-view, multimodal datasets that typically involve not only camera images but also range data such as Radar or LiDAR sensors have been developed~\cite{geiger2012we,caesar2020nuscenes,sun2020scalability}.
KITTI~\cite{geiger2012we} was a pioneering dataset that provides multimodal sensor readings, including front-facing stereo camera and LiDAR for 22 sequences, annotated with 200k 3D boxes and tasks of 3D object detection, tracking, stereo, and optical flow.
Subsequently, NuScenes~\cite{caesar2020nuscenes} and Waymo Open dataset~\cite{sun2020scalability} is the most recent multimodal datasets providing an orders-of-magnitude larger number of scenes (over 1K), with 1.4M and 993K annotated 3D boxes, respectively.
Despite remarkable progress, those datasets only aim at developing single-vehicle driving capability, which has been demonstrated to have limited ability to handle severe occlusions as well as long-range perception~\cite{xu2022opv2v,wang2020v2vnet,xu2022model,xu2022bridging}.

The recent development of V2V technologies has made it possible for vehicles to communicate and fuse multimodal features collaboratively, thus yielding a much broader perception range beyond the limit of single-view methods.
OPV2V~\cite{xu2022opv2v} builds the first-of-a-kind 3D cooperative detection dataset using CARLA and OpenCDA co-simulation.
V2XSet~\cite{xu2022v2xvit} and V2X-Sim~\cite{li2022v2x} further explore the viability of vehicle-to-everything (V2X) perception using synthesized data generated from CARLA simulator~\cite{dosovitskiy2017carla}.
Unlike the above-simulated datasets, DAIR-V2X is the first real-world dataset for cooperative detection.
However, DAIR-V2X only concentrates on V2I cooperation, neglecting the important V2V application, which can be more flexible and more likely to be scalable. As V2V and V2I perception has major differences, i.e., V2V perception needs to deal with more diverse traffic scenarios and occlusions~\cite{xu2022v2xvit}, a real-world dataset for V2V perception is needed. Furthermore, DAIR-V2X only spans limited road types~(i.e., only intersections) and constrained driving route length~(only 20km).

\subsection{3D Detection}

3D object detection plays a critical role in the success of autonomous driving.
Based on available sensor modality, 3D detection has roughly three categories.
(1) \textbf{Camera-based} detection denotes approaches that detect 3D objects from a single or multiple RGB images~\cite{reading2021categorical,roddick2018orthographic,rukhovich2022imvoxelnet,huang2021bevdet,wang2022detr3d}. For instance, ImVoxelNet~\cite{rukhovich2022imvoxelnet} builds a 3D volume in 3D world space and samples multi-view features to obtain the voxel representation. DETR3D~\cite{wang2022detr3d} models 3D objects using queries to index into extracted 2D multi-camera features, which directly estimate 3D bounding boxes in 3D spaces. Especially, an additional detection head and  attention modules can further improve small object detection accuracy for RGB images\cite{liu2022yolov5}
(2) \textbf{LiDAR-based} detection typically converts LiDAR points into voxels or pillars, resulting in 3D voxel-based~\cite{zhou2018voxelnet,yan2018second} or 2D pillar-based methods~\cite{lang2019pointpillars,yang2018pixor}. Since 3D voxels are usually expensive to process, PointPillars~\cite{lang2019pointpillars} propose to compress all the voxels along the $z$-axis into a single pillar, then predicting 3D boxes in the bird's-eye-view space. Benefiting from its fast processing and real-time performance, many recent 3D object detection models follow this pillar-based approach~\cite{wang2020pillar,fan2022embracing}.
(3) \textbf{Camera-LiDAR fusion} presents a recent trend in 3D detection that fuses information from both image and LiDAR points. One of the key challenges in multimodal fusion is how to align the image features with point clouds. Some methods~\cite{qi2018frustum, vora2020pointpainting} use a two-step framework, e.g., first detect the object in 2D images, then use the obtained information to further process point clouds; more recent works~\cite{prakash2021multi,li2022deepfusion} develop end-to-end fusion pipelines and leverage cross-attention~\cite{vaswani2017attention} to perform feature alignment.

\subsection{V2V/V2X Cooperative Perception}
\label{sec:2.3}
Due to the intrinsic limitation of camera/LiDAR devices, occlusions and long-distance perception are extremely challenging for single-vehicle systems, which can potentially cause catastrophic consequences in complex traffic environments~\cite {xu2022opv2v}.
Cooperative systems, on the other hand, can unlock the possibility of multi-vehicle detection that tackles the limitation of single-vehicle perception.
Among these, V2V (Vehicle-to-Vehicle) approaches center on collaborations between vehicles, while V2X (Vehicle-to-Everything) involves correspondence between vehicles and infrastructure.
V2V/V2X cooperative perception can be roughly divided into three categories: (1) {Early Fusion}~\cite{chen2019cooper} where raw data is shared among CAVs, and the ego vehicle makes predictions based on the aggregated raw data, (2) {Late Fusion}~\cite{rawashdeh2018collaborative} where detection outputs (\textit{e.g.}, 3D bounding boxes, confidence scores) are shared, then fused to a `consensus' prediction, and (3) {Intermediate Fusion}~\cite{wang2020v2vnet,chen2019f,xu2022opv2v,lu2022robust} where intermediate representations are extracted based on each agent’s observation and then shared with CAVs.

Recent state-of-the-art methods~\cite{wang2020v2vnet,chen2019f,xu2022opv2v} typically choose the intermediate neural features computed from each agent's sensor data as the transmitted features, which achieves the best trade-off between accuracy and bandwidth requirements.
For instance, V2VNet~\cite{wang2020v2vnet} adopted graph neural networks to fuse intermediate features.
F-Cooper~\cite{chen2019f} employed max-pooling fusion to aggregate shared Voxel features. Coopernaut~\cite{cui2022coopernaut} used Point Transformer~\cite{zhao2021point} to deliver point features and conduct experiments under AustoCastSim~\cite{autocast}.
CoBEVT~\cite{xu2022cobevt} proposed local-global sparse attention that captures complex spatial interactions across views and agents to improve the performance of cooperative BEV map segmentation. AttFuse~\cite{xu2022opv2v} proposed an agent-wise self-attention module to fuse the received intermediate features.
V2X-ViT~\cite{xu2022v2xvit} presented a unified vision transformer for multi-agent multi-scale perception and achieves robust performance under GPS error and communication delay. 

\section{V2V4Real Dataset}
To expedite the development of V2V Cooperative Perception for autonomous driving, we propose \textbf{V2V4Real}, the real-world, large-scale, multi-modal dataset with diverse driving scenarios. This dataset is annotated with both 3D bounding boxes and HDMaps for the research of multi-vehicle cooperative perception. In this section, we first detail the setup of data collection (\cref{ssec:data-acq}), and then describe the data annotation approach (\cref{ssec:data-annotation}), and finally analyze the data statistics (\cref{ssec:data_analysis}).

\begin{figure}[!htb]
\centering
\subfloat[Tesla Vehicle]{%
  \includegraphics[width=.49\columnwidth,height=.3\columnwidth]{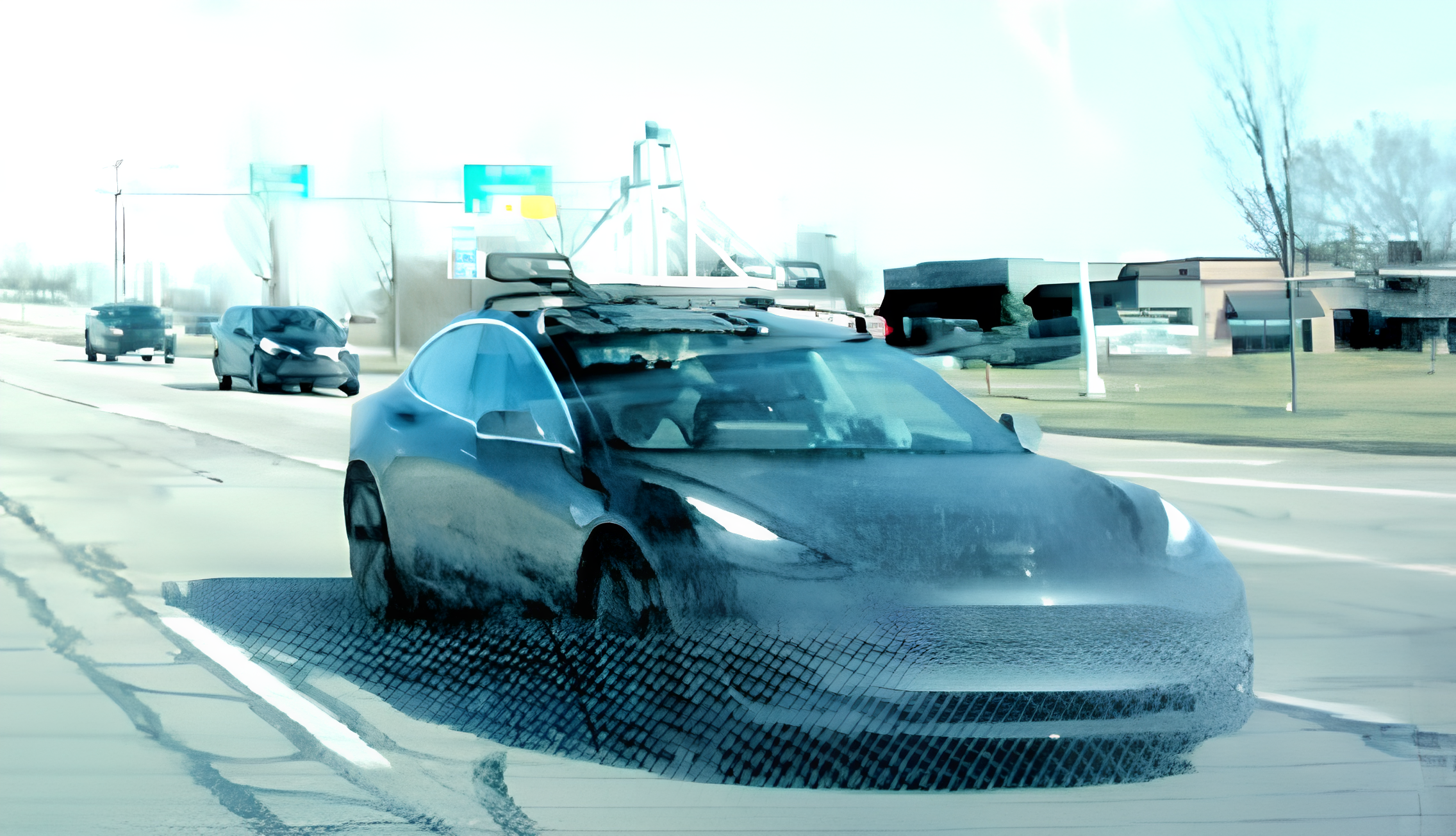}%
}
\hfil
\subfloat[Ford Fusion Vehicle]{%
  \includegraphics[width=.49\columnwidth,height=.3\columnwidth]{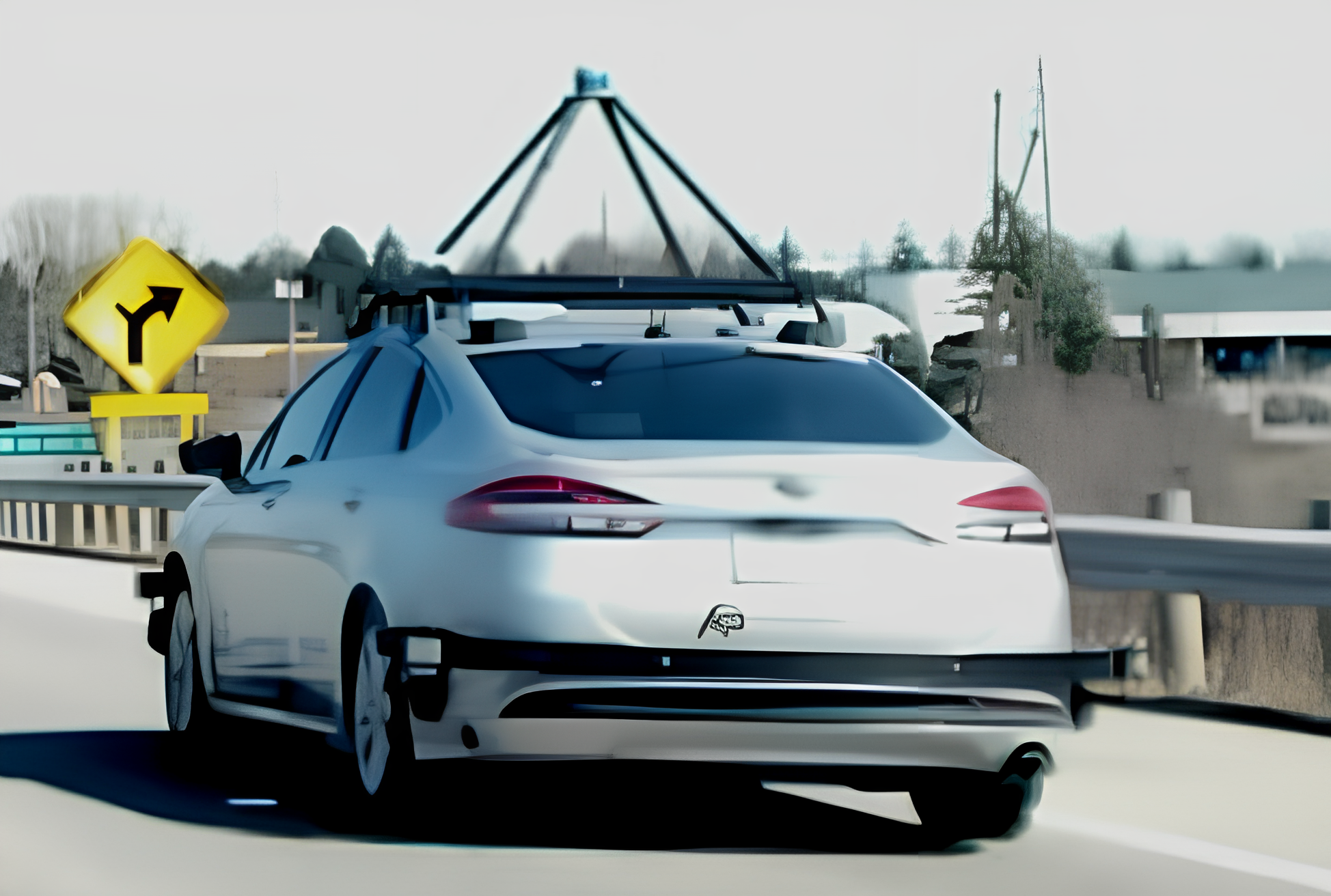}%
}
\hfil
\subfloat[Sensor setup for our data collection platform]{%
  \includegraphics[width=0.7\columnwidth]{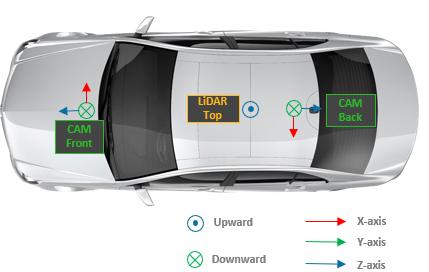}%
}
\caption{\textbf{The information of the collection vehicles}.a) The Tesla vehicle. b) The Ford Fusion vehicle. c) The sensor setup for both vehicles.Note that the photo of Tesla is taken from the rear camera of Ford, and that of Ford is taken from the front camera of Tesla. }
\label{fig:car}

\end{figure}

\subsection{Data Acquisition}
\label{ssec:data-acq}

\noindent\textbf{Sensor Setup.} We collect the V2V4Real via two experimental connected automated vehicles including a Tesla vehicle~(Fig.~\ref{fig:car}a)  and a Ford Fusion vehicle~(Fig.~\ref{fig:car}b) retrofitted by Transportation Research Center(TRC) company and AutonomouStuff~(AStuff)
Company respectively. Both vehicles are equipped with a Velodyne VLP-32 LiDAR sensor, two mono cameras (front and rear), and GPS/IMU integration systems. The sensor layout configuration can be found in Fig.~\ref{fig:car}c, and the detailed parameters are listed in Table.~\ref{table:sensor-details}.

\noindent\textbf{Driving Route.} The two vehicles drive simultaneously in Columbus, Ohio, and their distance is maintained within 150 meters to ensure overlap between their views. To enrich the diversity of sensor-view combinations, we vary the relative poses of the two vehicles across different scenarios~(see ~\cref{ssec:data_analysis} for details). We collect driving logs for three days that cover 347 km of highway road and 63 km of city road. The driving routes are visualized in ~\cref{fig:route}, wherein the red route is on day 1~(freeway with one to five lanes), the yellow route is on day 2~(city road, one to two lanes), and the green route is on day 3~(highway, two to four lanes).

\noindent\textbf{Data Collection.} We collect 19 hours of driving data of 310K frames. We manually select the most representative 67 scenarios, each 10-20 seconds long. We sample the frames at 10Hz, resulting in a total of 20K frames of LiDAR point cloud and 40K frames of RGB images. For each scene, we ensure that the asynchronizations between two vehicles' sensor systems are less than $50~ms$. All the scenarios are aligned with maps containing drivable regions, road boundaries, as well as dash lines.

\subsection{Data Annotation}
\label{ssec:data-annotation}

\begin{figure}[!htb]
    \centering
    \includegraphics[width=3.0in]{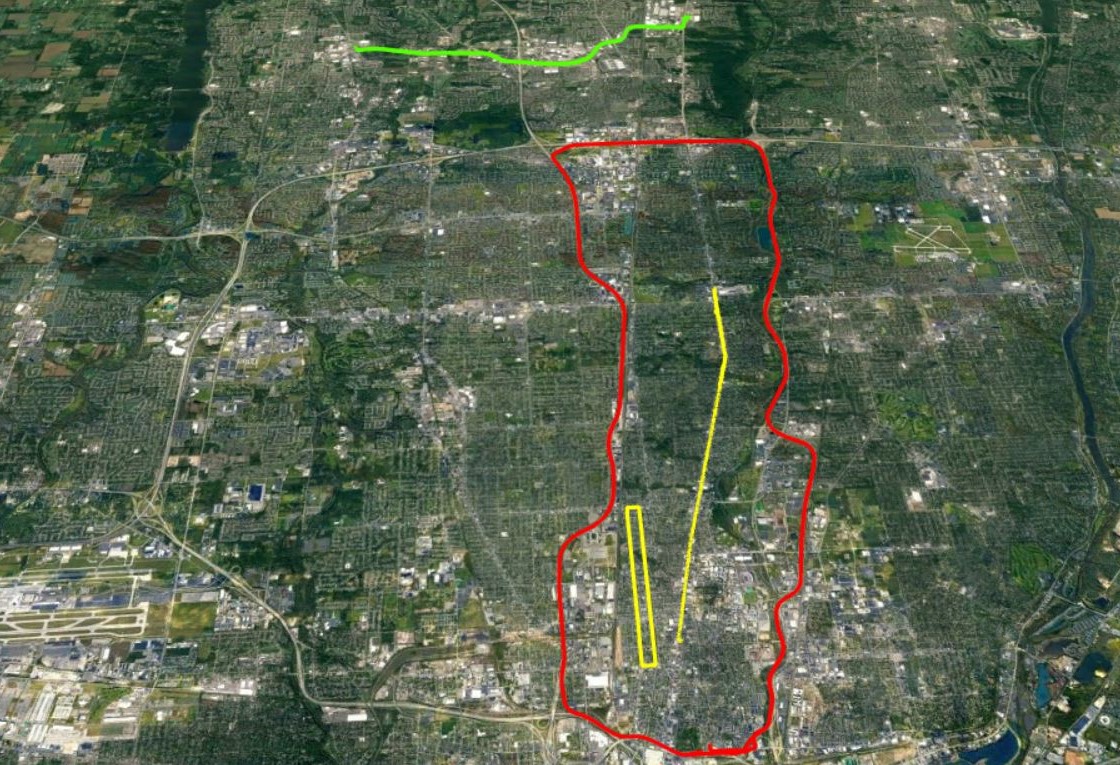}
    \caption{Driving routes of our two collection vehicles. Different colors represent the routes collected on different days. }
    \label{fig:route}
\end{figure}

\noindent\textbf{Coordinate System.} Our dataset includes four different coordinate systems: the LiDAR coordinate system for Tesla and Ford Fusion, the HDmap coordinate, and the earth-earth, fixed-coordinate(ECEF). We annotate the 3D bounding boxes separately based on each vehicle's LiDAR coordinate system such that each vehicle's sensor data alone can also be treated as single-agent detection tasks.  We utilize the positional information provided by GPS on the two vehicles to initialize the relative pose of the two vehicles for each frame. The origin of the HDMap aligns with the initial frame of Tesla for each driving route. 

\noindent\textbf{3D Bounding boxes annotation.} We employ SusTechPoint~\cite{li2020sustech}, a powerful opensource labeling tool, to annotate 3D bounding boxes for the collected LiDAR data. We hire two groups of professional annotators. One group is responsible for the initial labeling, and the other further refines the annotations. There are five object classes in total, including cars, vans, pickup trucks, semi-truck, and buses. For each object, we annotate its 7-degree-of-freedom 3D bounding box containing $x, y, z$ for the centroid position and $l, w, h, yaw$ for the bounding box extent and yaw angles. We also record each object's driving state (\ie dynamic or parking). To facilitate  downstream applications such as tracking and behavior prediction,
we assign consistent id and size for the same object in different timestamps. 

\begin{table}[!tb]
\centering
\small
\setlength{\tabcolsep}{4pt}
\begin{tabular}{l|l}
\hline
Sensors    & Details                                \\ \hline
2x Camera  & RGB, Tesla:$1280 \times 720$, Ford:$2064 \times 1544$    \\ \hline
1x LiDAR & \begin{tabular}[c]{@{}l@{}}$32$~channels, $1.2~M$ points per second, \\ $200~m$ capturing range, $-25\degree$ to $15\degree$ \\ vertical FOV, $\pm3~cm$ error, 10Hz\end{tabular} \\ \hline
GPS \& IMU & Tesla: RT3000, Ford: Novatel SPAN E1  \\ \hline
\end{tabular}
\vspace{-2mm}
\caption{Sensor specifications for each vehicle.}
\label{table:sensor-details}
\vspace{-3mm}
\end{table}

Since the bounding boxes are annotated separately for the two collection vehicles, an object in the Tesla's frame could have the same id as a different object in Ford Fusion's frame. To avoid such issues, all the object ids in Tesla are labeled between $0-1000$, while ids in Ford Fusion range from $1001-2000$. Moreover, identical objects could have different ids in the annotation files of the two collection vehicles. To solve this issue, we transform the objects from different coordinates to a unified coordinate system and calculate the BEV IoU between all objects. For the objects that have IoU larger than a certain threshold, we assign them the same object id and unify their bounding box sizes.

\noindent\textbf{Map Annotation.} The HD map generation pipeline refers to generating a global point cloud map and vector map. To generate the point cloud map, we fuse a sequence of point cloud frames together. More specifically, we first pre-process each LiDAR frame by removing the dynamic objects while keeping the static elements. Then, a Normal Transformation Distribution scan matching algorithm is applied to compute the relative transformation between two consecutive LiDAR frames. The LiDAR odometry can then be constructed by taking the transformation. However, the noise imbued in the LiDAR data can lead to accumulated errors in the estimated transformation matrix as the frame index increases. Therefore, we compensate for these errors by further integrating the translation and heading information provided by the on-vehicle GPS/IMU system and applying Kalman filter~\cite{chui2017kalman}. Finally, all the points in different frames are transformed onto the map coordinate to form a global point cloud map. The aggregated point cloud maps will be imported to RoadRunner~\cite{crescenzi2001roadrunner} to produce the vector maps. The road is drawn and inferred from the intensity information visualized by distinct colors in Roadrunner. We then output the OpenDRIVE~(Xodr) maps and convert them to lanelet maps~\cite{bender2014lanelets} as the final format.

\begin{figure}[!t]
\centering
 \includegraphics[width=0.9\columnwidth]{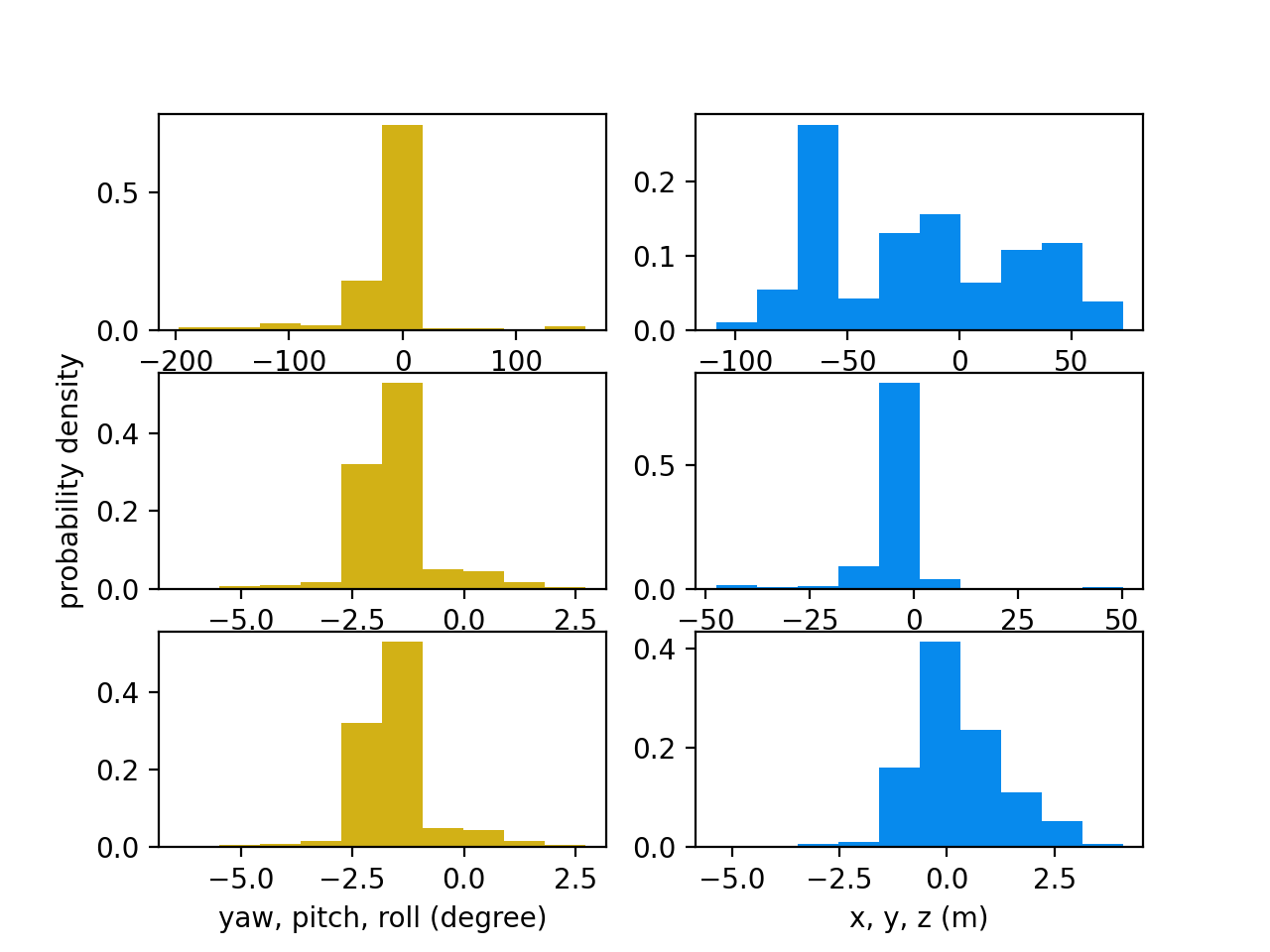}%
\caption{The distribution of the relative poses between the two collection vehicles.}
\label{fig:pos}
\end{figure}

\begin{figure}[!t]
\centering
 \includegraphics[width=0.8\columnwidth]{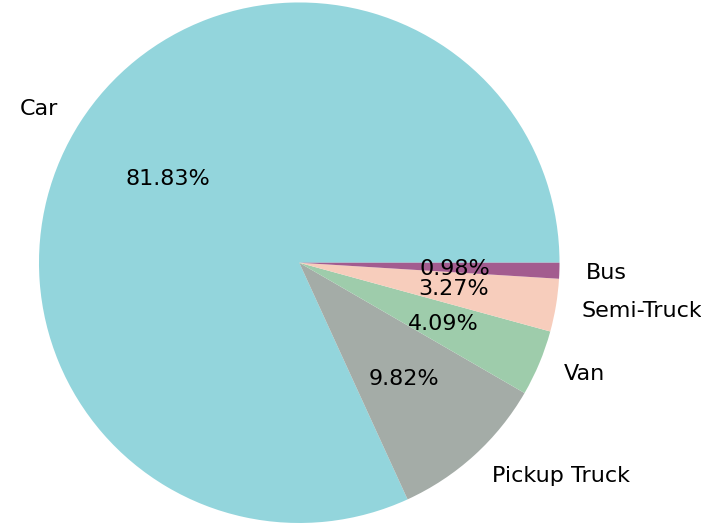}%
\caption{The distribution of vehicle types in collected dataset.}
\label{fig:pie}
\end{figure}

\subsection{Data Analysis}
\label{ssec:data_analysis}
\begin{figure}[!t]
\centering
 \includegraphics[width=.9\columnwidth]{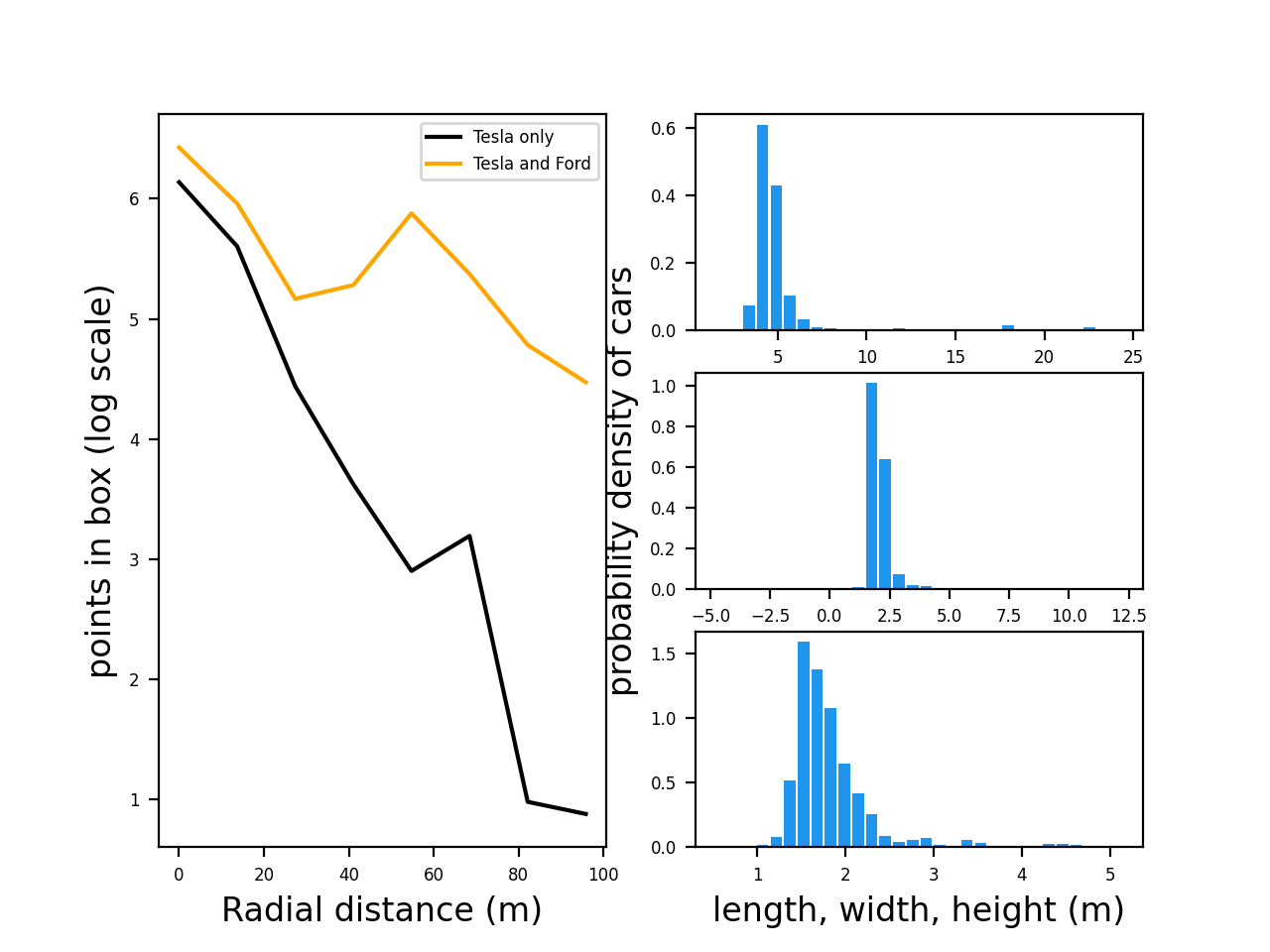}%
\caption{\textbf{Left}: Number of LiDAR points in $e$-based log scale within the ground truth bounding boxes with respect to radial distance from the ego vehicle. \textbf{Right}: Bounding box size distributions.}
\label{fig:data_analysis1}
\end{figure}

\cref{fig:pos} reveals the distribution of relative poses between the two collection vehicles across all scenarios. It can be observed that the two vehicles have a variety of relative poses, generating diverse view combinations of scenes. As \cref{fig:pie} describes, most of the objects in V2V4Real belong to the Car class, while Pickup Truck ranks second. The number of Vans and Semi-Trucks are similar, while Bus has the least quantities. \cref{fig:data_analysis1} shows the LiDAR points density distribution inside different objects bounding boxes and the bounding boxes' size distribution. As we may see in the left figure, when there is only one vehicle~(Tesla) scanning the environment, the number of LiDAR points 
within bounding boxes drops dramatically as the radial distance increases. Enhanced by the shared visual information from the other vehicle~(Ford Fusion), the LiDAR point density of each object increases significantly and still retains at a high level even when the distance reaches $100$~m. This validates the great benefits that cooperative perception can bring to the system. As the right figure reveals, the annotated objects have diverse bounding box sizes, with lengths ranging from $2.5$~m to $23$~m, widths ranging from  $1.5$~m to $4.5$~m, and heights ranging from $1$~m to $4.5$~m, demonstrating the diversity of our data.

\section{Tasks}
Our dataset supports multiple cooperative perception tasks, including detection, tracking, prediction, localization, etc. In this paper, we focus on cooperative detection, tracking, and Sim2Real transfer learning tasks.

\subsection{Cooperative 3D Object Detection}
\noindent\textbf{Scope.} The V2V4Real detection task requires users to leverage multiple LiDAR views from different vehicles to perform 3D object detection on the ego vehicle. Compared to the single-vehicle detection task, cooperative detection has several domain-specific challenges:
\begin{itemize}
     \setlength{\parskip}{0pt}
  \setlength{\itemsep}{0pt plus 0pt}
    \item \textbf{GPS error:} There exists unavoidable error in the relative pose of the collaborators~\cite{liu2021automated}, which can produce global misalignments when transforming the data into a unified coordinate system. 
    \item \textbf{Asynchronicity:} The sensor measurements of collaborators are usually not well-synchronized, which is caused by the asynchrony of the distinct sensor systems as well as the communication delay during the data transmission process~\cite{xu2022v2xvit}.
    \item \textbf{Bandwidth limitation:} Typical V2V communication technologies require restricted bandwidth, which limits the transmitted data size~\cite{rawashdeh2018collaborative,xu2022v2xvit, wang2020v2vnet}. Therefore, cooperative detection algorithms must consider the trade-off between accuracy and bandwidth requirements.
\end{itemize}
The major mission of this track is to design efficient cooperative detection methods to handle the above challenges.

\noindent\textbf{Groundtruth.} During training or testing, one of the two collection vehicles will be selected as the ego vehicle, and the other will transform its annotated bounding boxes to the ego's coordinate. In this way, the groundtruth is defined in a unified (the ego) coordinate system. Note that in the training phase, the ego vehicle is randomly picked, while during testing, we fix Tesla as ego. Due to asynchronicity and localization errors, the bounding boxes from two vehicles corresponding to the same object have some offsets. In such a case, we select the one annotated in the ego vehicle as the groundtruth.

\noindent\textbf{Evaluation.}The evaluation range in $x$ and $y$ direction are $[-100, 100]$~m and $[-40, 40]$~m with respect to the ego vehicle. Similar to DAIR-V2X~\cite{yu2022dair}, we categorize different vehicle types as the same class and focus only on vehicle detection. We 
use the Average Precision~(AP) at Intersection-over-Union~(IoU) 0.5 and 0.7 as the metric to evaluate the performance of vehicle detection. To assess the transmission cost, Average MegaByte~(AM) is employed, which represents the transmitted data size specified by the algorithm. Following \cite{yu2022dair, xu2022v2xvit}, we evaluate all the models under two settings:
1) \textit{Sync} setting, under which the data transmission is regarded as instantaneous, whereas the asynchrony is only induced by the distinct cycles of the sensor systems. 2) \textit{Async} setting, where we consider the data transmission delay as $100$~ms. We simulate such communication delay by retrieving the LiDAR data from the previous timestamp from the non-ego vehicle. 

\begin{figure}[!t]
\centering
\includegraphics[width=0.8\columnwidth]{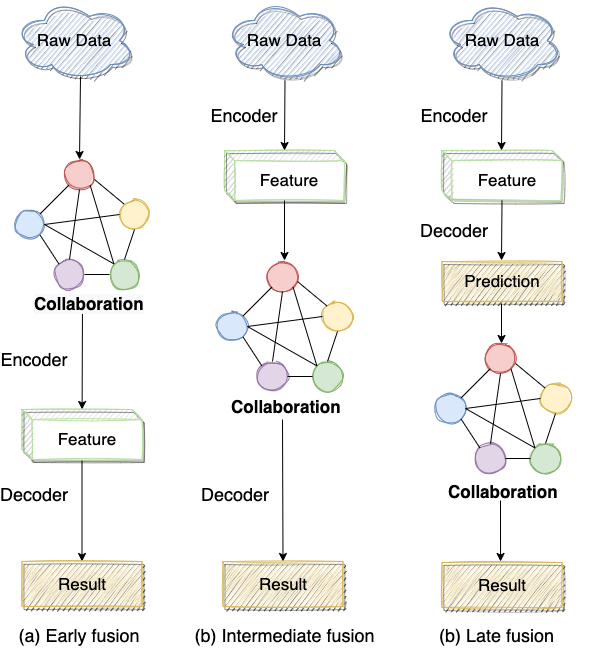}%

\caption{\textbf{The three different fusion strategies:} (a) Early Fusion, (b) Intermediate Fusion, and (c) Late Fusion.}
\label{fig:fusion}
\end{figure}

\begin{table*}[!t]
\setlength{\tabcolsep}{5.5pt}

\centering
\begin{tabular}{l|cccc|cccc|c}
\cline{2-5}
\hline

\multirow{2}{*}{\textbf{Method}} & \multicolumn{4}{c|}{\textbf{Sync} (AP@IoU=0.5/0.7)} & \multicolumn{4}{c|}{\textbf{Async} (AP@IoU=0.5/0.7)} &  \multirow{2}{*}{\begin{tabular}[c]{@{}c@{}}\textbf{AM}\\ (MB)\end{tabular}}\\
                        & Overall  & 0-30m & 30-50m & 50-100m & Overall  &0-30m & 30-50m & 50-100m  & \\ \hline
No Fusion               & 39.8/22.0         & 69.2/42.6     &  29.3/14.4      &  4.8/1.6     &   39.8/22.0         & 69.2/42.6     &  29.3/14.4      &  4.8/1.6 & 0    \\\arrayrulecolor{lightgray}\hline
Late Fusion             & 55.0/26.7     &  73.5/36.8     &   43.7/22.2     &   36.2/17.3     &  50.2/22.4       &     70.7/34.2 &    41.0/19.8     &    26.1/7.8 & 0.003     \\\arrayrulecolor{lightgray}\hline
Early Fusion            & 59.7/32.1     &  76.1/46.3    &    42.5/20.8     &   47.6/\textbf{21.1}     & 52.1/25.8        &  74.6/43.6     &  34.5/16.3     &   30.2/\textbf{9.5} & 0.96      \\\arrayrulecolor{lightgray}\hline
F-Cooper~\cite{chen2019f}                & 60.7/31.8     &  80.8/46.9     &    45.6/23.6    &    32.8/13.4      & 53.6/26.7         &   79.0/44.1   &   38.7/19.5     &   18.1/6.0 & 0.20        \\
V2VNet~\cite{wang2020v2vnet}                  & 64.5/34.3     & 80.6/51.4    &  52.6/26.6    &   42.6/14.6     & 56.4/28.5       &  78.6/48.0    &   44.2/21.5    &   25.6/6.9 & 0.20        \\
AttFuse~\cite{xu2022opv2v}                 & 64.7/33.6     &  79.8/44.1     &   \textbf{53.1}/\textbf{29.3}     &    43.6/19.3      &  57.7/27.5        & 78.6/41.4      &     \textbf{45.5}/\textbf{23.8}    &     27.2/9.0  & 0.20    \\
V2X-ViT~\cite{xu2022v2xvit}                 & 64.9/\textbf{36.9}    &  82.0/\textbf{55.3}    &    51.7/26.6      &   43.2/16.2      &  55.9/29.3       &  79.7/\textbf{50.4}    &    43.3/21.1    &    24.9/7.0  & 0.20      \\
CoBEVT~\cite{xu2022cobevt}                  & \textbf{66.5}/36.0  &  \textbf{82.3}/51.1     &   52.1/28.2    &   \textbf{49.1}/19.5    & \textbf{58.6}/\textbf{29.7}        & \textbf{80.3}/48.3
&    44.7/22.8  &   \textbf{30.5}/8.7 & 0.20 
\\\arrayrulecolor{black}\hline
\end{tabular}
\vspace{-2mm}
\caption{\textbf{Cooperative 3D object detection benchmark.}}
\label{tab:3d_detection}
\vspace{-3mm}
\end{table*}

\noindent\textbf{Benchmarking methods.} We evaluate most commonly adopted fusion strategies as \cref{fig:fusion} demonstrated for cooperative perception with state-of-the-art methods in the domain. In total, four fusion strategies are considered:
\begin{itemize}
     \setlength{\parskip}{0pt}
  \setlength{\itemsep}{0pt plus 0pt}
    \item \textit{No Fusion:} Only ego vehicle's point cloud is used for visual reasoning. This strategy serves as the baseline.
    \item \textit{Late Fusion:} Each vehicle detects 3D objects utilizing its own sensor observations and delivers the predictions to others. Then the receiver applies Non-maximum suppression to produce the final outputs.
    \item \textit{Early Fusion:} The vehicles will directly transmit the raw point clouds to other collaborators and the ego vehicle will aggregate all the point clouds to its own coordinate frame, which preserves complete information but requires large bandwidths.
    \item \textit{Intermediate Fusion:} The collaborators will first project their LiDAR to the ego vehicle's coordinate system and then extract intermediate features using a neural feature extractor. Afterward, the encoded features are compressed and broadcasted to the ego vehicle for cooperative feature fusion. We benchmark a number of leading intermediate methods, including AttFuse~\cite{xu2022opv2v}, F-Cooper~\cite{chen2019f}, V2VNet~\cite{wang2020v2vnet}, V2X-Vit~\cite{xu2022v2xvit}, and CoBEVT~\cite{xu2022cobevt}~(see Sec.~\ref{sec:2.3} for detail descriptions). Similar to previous works \cite{xu2022cobevt, xu2022opv2v, xu2022v2xvit}, we train a simple auto-encoder to compress the intermediate features by $32\times$ to save bandwidth and decompress them to the original size on the ego side.
    
\end{itemize}

\subsection{Object Tracking}
\noindent\textbf{Scope.} In this track, we study whether and how object tracking models can obtain benefits from the cooperative system. There are two major approaches to tracking algorithms: joint detection and tracking and tracking by detection. In this paper, we focus on the second class.

\noindent\textbf{Evaluation.} We employ the same evaluation metrics in \cite{weng20203d, caesar2020nuscenes} for object tracking, including 1) Multi Object Tracking Accuracy~(MOTA), 2) Mostly Tracked Trajectories~(MT), 3) Mostly Lost Trajectories~(ML), 4) Average Multiobject Tracking Accuracy~(AMOTA), 5) Average Multiobject Tracking Precision~(AMOTP), and 6) scaled Average Multiobject Tracking Accuracy~(sAMOTA). Specifically, the AMOTA and AMOTP average MOTA and MOTP across all recall thresholds, which takes into account the prediction confidence, compared to traditional MOTA and MOTP metrics. sAMOTA is proposed by~\cite{caesar2020nuscenes} to guarantee a more linear span over the entire $[0,1]$ range significantly difficult tracking tasks.

\begin{table*}[!tb]
\centering
\setlength{\tabcolsep}{10pt}
\begin{tabular}{l|c|c|c|c|c|c}
\hline
Method       & AMOTA($\uparrow$) & AMOTP($\uparrow$) & sAMOTA($\uparrow$) & MOTA($\uparrow$)  & MT($\uparrow$)    & ML($\downarrow$)    \\ \hline
No Fusion    & 16.08    & 41.60  & 53.84     & 43.46     &  29.41     &    60.18   \\\arrayrulecolor{lightgray}\hline
Late Fusion  & 29.28 & 51.08 & 71.05  & 59.89 & 45.25 & 31.22    \\\arrayrulecolor{lightgray}\hline
Early Fusion & 26.19 & 48.15 & 67.34  & 60.87  & 40.95 & 32.13 \\\arrayrulecolor{lightgray}\hline
F-Cooper~\cite{chen2019f}    &   23.29 &  43.11   & 65.63    &  58.34     &    35.75  &  38.91        \\
AttFuse~\cite{xu2022opv2v}      & 28.64 & 50.48 & 73.21 &  63.03 & 46.38 & 28.05 \\
V2VNet~\cite{wang2020v2vnet}       & 30.48 & 54.28 & 75.53  & \textbf{64.85}  & \textbf{48.19} & 27.83 \\
V2X-ViT~\cite{xu2022v2xvit}      &  30.85 & 54.32 & 74.01  & 64.82  & 45.93 & \textbf{26.47} \\
CoBEVT~\cite{xu2022cobevt}       & \textbf{32.12}  & \textbf{55.61} & \textbf{77.65}  & 63.75 & 47.29 & 30.32
\\ \arrayrulecolor{black}\hline
\end{tabular}
\vspace{-2mm}
\caption{\textbf{Cooperative Tracking benchmark}. All numbers represent percentages.}
\label{tab:track}
\vspace{-2mm}
\end{table*}

\begin{table}[htb]
\centering
\setlength{\tabcolsep}{8pt}
\begin{tabular}{l|c|c}
\hline
Method  & \multicolumn{1}{c|}{AP@IoU=0.5} & \multicolumn{1}{|c}{AP drop} \\ \hline
AttFuse~\cite{xu2022opv2v} & 22.5  & 42.2                       \\
AttFuse \textbf{w/ D.A.}  & 23.4 \color{gray}{(+0.9)} & 41.3                           \\\arrayrulecolor{lightgray}\hline
F-Cooper~\cite{chen2019f} & 23.6  & 37.1                         \\
F-Cooper \textbf{w/ D.A.}   & 37.3 \color{gray}{(+13.7)} & 23.4             \\\arrayrulecolor{lightgray}\hline
V2VNet~\cite{wang2020v2vnet}   & 23.2  &  41.3                        \\
V2VNet \textbf{w/ D.A.}   & 26.3 \color{gray}{(+3.1)}  & 38.2          \\\arrayrulecolor{lightgray}\hline
V2X-ViT~\cite{xu2022v2xvit} & 27.4  &  37.5          \\
V2X-ViT \textbf{w/ D.A.}    & 39.5 \color{gray}{(+12.1)}  &  25.4      \\\arrayrulecolor{lightgray}\hline
CoBEVT~\cite{xu2022cobevt} & 32.6 &  33.9          \\
CoBEVT \textbf{w/ D.A.}   & 40.2 \color{gray}{(+7.6)} &  26.3                          \\ \arrayrulecolor{black}\hline
\end{tabular}
\vspace{-2mm}
\caption{\textbf{Domain Adaptation benchmark}. The number in the bracket indicates the precision gain when using domain adaptation. AP drop refers to the precision gap compared to directly training on the V2V4Real dataset.}
\label{tab:da}
\vspace{-3mm}
\end{table}

\noindent\textbf{Baselines tracker.} We implement AB3Dmot tracker~\cite{weng20203d} as our baseline tracker. Given the detection results from the cooperative detection models, AB3Dmot combines the 3D Kalman Filter with Birth and Death Memory technique to achieve an efficient and robust tracking performance.

\subsection{Sim2Real Domain Adaptation}
\noindent\textbf{Scope.} Data labeling is time-consuming and expensive for the perception system~\cite{xiang2022v2xp}. When it comes to cooperative perception, the cost can dramatically expand as the labelers need to annotate multiple sensor views, which is impossible to scale up. A potential solution is to employ infinite and inexpensive simulation data. However, it is known that there is a significant domain gap between simulated and real-world data distributions. Therefore, this track investigates how to utilize domain adaptation methods to reduce domain discrepancy in the cooperative 3D detection task.

\noindent\textbf{Training.} We define the target domain as the V2V4Real dataset and the source domain as a large-scale open simulated OPV2V dataset~\cite{xu2022opv2v}. The training data consists of two parts:
 the OPV2V training set with provided annotations, and V2V4Real training set's LiDAR point cloud without access to the labels. Participants should leverage domain adaption algorithms to enable the cooperative detection models to generate domain-invariant features.

 \noindent\textbf{Evaluation.} The evaluation will be conducted on the test set of V2V4Real dataset under the \textit{Sync} setting, and the assessment protocol is the same as the cooperative 3D object detection track.

 \noindent\textbf{Evaluated methods.} The baseline method is to train the detection models on OPV2V  and directly test on V2V4Real  without any domain adaptation. To demonstrate the effectiveness of domain adaptation, we implement a similar method as in~\cite{chen2018domain}, which applies two domain classifiers for feature-level and object-level adaption and utilizes gradient reverse layer~(GRL)~\cite{ganin2015unsupervised} to backpropagate the gradient to assist the model for generating domain-invariant features.

\section{Experiments}
\subsection{Implementation Details}
The dataset is split into the train/validation/test set with 14,210/2,000/3,986 frames, respectively, for all three tasks. All the detection models employ PointPillar~\cite{lang2019pointpillars} as the backbone to extract 2D features from the point cloud. we train all models with 60 epochs, a batch size of 4 per GPU~(RTX3090), a learning rate of 0.001
, and we decay the learning rate with a cosine annealing~\cite{loshchilov2017decoupled}. Early stopping is used to find the best epoch.  We also add normal point cloud data augmentations for all experiments, including scaling, rotation, and flip~\cite{lang2019pointpillars}.  We employ AdamW~\cite{kingma2014adam} with a weight decay of $1\times 10^{-2}$ to optimize our models. For the tracking task, we take the  previous 3 frames together with the current frame as the inputs.
\subsection{3D LiDAR Object Detection}
\cref{tab:3d_detection} demonstrates the quantitive comparison between various cooperative 3D detection models on our V2V4Real dataset. We can observe that:
\begin{itemize}
     \setlength{\parskip}{0pt}
  \setlength{\itemsep}{0pt plus 0pt}
    \item Compared to the single-vehicle perception baseline, all cooperative perception methods can significantly boost performance by at least 15.2\% in terms of overall AP at IoU 0.5. Furthermore, the accuracy of all evaluation ranges is improved, whereas long-range detection has the most benefits with a minimum of 28.0\% and 11.8\% gain for AP@0.5 and AP@0.7, respectively.
    \item Under both \textit{Sync} and \textit{Async} settings, intermediate fusion methods achieve the best trade-off between accuracy and transmission cost. Among all the intermediate fusion methods, CoBEVT has the best performance in terms of AP@0.5, 1.6\% higher than the second best model V2X-Vit, 6.8\% higher than \textit{Early Fusion}, and 11.5\% higher than \textit{Late Fusion} in the \textit{Sync} setting.
    \item Except for \textit{No Fusion}, all other methods' AP dropped significantly when the communication delay was introduced. For instance, CoBEVT, V2X-ViT, and V2VNet drops 6.3\%, 7.6\%, and 5.8\% at AP@0.7, respectively. This observation highlighted the importance of robustness to the asynchrony for cooperative perception methods.
\end{itemize}
\subsection{3D Object Tracking}
\cref{tab:track} shows the benchmark results for cooperative tracking. It can be seen that  when AB3Dmot combines with cooperative detection, the performance is dramatically better than the single-vehicle tracking method. Similar to the cooperative detection track, CoBEVT~\cite{xu2022cobevt} achieves the best performance in most of the evaluation metrics, including AMOTA~(16.04\% higher than baseline), sAMOTA~(23.81\% higher than baseline), and AMOTP~(14.01\% better than baseline).

\subsection{Sim2Real Domain Adaptation}
As \cref{tab:da} reveals, there exist serious domain gaps between the simulated dataset OPV2V and our real-world dataset V2V4Real. Without any domain adaptation, only seeing the simulated data will decrease the accuracy of the detection models by $42.2\%$, $37.1\%$, $41.3\%$, $37.5\%$, $33.9$ for AttFuse, F-Cooper, V2VNet, V2X-ViT, and CoBEVT. Applying the domain adaption technique alleviates the performance drop by an average of $7.46\%$. Furthermore, the strongest model, CoBEVT, can reach 40.2\% after employing the domain adaptation, which is higher than the \textit{No Fusion} baseline method that uses real-world data for training.

\section{Conclusion}

We present V2V4Real, a large-scale real-world dataset that covers up to 410 km driving areas, contains 20K LiDAR frames, 40K RGB images, and are annotated with 240K bounding boxes as well as HDMaps, to promote V2V cooperative perception research.
We further introduce three V2V perception benchmarks involving 3D object detection, object tracking, and Sim2Real domain adaptation, which opens up the possibility for future task development.
V2V4Real will be made fully available to the public to accelerate the progress of this new field.
We plan to release the benchmarks and baseline models for HDMap learning tasks and camera images in the next version. 

\noindent\textbf{Broader impact.} Although the proposed benchmark covers various driving scenes for V2V perception, there may still exist extremely challenging scenarios that do not appear in our training set. In such cases, the models should be trained more carefully in order not to hinder generalization abilities. Out-of-distribution detection is also an important topic that has not been investigated within the scope of this paper. These issues should be taken care of by future related research for robust and safe autonomous perception.

\section{Acknowledgement}
The project belongs to OpenCDA ecosystem~\cite{10045043} and is funded in part by  the Federal Highway Administration project and California RIMI Program. Special thanks go to Transportation Research Center Inc for their collaboration in experimental data collection and processing.
{\small
\bibliographystyle{ieee_fullname}
\bibliography{egbib}
}

\appendix
\section{Appendix}
\label{sec:appendix}

This supplementary document is organized as follows:

\begin{itemize}
    \item We present additional information and visualization of the V2V4Real dataset in~\cref{sec:sup-datasets}.
    \item Implementation details of the evaluated models are covered in~\cref{sec:imp-details}.
        \item More ablation studies are covered in~\cref{sec:ablation-details}.
    \item More qualitative 3D object detection results are shown in~\cref{sec:sup-detection}.
    \item More qualitative Sim2Real domain adaptation results are shown in~\cref{sec:sup-da}
\end{itemize}

\section{Dataset Visualization}
\label{sec:sup-datasets}
\begin{figure*}[!ht]
\centering
\footnotesize
\def\xwidth{0.88}
\begin{tabular}{c}
\includegraphics[width=0.7\linewidth]{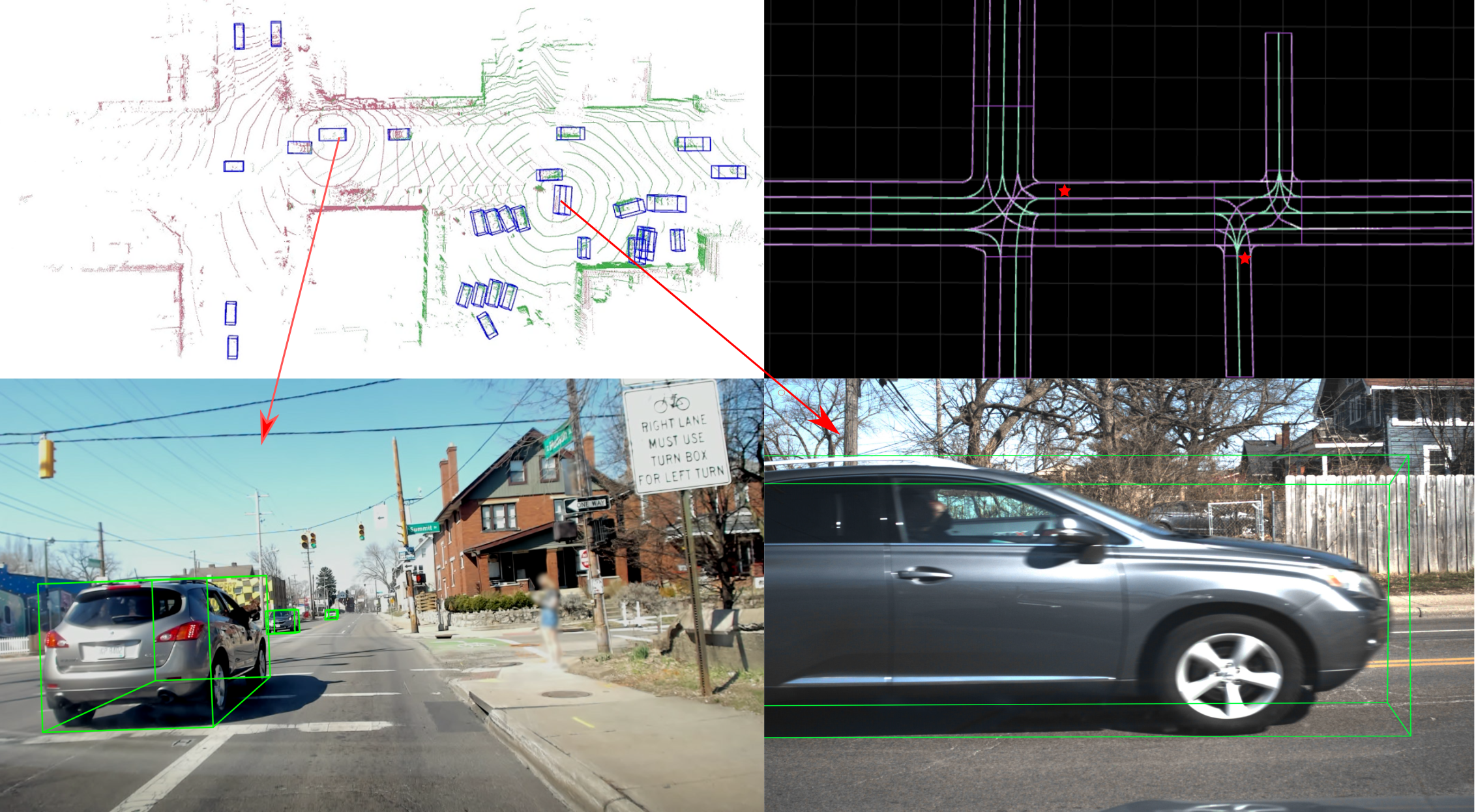} \\
(a) A cityroad example \\
\includegraphics[width=0.7\linewidth]{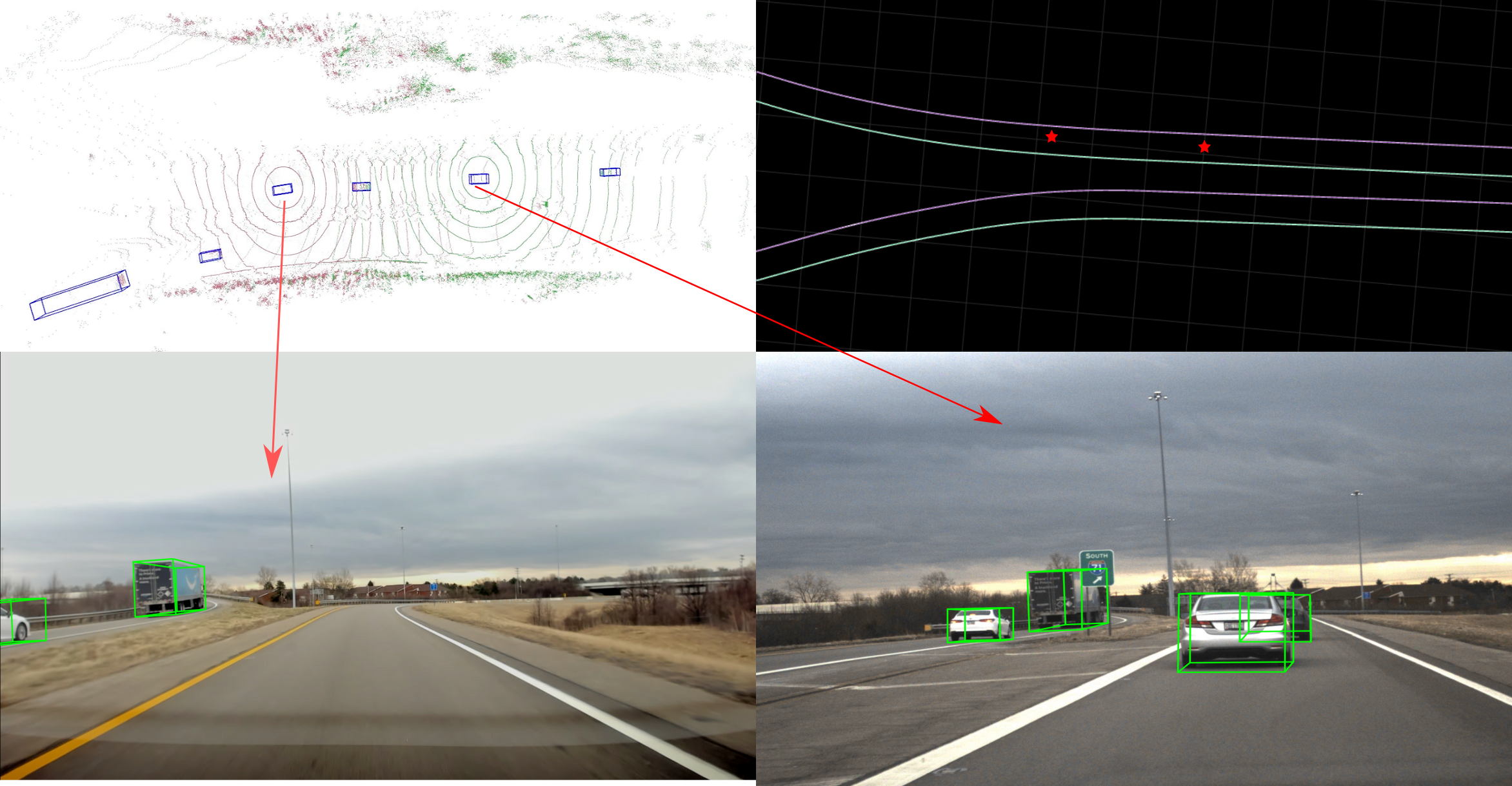} \\
(b)  A highway example \\
\includegraphics[width=0.7\linewidth]{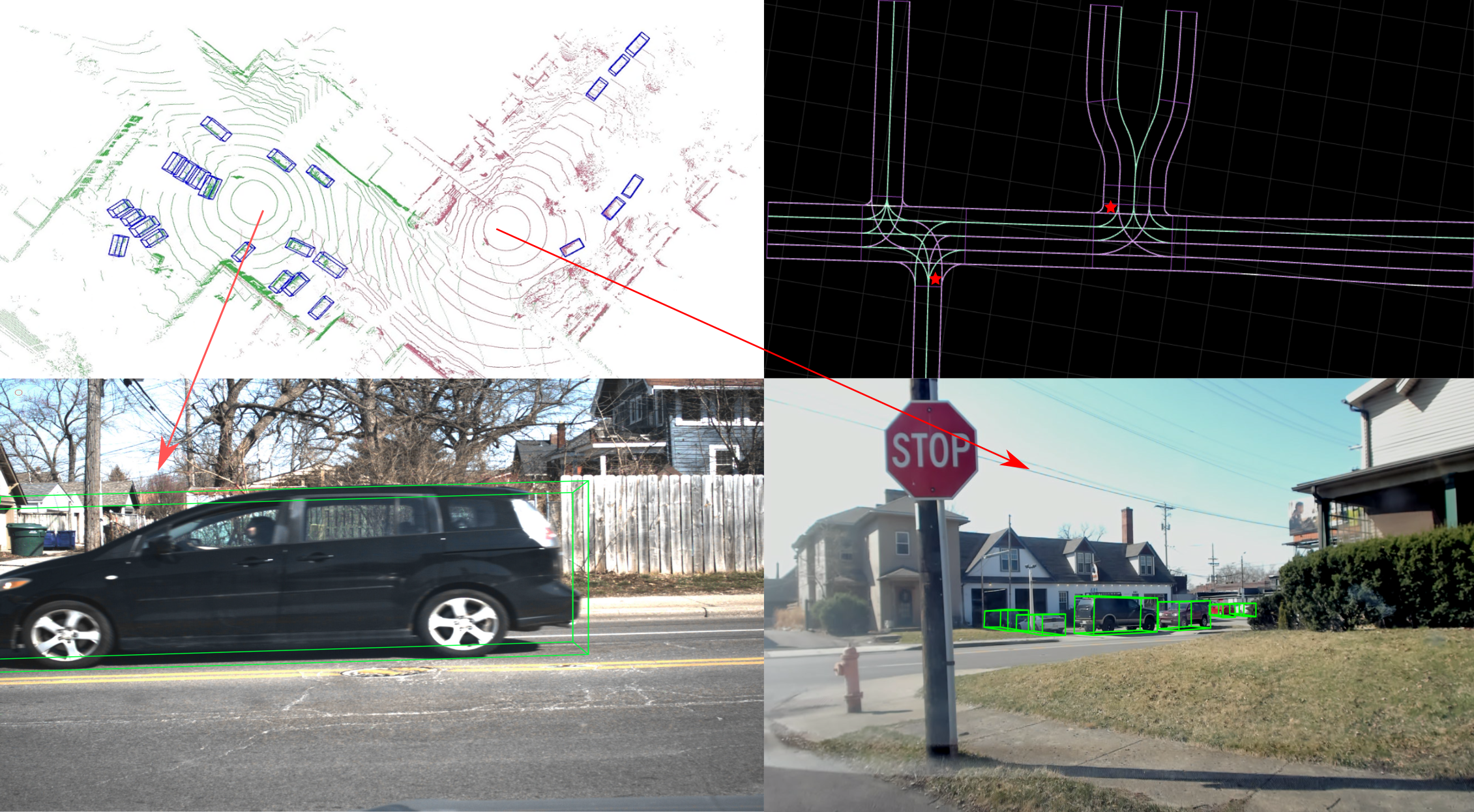} \\
(b)  A cityroad example \\
\end{tabular}
\caption{\textbf{3 different scenarios in V2V4Real.} \textit{Left Up}: The aggregated 3D LiDAR. \textit{Right Up}: The annotated HDMap, where the stars indicate the position of the two collection vehicles. \textit{Bottom Row}: The front cameras of the two vehicles. }
\label{fig:sup-dataset}
\end{figure*}

We demonstrate what each object class looks like in the LiDAR data in~\cref{fig:route}.
We show more visualizations of the proposed V2V4Real dataset in~\cref{fig:sup-dataset}. 3 different scenes are presented: a cityroad, a highway, and another cityroad example. For each figure, we demonstrate four images. Upper left: the aggregated 3D LiDAR points; Upper right: the annotated HDMap; Bottom row: the front camera view of the two vehicles, with the green and red LiDAR corresponding to lower left and lower right images. The 3D bounding boxes drawn on the images are projected from the labels annotated in LiDAR frames using extrinsics and intrinsics.

\begin{figure}[!htb]
    \centering

    \includegraphics[width=\columnwidth]{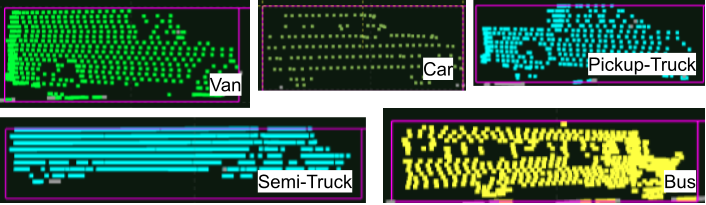}
    \caption{The LiDAR appearance for each object class in V2V4Real.  }
    \label{fig:route}
    \vspace{-5mm}
\end{figure}

\section{Implementation Details}
\label{sec:imp-details}

We provide additional details on the implemented baseline methods in our experiments.
\subsection{Cooperative 3D Object Detection}
\textbf{PointPillars backbone.}
For all the experiments, we set the PointPillars backbone~\cite{lang2019pointpillars} to have a voxel resolution of 0.4 meters on $x$ and $y$ direction. We set the maximum points per voxel as $32$ and the maximum voxel numbers as $32000$.

\textbf{Fusion models.}
We have implemented five different fusion methods including F-Cooper~\cite{chen2019cooper}, AttFuse~\cite{xu2022opv2v}, V2VNet~\cite{wang2020v2vnet}, V2X-ViT~\cite{xu2022v2xvit}, and CoBEVT~\cite{xu2022cobevt}. We mainly follow the implementation and configurations from the original authors, except for V2X-ViT, wherein we regard the two vehicles as the same object type (i.e., vehicle) since there is no infrastructure in V2V4Real.

\textbf{Detection head.}
For 3D object detection, we apply two channel-wise convolution $1\times 1$ layers on top of the fused feature maps to obtain two heads for box regression and classification, respectively. The regression head yields $(x,y,z,w,l,h,\theta)$, denoting the position $x,y,z$, size $w,l$, and yaw angle $\theta$ of the predefined anchor boxes. The classification head outputs the confidence score of being an object or background for each anchor box, respectively.
We employ the smoothed $\ell_1$ loss and a focal loss for regression and classification heads.

\begin{table*}[!t]
\centering
\begin{tabular}{l|cccc|cccc|c}
\cline{2-5}
\hline

\multirow{2}{*}{\textbf{Method}} & \multicolumn{4}{c|}{\textbf{Sync} (AP@IoU=0.5)} & \multicolumn{4}{c|}{\textbf{Async} (AP@IoU=0.5)} &  \multirow{2}{*}{\begin{tabular}[c]{@{}c@{}}\textbf{AM}\\ (MB)\end{tabular}}\\
                        & Overall  & 0-30m & 30-50m & 50-100m & Overall  &0-30m & 30-50m & 50-100m  & \\ \hline
No Fusion               & 28.7\color{gray}{(-11.1)}         & 50.0      &  22.9       &   4.9       &    28.7\color{gray}{(-11.1)}      &   50.0    &      22.9   &     4.9 & 0    \\\arrayrulecolor{lightgray}\hline
Late Fusion             & 43.0\color{gray}{(-12.0)}       &  55.1     &    34.4     &   31.9      &  40.9\color{gray}{(-9.3)}         &      54.6 &    33.5     &   30.9 & 0.003     \\\arrayrulecolor{lightgray}\hline
Early Fusion            & 48.2\color{gray}{(-11.5)}       &  64.3     &    33.2     &   34.0       & 41.0 \color{gray}{(-11.1)}         &  62.5     &  27.3      &   18.1 & 0.96      \\\arrayrulecolor{lightgray}\hline
F-Cooper~\cite{chen2019f}                & 45.6\color{gray}{(-15.1)}       &  65.3     &    35.3     &    25.9      & 37.6\color{gray}{(-16.0)}            &   62.1    &    28.6     &   13.4 & 0.20        \\
V2VNet~\cite{wang2020v2vnet}                  & 49.0\color{gray}{(-15.5)}       &  69.2     &   35.0      &    30.6      & 41.5\color{gray}{(-14.9)}         &  65.5     &    32.4     &   13.7 & 0.20        \\
AttFuse~\cite{xu2022opv2v}                 & 47.9\color{gray}{(-16.8)}     &  67.9    &   34.6     &  26.8     & 40.8\color{gray}{(-16.9)}    & 65.8     &    28.6    &     13.9  & 0.20    \\
V2X-ViT~\cite{xu2022v2xvit}                 & 48.9\color{gray}{(-16.0)}    &  66.0   &    38.1    &   30.0      &  41.6\color{gray}{(-14.3)}    &  62.8     &    32.9    &    17.5  & 0.20      \\
CoBEVT~\cite{xu2022cobevt}                  & 51.1\color{gray}{(-15.4)}    &  69.3     &   40.0     &   32.4  & 44.9 \color{gray}{(-13.7)}      & 65.2
&    35.6     &   19.8 & 0.20  
\\\arrayrulecolor{black}\hline
\end{tabular}
\vspace{-2mm}
\caption{Cooperative 3D object detection benchmark \textbf{without data augmentation} . The numbers in the bracket indicate the performance drop compared to the same model with data augmentation.}
\label{tab:augmentation}
\end{table*}
\subsection{Cooperative Tracking}
The proposed cooperative tracking framework in our benchmark follows the widely adopted tracking-by-detection paradigm but differs from the existing object tracking methods: the detection results are gained from shared visual information instead of individuals.

\textbf{Problem definition:}  Assume there are $N_t$ bounding boxes $D(t) = \{D_t^i\}_{i=1}^{N_t} $ from our cooperative detection algorithm  at current frame $t$, where $N_t$ represent the number of cooperatively detected objects and $D_t^i = (x_t^i , y_t^i , z_t^i , \theta_t^i , w_t^i , h_t^i , l_t^i , s_t^i )$. There are $M_{t-1}$  previous associated trajectories $T(t-1) = \{T_{t-1}^j\}_{j=1}^{M_{t-1}}$ at frame $t-1$, where $M_{t-1}$ represent the number of previous trajectories and $T_{t-1}^j = (x_{t-1}^j , y_{t-1}^j , z_{t-1}^j , \theta_{t-1}^j , w_{t-1}^j , h_{t-1}^j , l_{t-1}^j , s_{t-1}^j, v_{x_{t-1}}^j, v_{y_{t-1}}^j, 
\\v_{z_{t-1}}^j)$. We aim to map each detected object $D_t^i$ to their corresponding trajectory $M_{t-1}^j$. The $(x, y, z)$ corresponds to the object center, and $(w, h, l)$ represents object size in point cloud space. $\theta$ is the heading angle of the object. $s$ is the detection confidence score, which depends on the cooperative detection network. The additional variables $(v_x, v_y, v_z)$ in trajectories represent the object velocity in $x$, $y$, and $z$ directions. 

\textbf{Detection results:} The inputs of the tracking system are the detected bounding boxes, which are obtained from the cooperative 3D object detection task described in the previous section.

\textbf{Trajectory prediction and association:} With the detected bounding boxes from the cooperative detection module, the goal of object tracking is to find all valid matches between detected bounding boxes $D(t)$ and trajectory $T(t-1)$. A Kalman filter is applied to predict the trajectories $T(t-1)$ of objects based on a constant velocity kinematic vehicle model. These predicted spatial information of trajectories combined with the information of detected objects would be used to calculate the affinity matrix in the Hungarian to determine whether currently detected objects in $D(t)$ can be matched to trajectories in $T(t-1)$. Specifically, given a sequence of trajectories 
\begin{align}
T(t-1) = \{T_{t-1}^j\}_{j=1}^{M_{t-1}}
\end{align} 
at frame $t-1$, a constant velocity kinematic vehicle model in the Kalman filter is used to predict the position of the object in each trajectory in $T(t-1)$ as follows:
\begin{align}
x_{t,pred}^j = x_{t-1}^j + v_{x_{t-1}}^j \\
y_{t,pred}^j = y_{t-1}^j + v_{y_{t-1}}^j \\
z_{t,pred}^j = z_{t-1}^j + v_{z_{t-1}}^j
\end{align} 
Therefore, the final predicted trajectory is
\begin{align}
T_{t,pred}^j = (x_{t,pred}^j , y_{t,pred}^j , z_{t,pred}^j , \theta_{t-1}^j ,w_{t-1}^j,\\ h_{t-1}^j , l_{t-1}^j , s_{t-1}^j,v_{x_{t-1}}^j, v_{y_{t-1}}^j, v_{z_{t-1}}^j)
\end{align}

After predicting the set of trajectories $T(t)_{pred}$, the 3D Intersection of Union(IoU) is used to compute the data affinity matrix  $A \in \mathbf{R}^ {M_{t-1} \times N_{t}}$ to determine the similarity between predicted trajectories and detected bounding boxes $D(t)$, where each element $A_{i,j}$ is the 3D IoU for the predicted trajectory $i$ and the 3D bounding box $j$ at frame $t$. The affinity matrix will be solved by the Hungarian algorithm, which considers the association as a bipartite matching problem, to solve the association problem. 

\textbf{State update and trajectory management:} After gaining the predicted trajectories and association results from the Hungarian algorithm, the state update is to make the trajectory more accurate by considering the current detection results. The Kalman Filter is used to update the state of the predicted trajectory by considering the current detection information and accounting for uncertainties from the detection errors. Accordingly, we have: 
\begin{align}
T_{t}^{m} = KF(T_{t}^{m}, D_{t}^k)
\end{align}
,where $D_{t}^k \in D(t)$ and $T_{t}^{m} \in T(t)$ are the associated pair obtained from Hungarian algorithm, $k \in \{1, 2, ..., N_{t}\}$, $m \in \{1, 2, ..., M{t}\}$. The updated state of the corresponding predicted trajectory $T_{t}^{m} \in T(t)$ is a weighted average between the related $D_{t}^k \in D(t)$ and $T_{t}^{m} \in T(t)$.

Trajectory management is organizing new and old trajectories. When an object starts to appear at frame $t$, it could either be a false positive due to the detector or it naturally enters the field of view. Similarly, when an object starts to disappear at frame $t$,  it could either be a miss or it naturally leaves the LiDAR range. Both scenarios are handled by tracking objects in additional frames. Specifically, when $D_t^l$ is an unmatched object entering the field of view, we will treat it as a new trajectory if $D_t^l$ can be matched in the next few frames to prevent adding false positive detection as a new trajectory. When $T_t^p$ is an unmatched trajectory leaving the field of view, we will treat it as a dead trajectory if $T_t^p$ cannot be matched with any detected bounding boxes in the next couple of frames to prevent removing the true positive trajectory. 

\subsection{Domain Adaption}
\textbf{Feature-level domain discriminator:} The feature-level domain discriminator will take the fused features after the fusion modules as input and classify whether the feature belongs to target domain~(V2V4Real) or source domain~(OPV2V). The discriminator consists of two convolution layers with $3\times3$ kernel size, and the second convolution will map the feature channel number to $1$. 

\textbf{Objec-level domain discriminator:} The object-level domain clarifier will take the score map obtained from the detection classification head as the input. It includes three linear projection layers with ReLU activation functions.

\textbf{Loss:} Both discriminators employ binary cross-entropy to compute the loss and use gradient reverse layer~(GRL)~\cite{ganin2015unsupervised} to backpropagate the gradients.

\section{Ablation Studies}
\label{sec:ablation-details}

\noindent\textbf{Effects of Data Augmentation.} Data augmentation has been shown to be highly effective in single-vehicle perception tasks, such as 3D object detection using pointclouds~\cite{lang2019pointpillars,zhou2018voxelnet}. In this work, we evaluate the impact of data augmentation on cooperative perception by conducting an ablation study that removes pointcloud rotation, flipping, and scaling augmentations. Our evaluation is performed on 3D object detection. As depicted in Table \ref{tab:augmentation}, all methods show a significant decrease in performance without data augmentation, such as 15.4\% for CoBEVT and 11.5\% for Early fusion. Additionally, the intermediate fusion methods show more benefits from data augmentation, which is possibly due to their more complex models and requirement for more data.
\section{Detection Results}
\label{sec:sup-detection}

\begin{figure*}[!ht]
\centering
\footnotesize
\def\xwidth{0.36}
\def\yheight{0.18}
\def\xem{-2pt}
\def\im_shift{0.1\textwidth}
\setlength{\tabcolsep}{0.5pt}
\begin{tabular}{cccc}
 & Scene 1 & Scene 2\\
 \multirow[t]{1}{*}[\im_shift]{\begin{sideways} No Fusion \end{sideways}} &
\includegraphics[ width=\xwidth\linewidth]{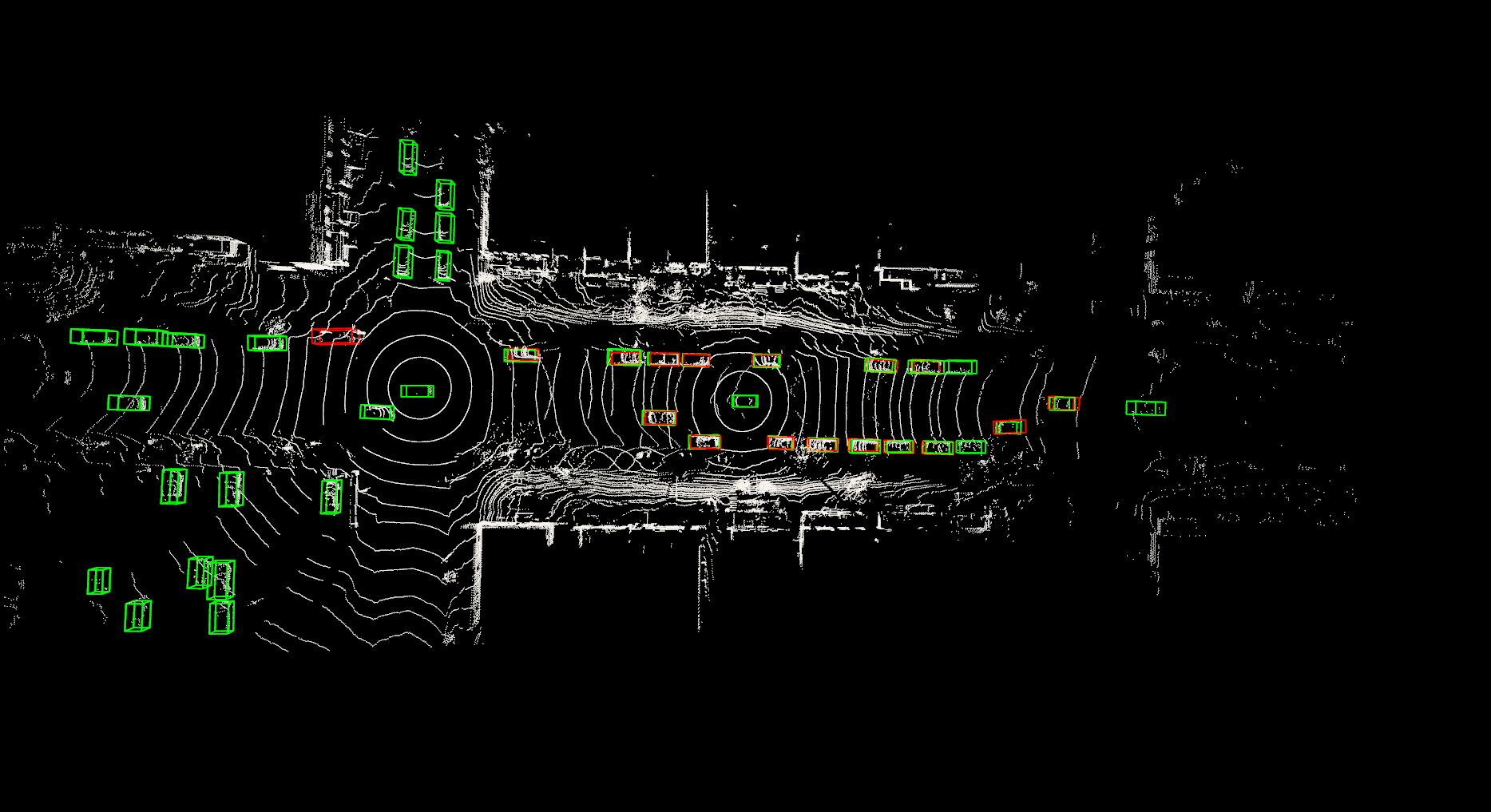}
& \includegraphics[ width=\xwidth\linewidth]{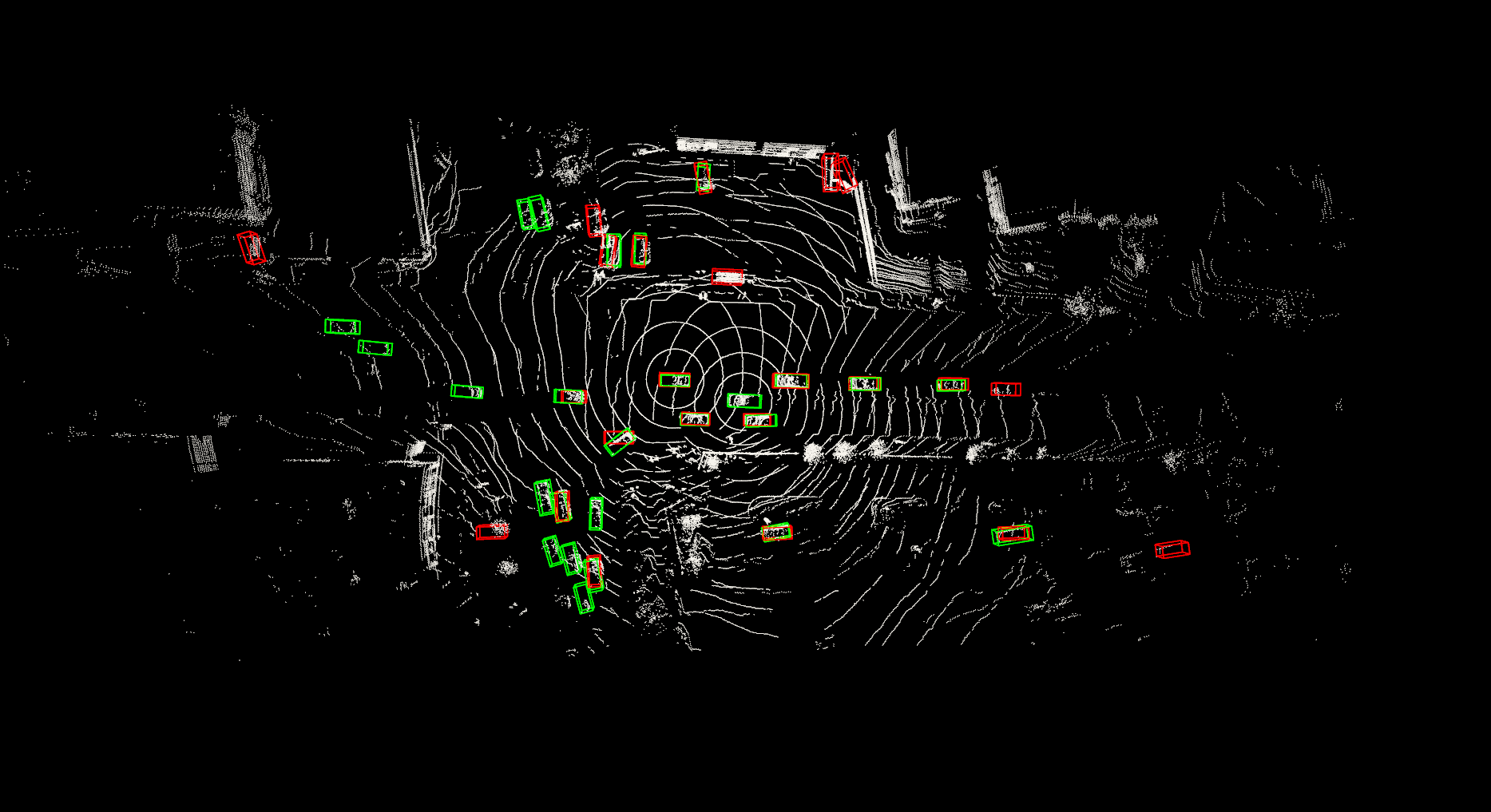}\\

\multirow[t]{1}{*}[\im_shift]{\begin{sideways}  Early Fusion  \end{sideways}}  &
 \includegraphics[ width=\xwidth\linewidth]{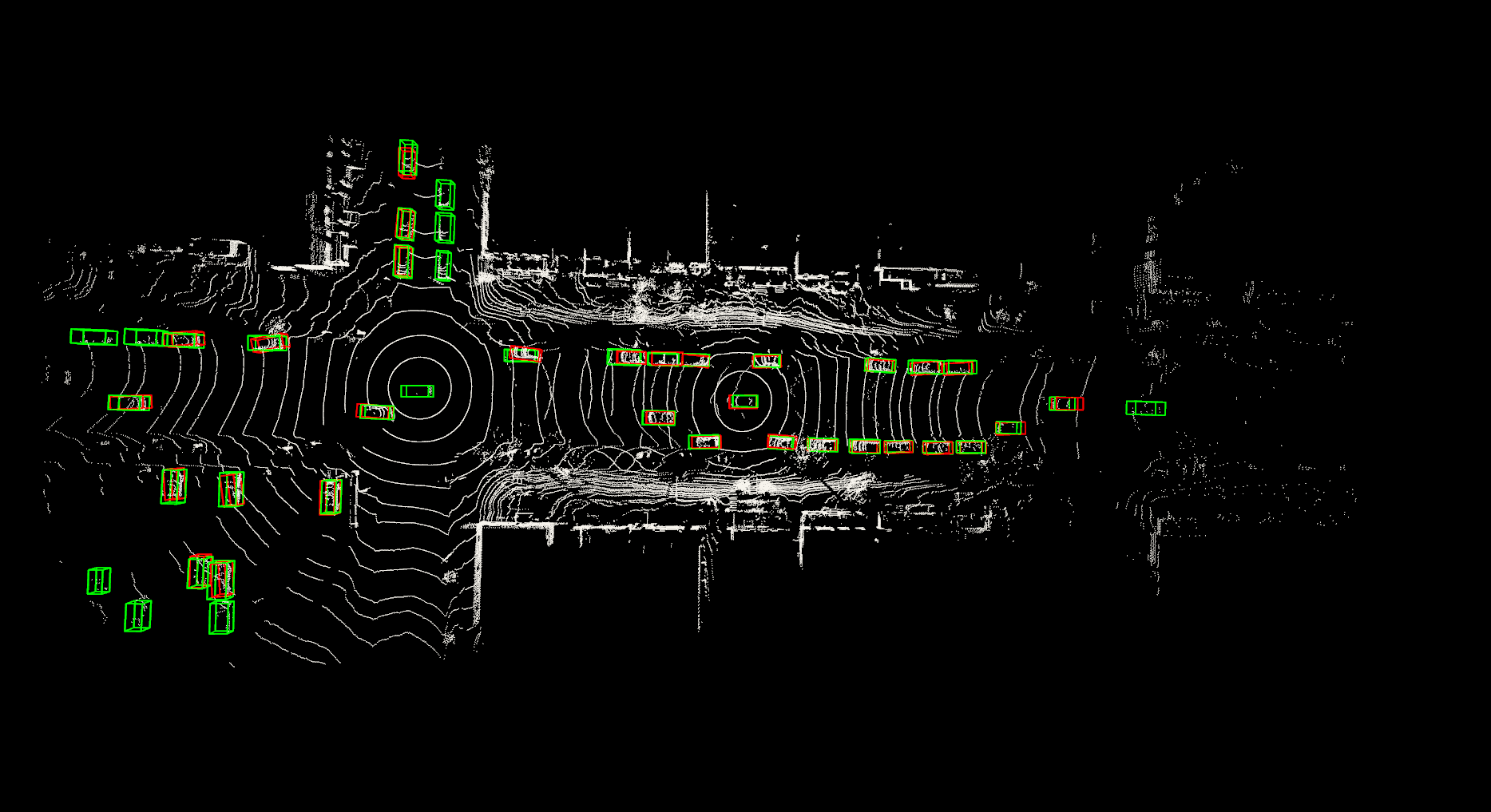}
& \includegraphics[ width=\xwidth\linewidth]{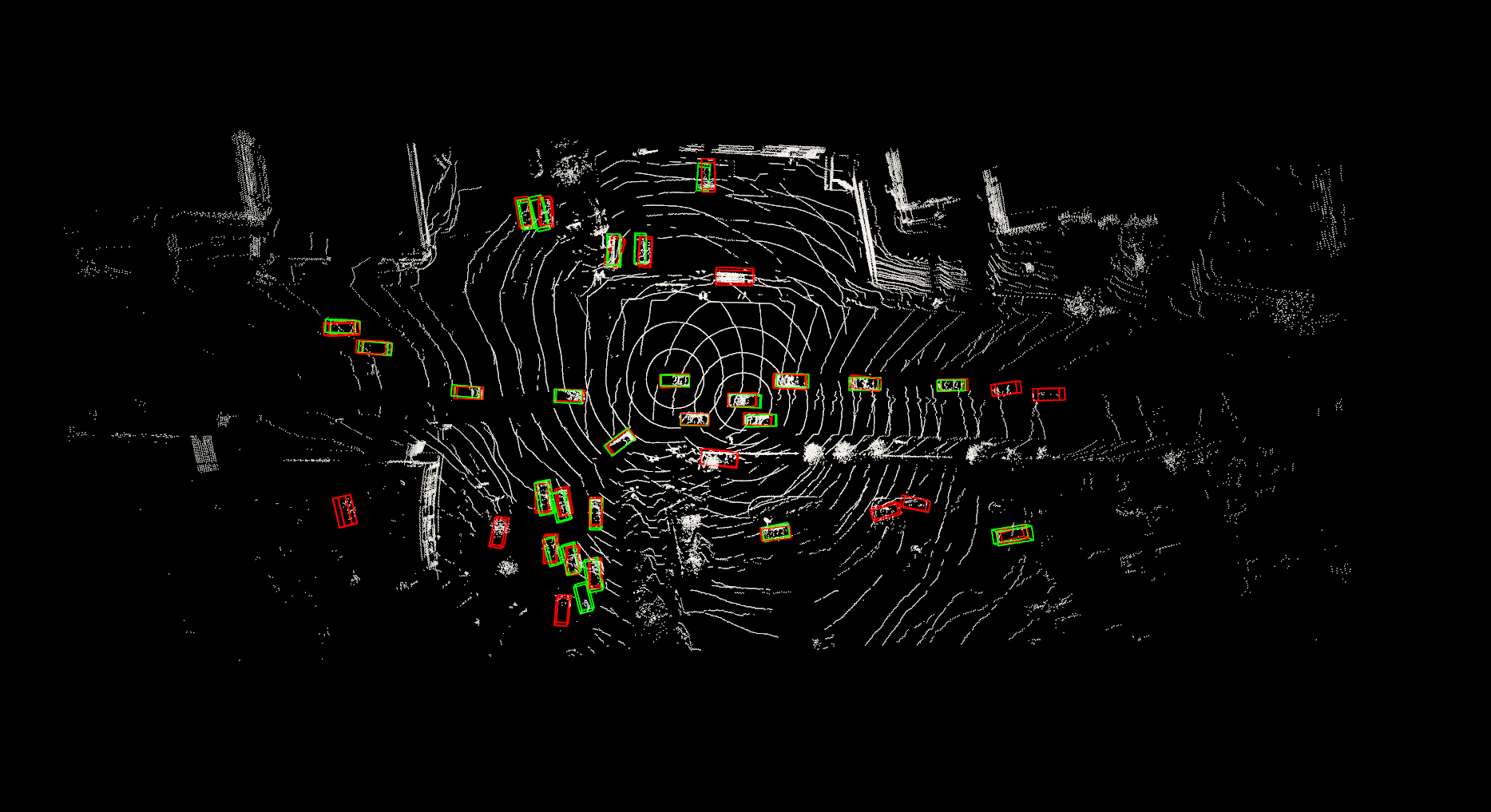}\\

\multirow[t]{1}{*}[\im_shift]{\begin{sideways}  Late Fusion  \end{sideways}}  &
 \includegraphics[ width=\xwidth\linewidth]{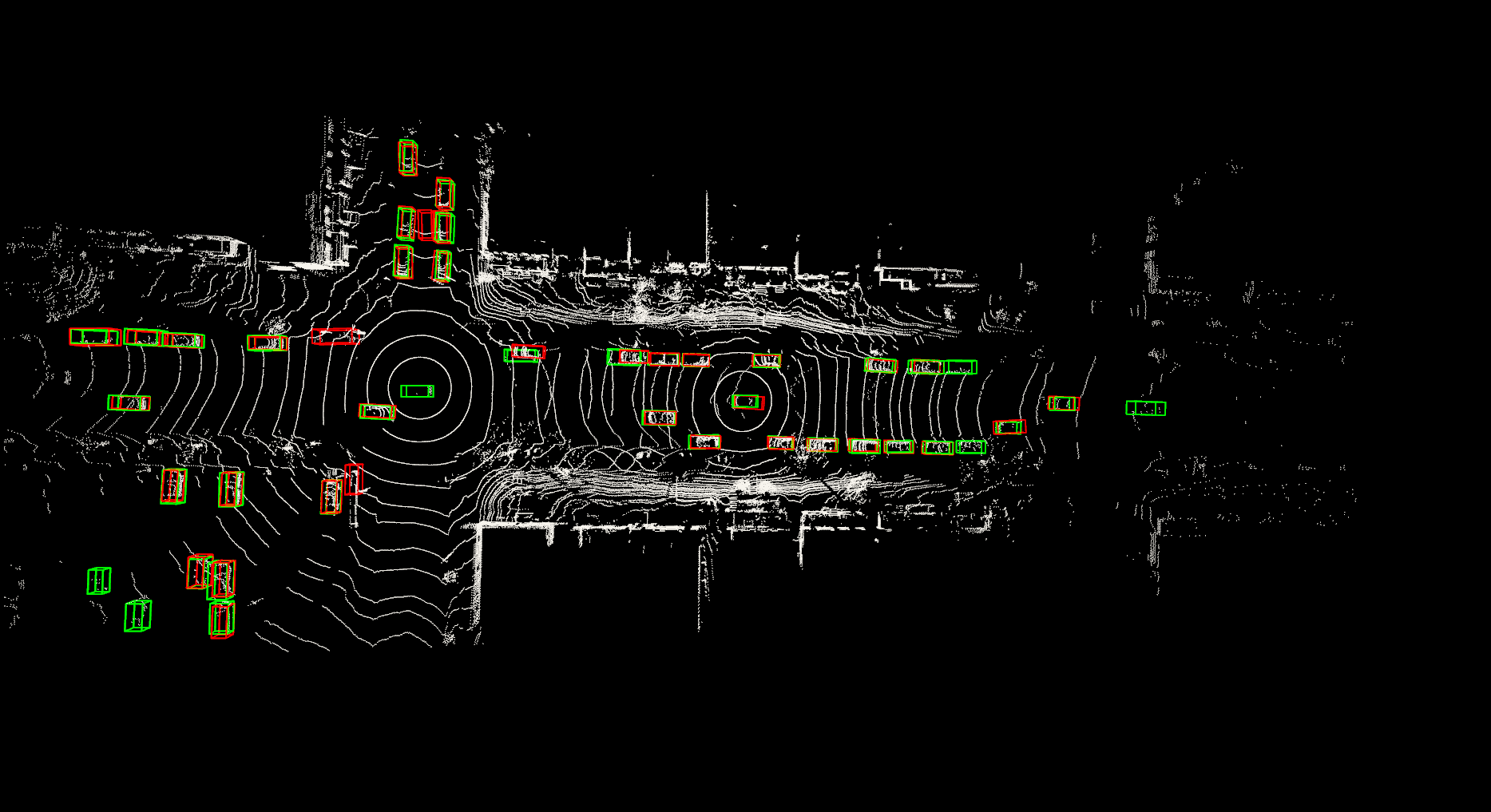}
& \includegraphics[ width=\xwidth\linewidth]{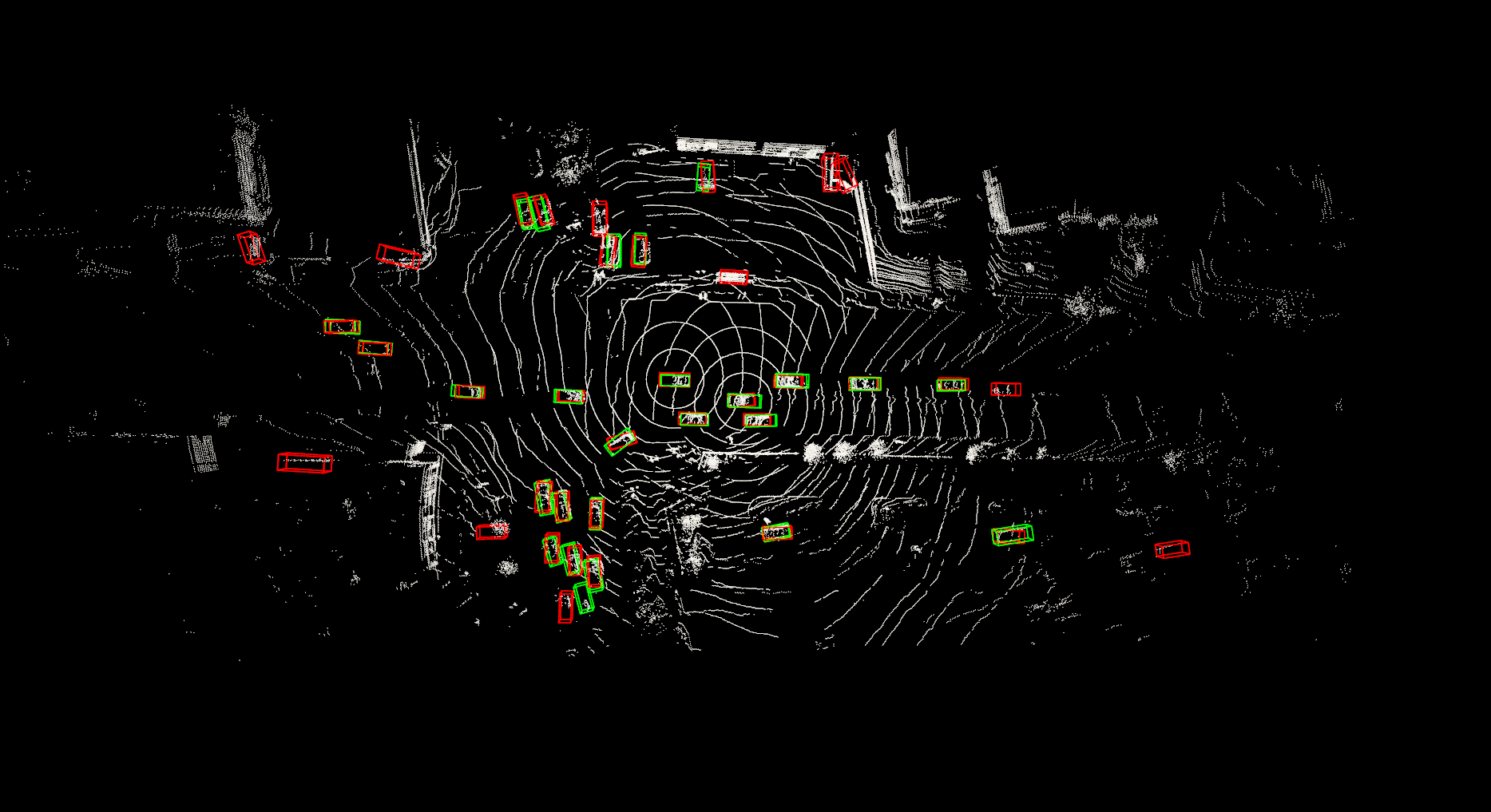}\\

\multirow[t]{1}{*}[\im_shift]{\begin{sideways}  V2VNet~\cite{wang2020v2vnet} \end{sideways}} &
\includegraphics[width=\xwidth\linewidth]{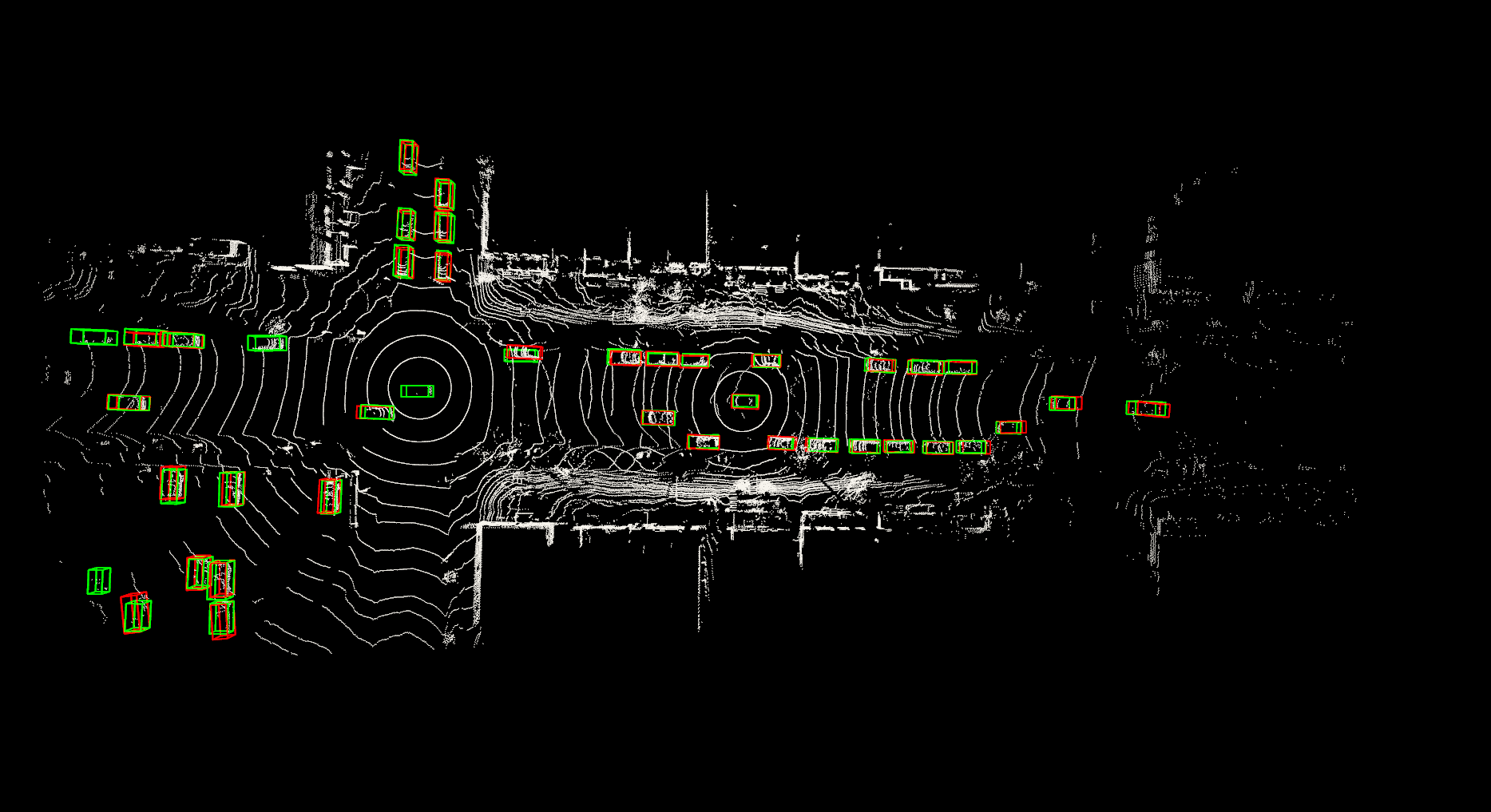}
& \includegraphics[width=\xwidth\linewidth]{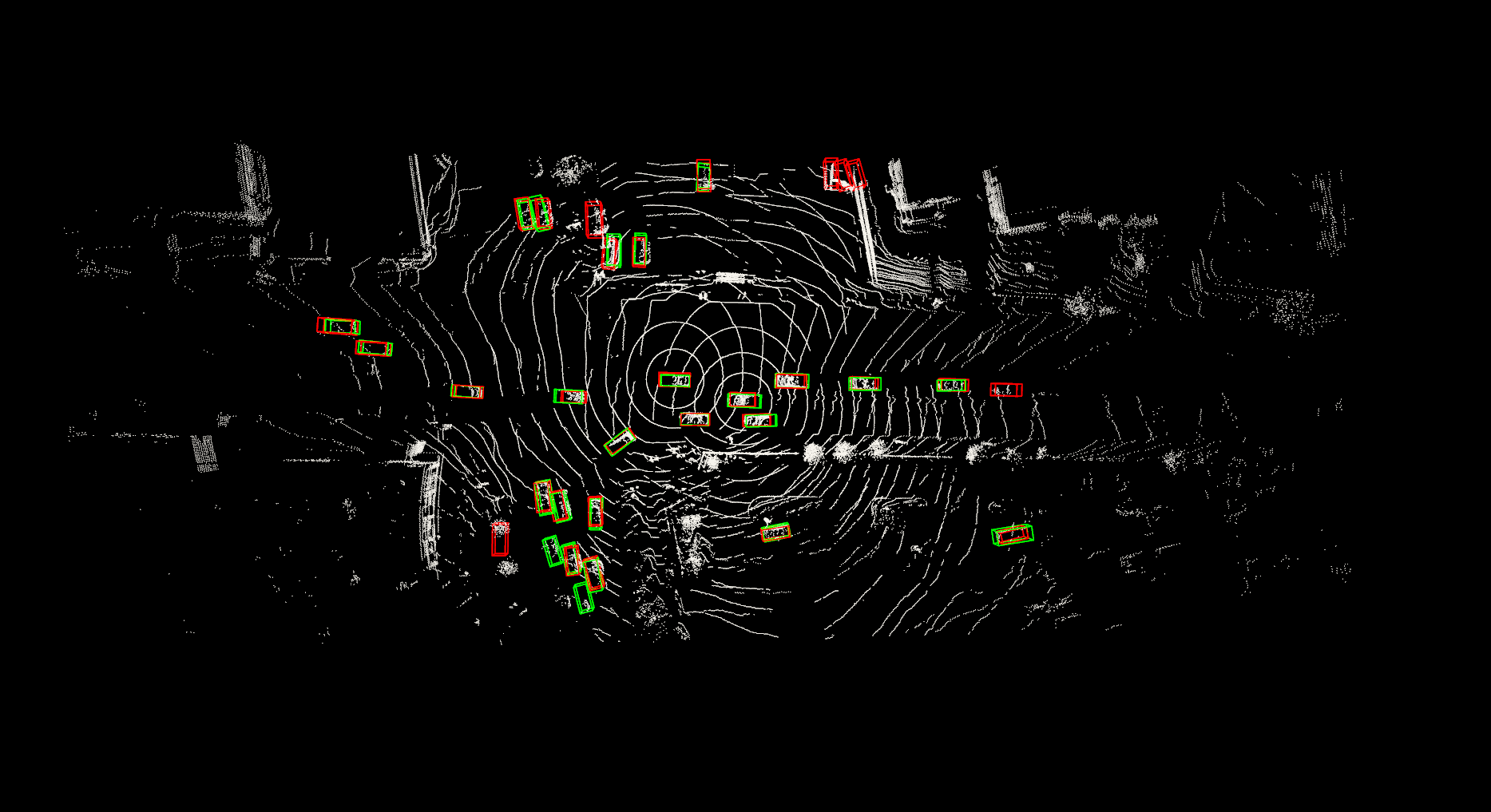}\\

\multirow[t]{1}{*}[\im_shift]{\begin{sideways}  V2X-ViT~\cite{xu2022v2xvit} \end{sideways}} &
 \includegraphics[ width=\xwidth\linewidth]{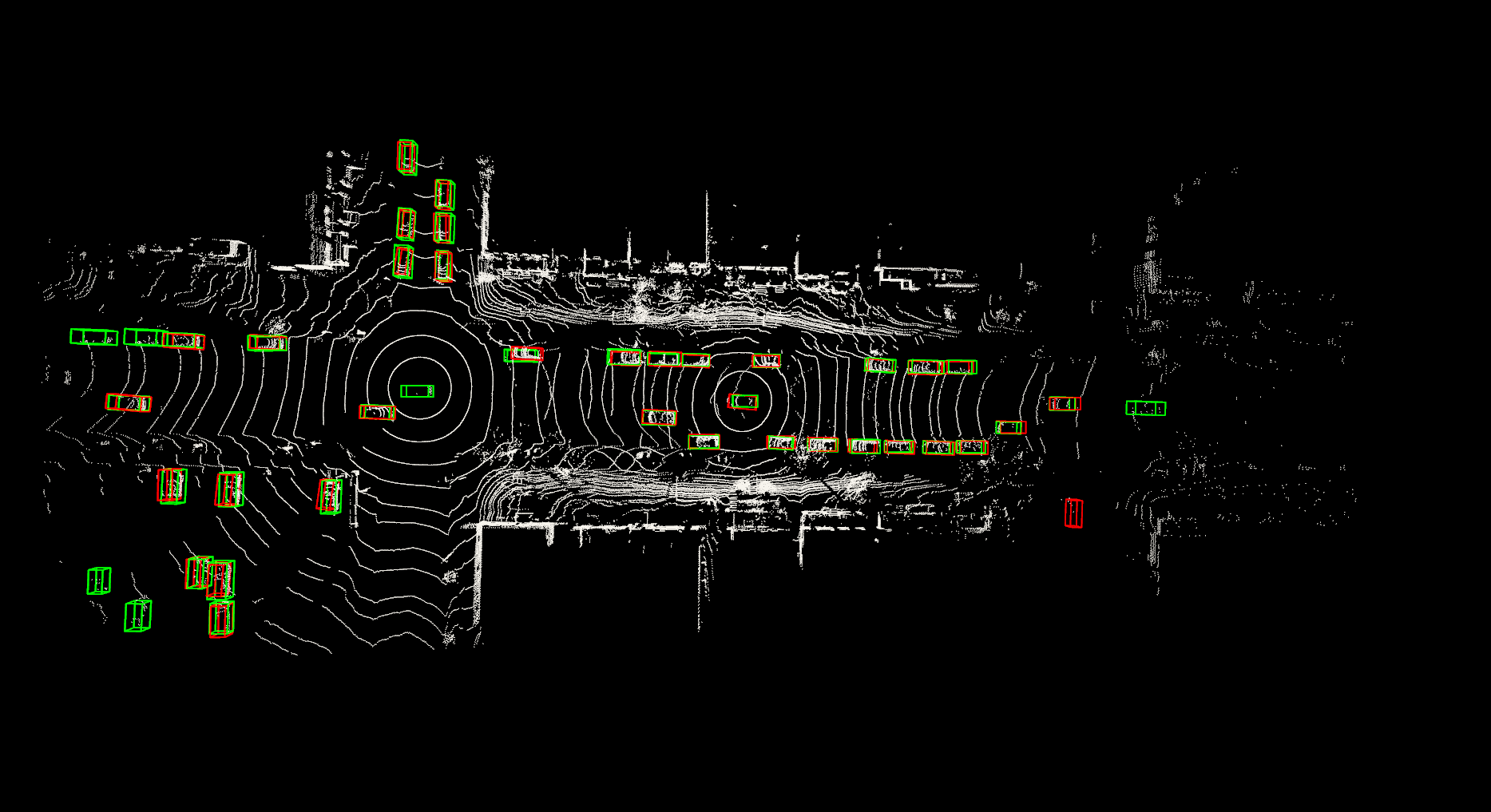}
& \includegraphics[ width=\xwidth\linewidth]{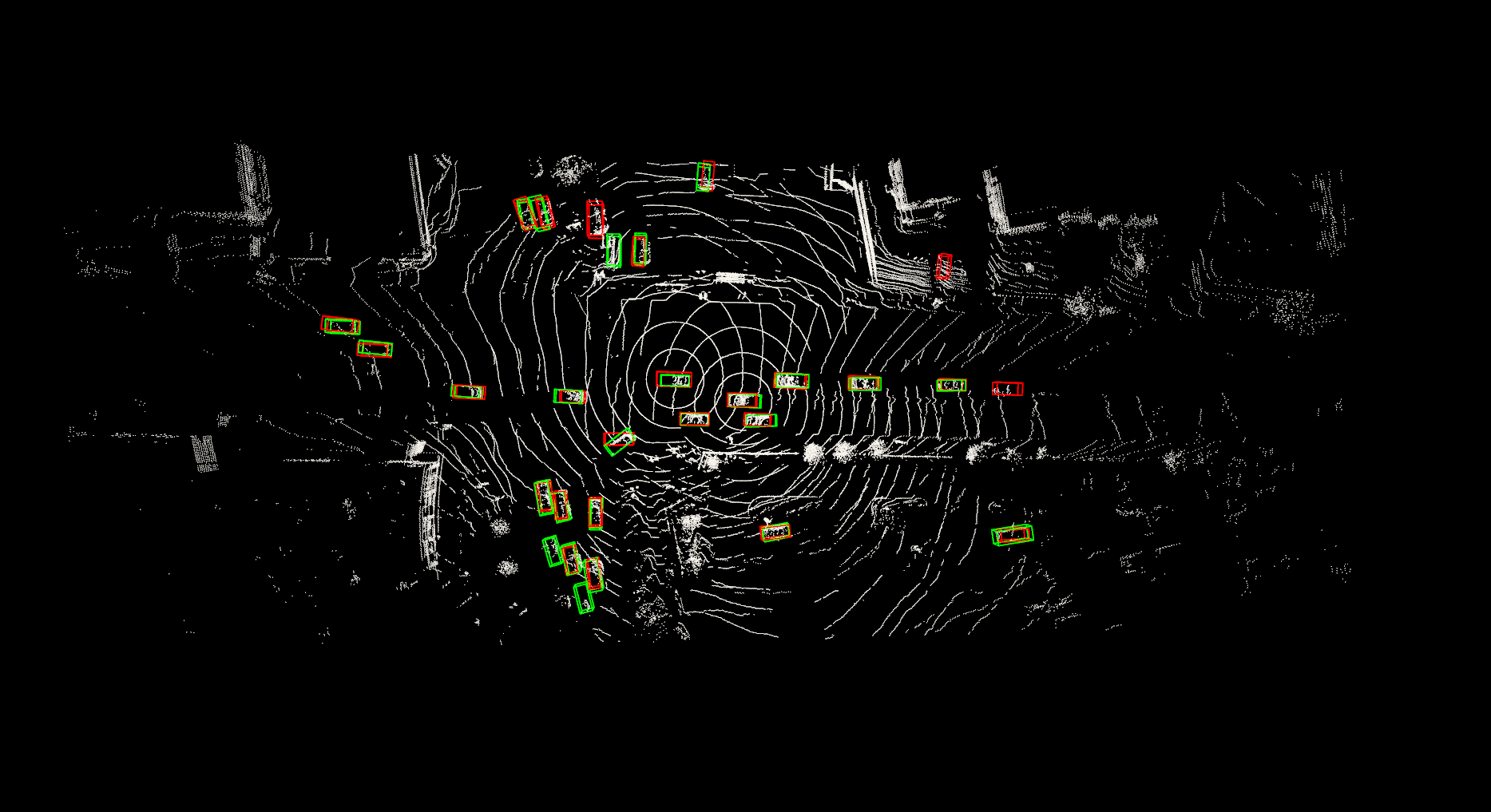}\\

\multirow[t]{1}{*}[\im_shift]{\begin{sideways}  CoBEVT~\cite{xu2022cobevt} \end{sideways}} &
 \includegraphics[width=\xwidth\linewidth]{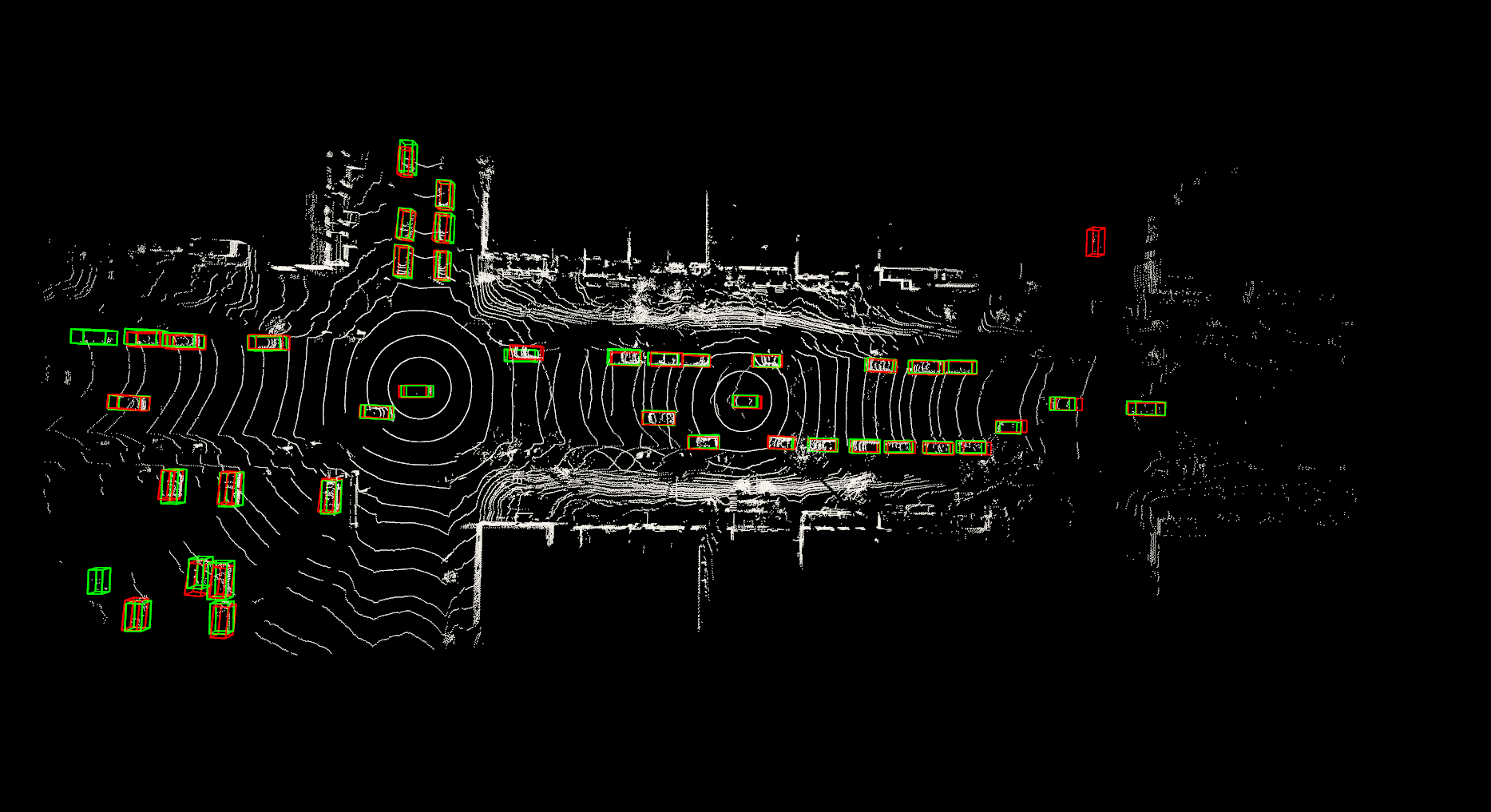}
 
& \includegraphics[width=\xwidth\linewidth]{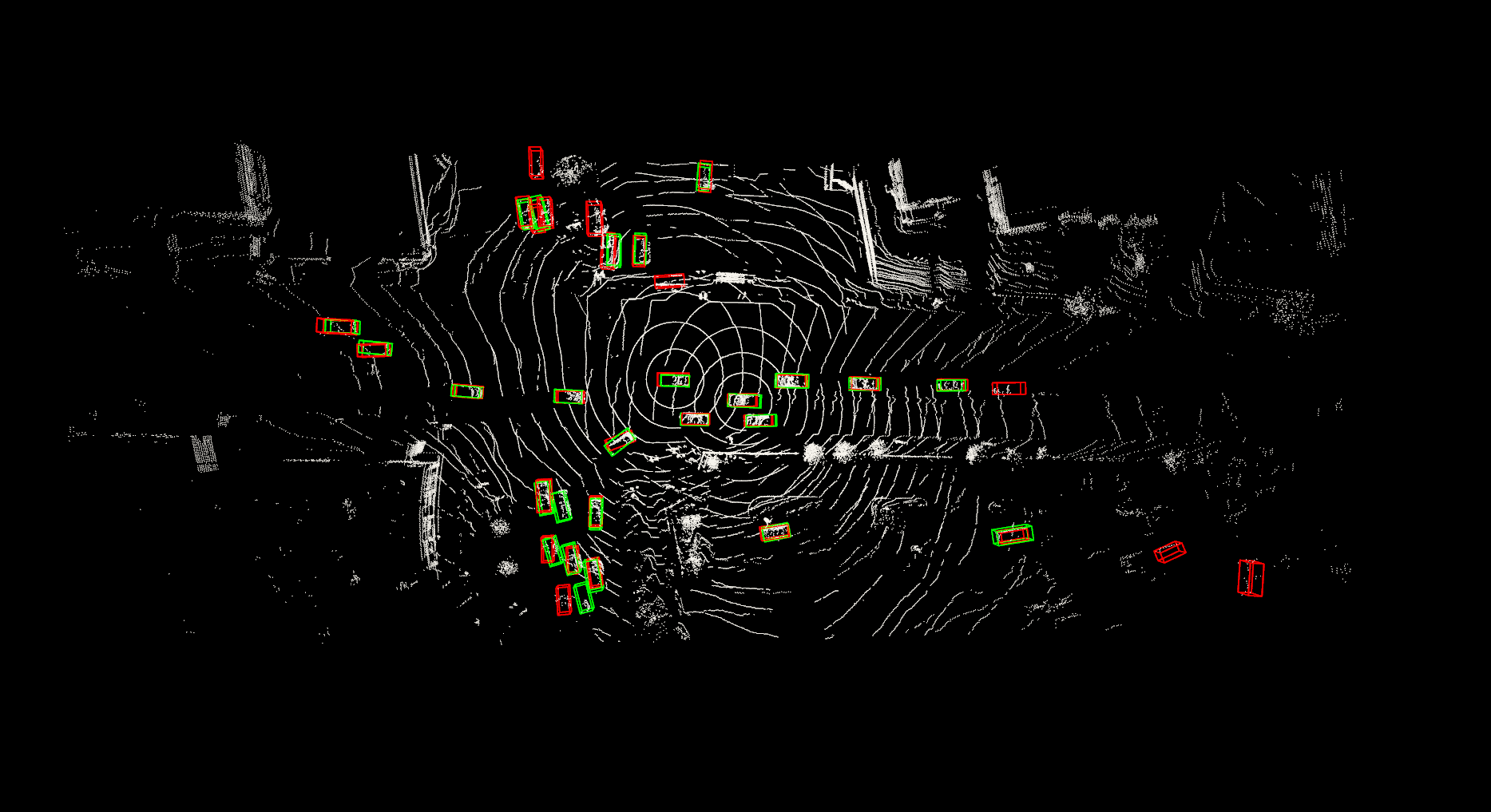}\\

\end{tabular}
\vspace{-3mm}
\caption{\textbf{Qualitative results of cooperative 3d object detection in two urban scenarios.} \textcolor{green}{Green} and \textcolor{red}{red} 3D bounding boxes represent the groundtruth and prediction, respectively.}
\label{fig:sup-qualitive3}
\end{figure*}

\begin{figure*}[!ht]
\centering
\footnotesize
\def\xwidth{0.36}
\def\yheight{0.18}
\def\xem{-2pt}
\def\im_shift{0.1\textwidth}
\setlength{\tabcolsep}{0.5pt}
\begin{tabular}{cccc}
 & Scene 3 & Scene 4\\
 \multirow[t]{1}{*}[\im_shift]{\begin{sideways} No Fusion \end{sideways}} &
\includegraphics[ width=\xwidth\linewidth]{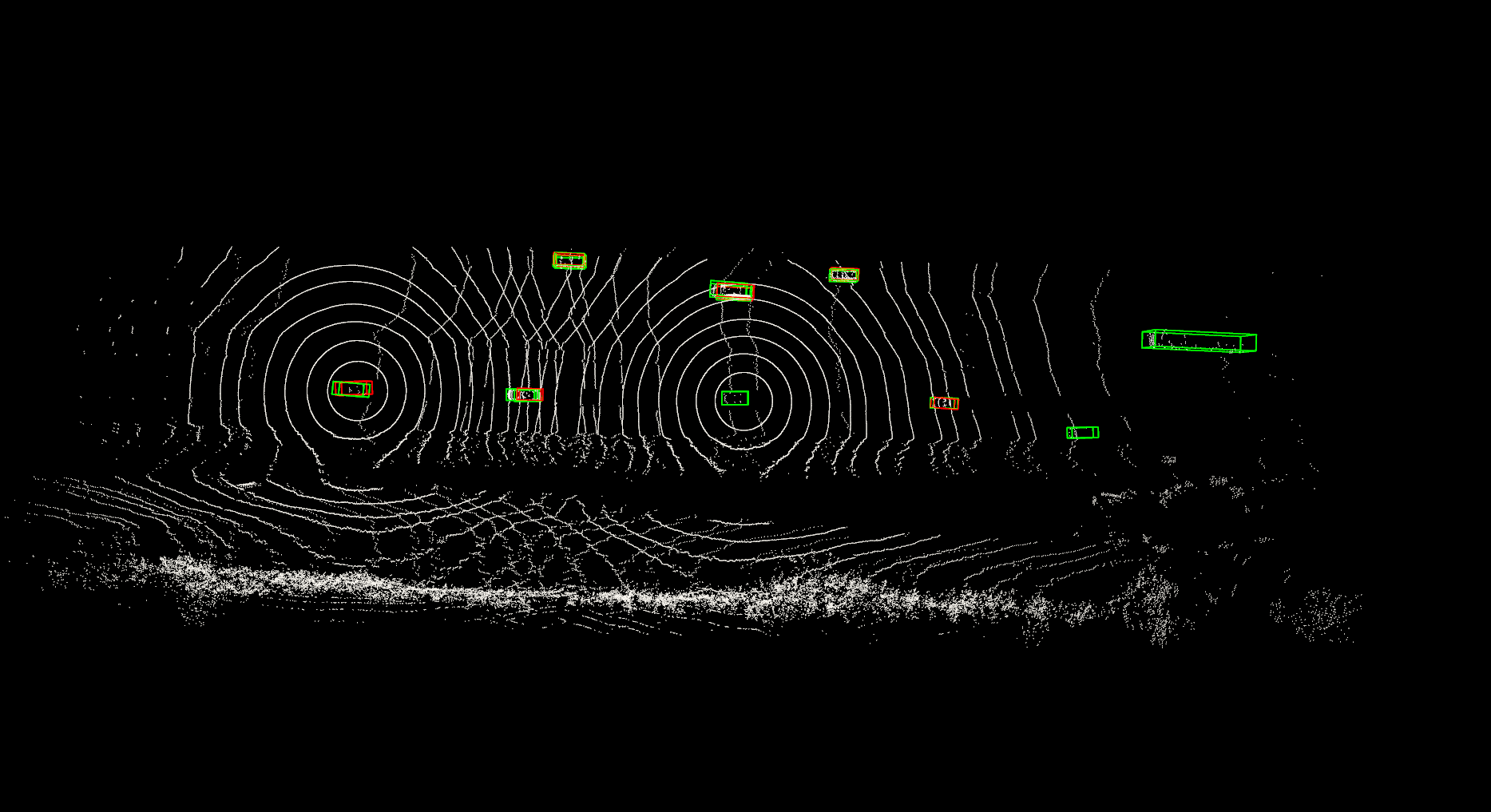}
& \includegraphics[ width=\xwidth\linewidth]{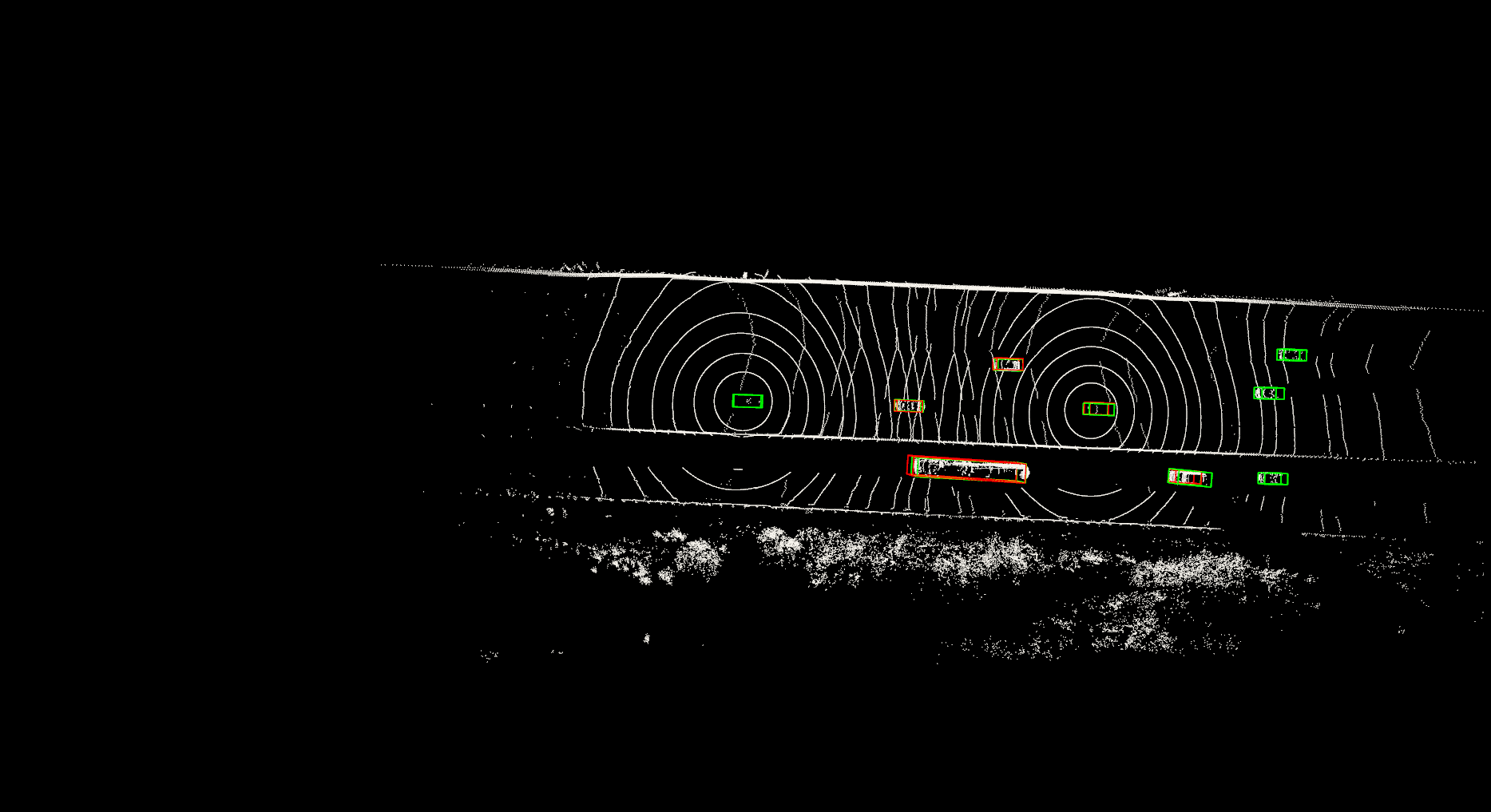}\\

\multirow[t]{1}{*}[\im_shift]{\begin{sideways}  Early Fusion  \end{sideways}}  &
 \includegraphics[ width=\xwidth\linewidth]{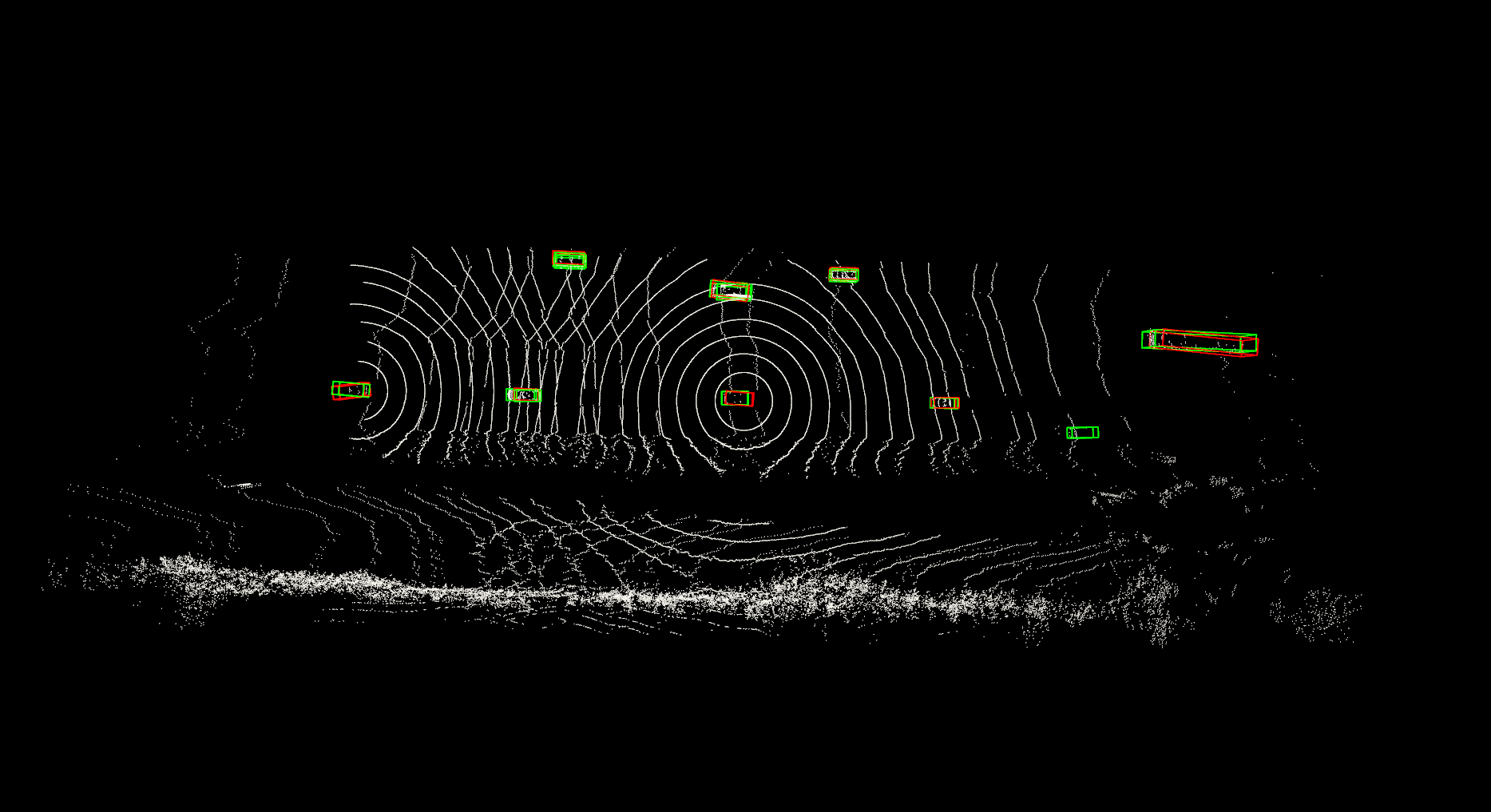}
& \includegraphics[ width=\xwidth\linewidth]{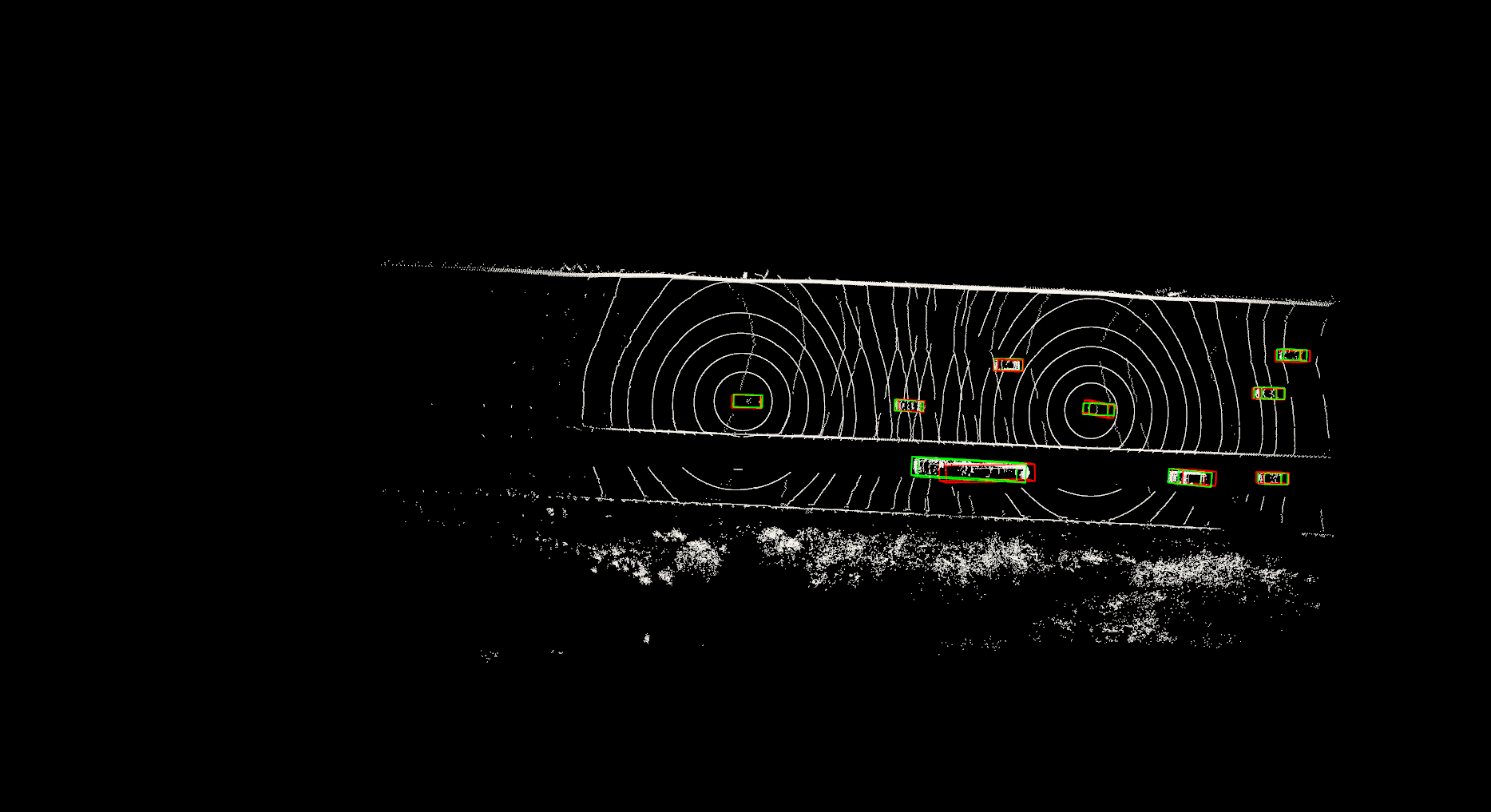}\\

\multirow[t]{1}{*}[\im_shift]{\begin{sideways}  Late Fusion  \end{sideways}}  &
 \includegraphics[ width=\xwidth\linewidth]{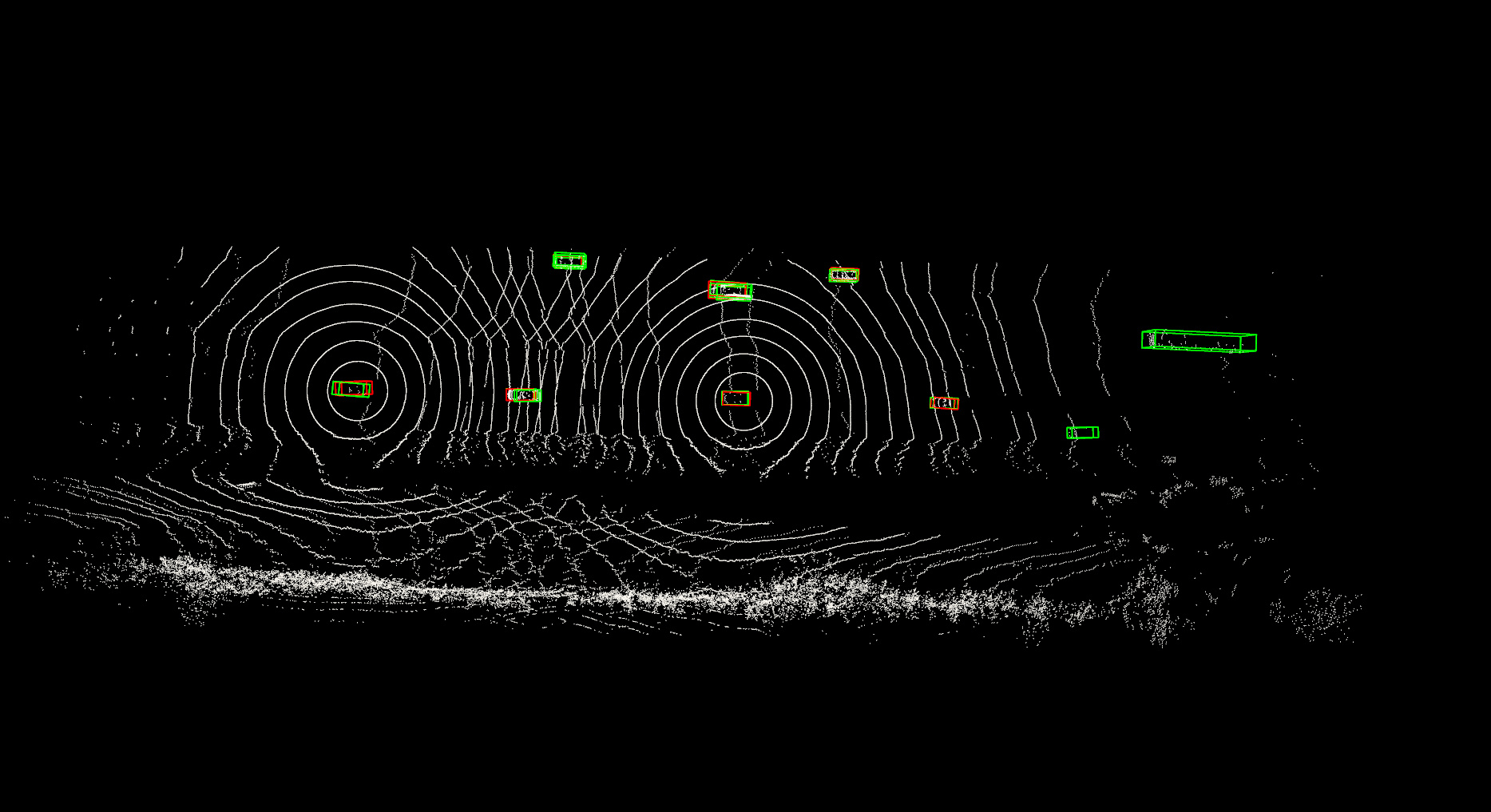}
& \includegraphics[ width=\xwidth\linewidth]{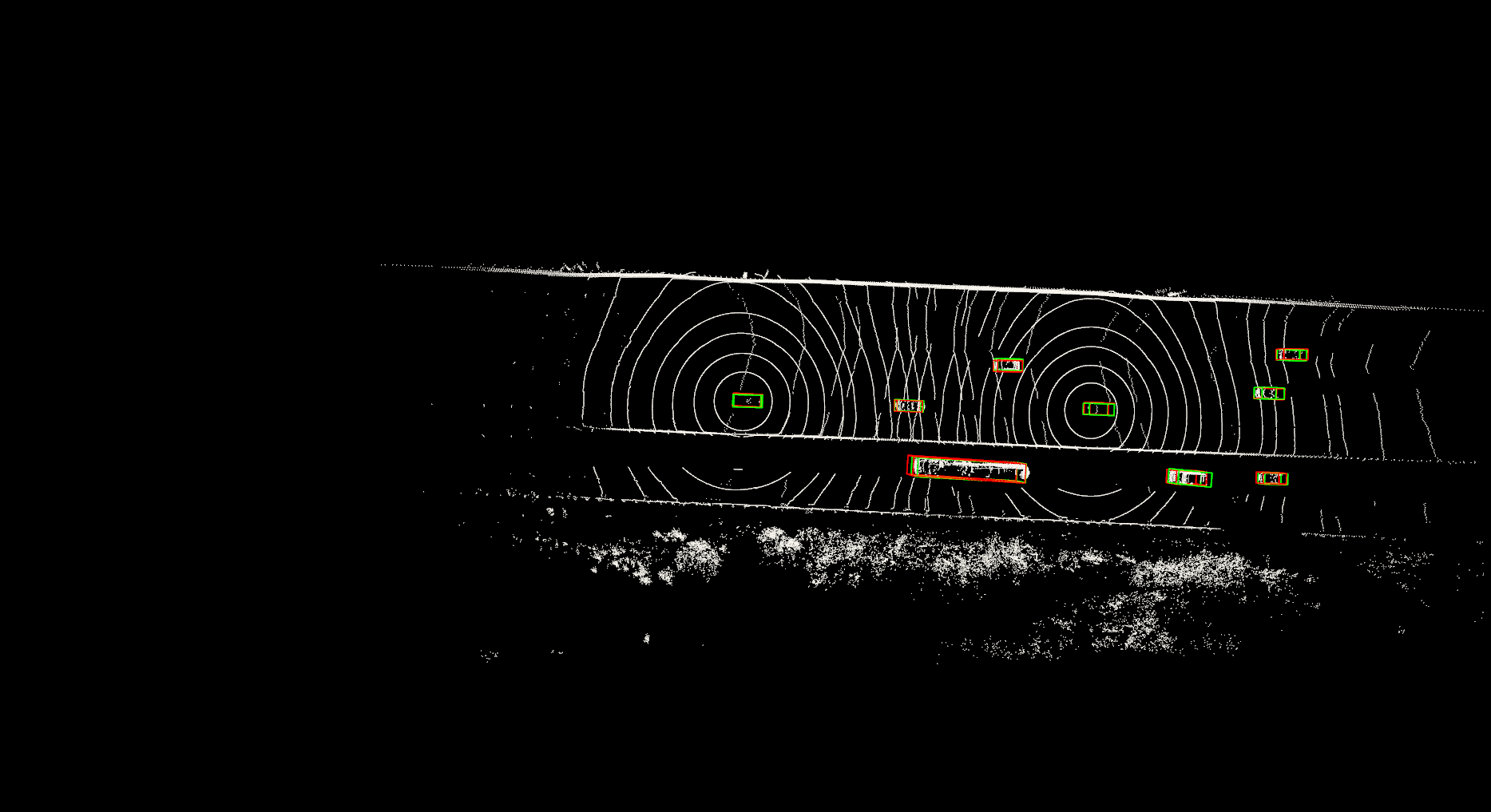}\\

\multirow[t]{1}{*}[\im_shift]{\begin{sideways}  V2VNet~\cite{wang2020v2vnet} \end{sideways}} &
\includegraphics[width=\xwidth\linewidth]{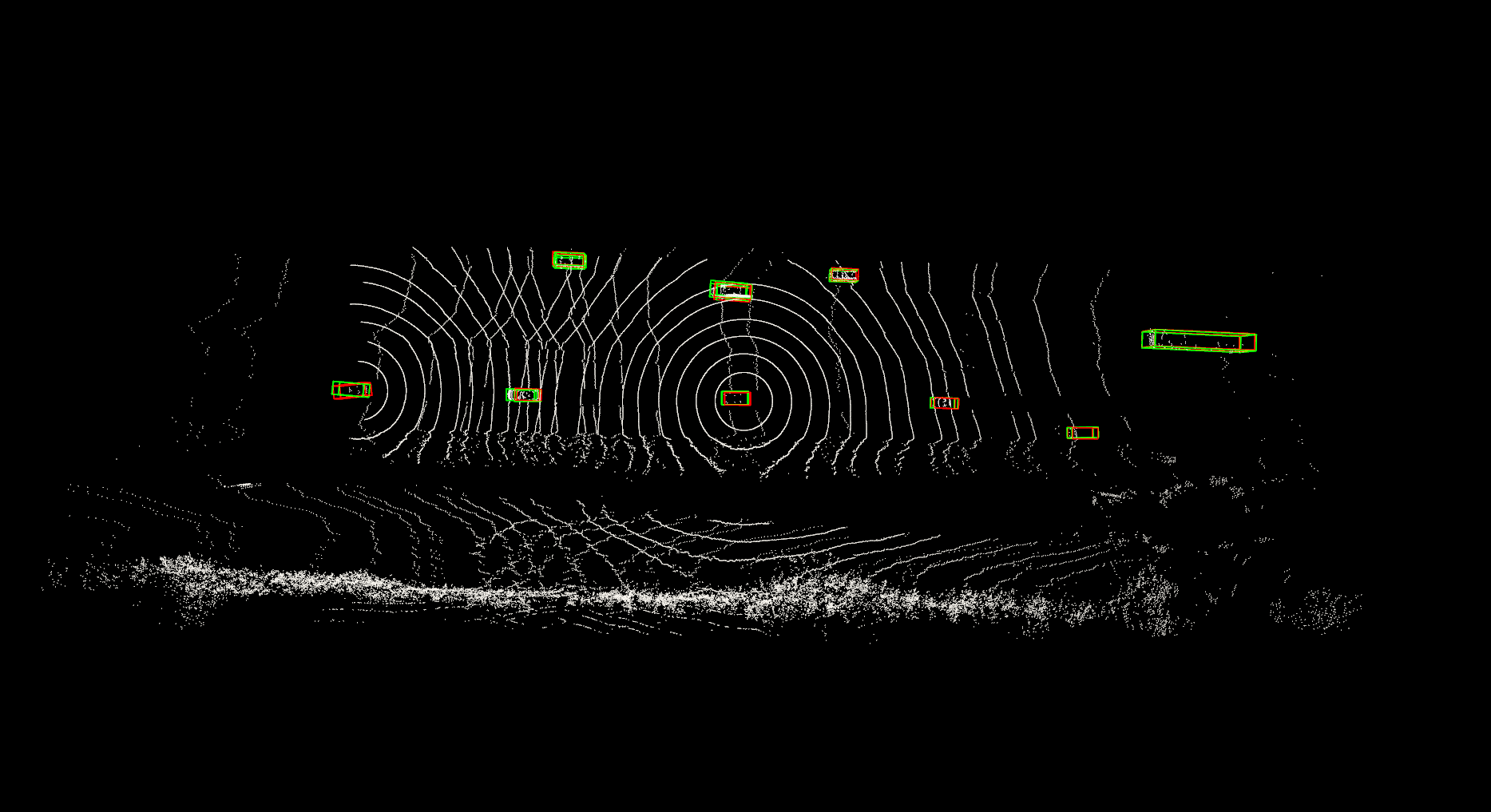}
& \includegraphics[width=\xwidth\linewidth]{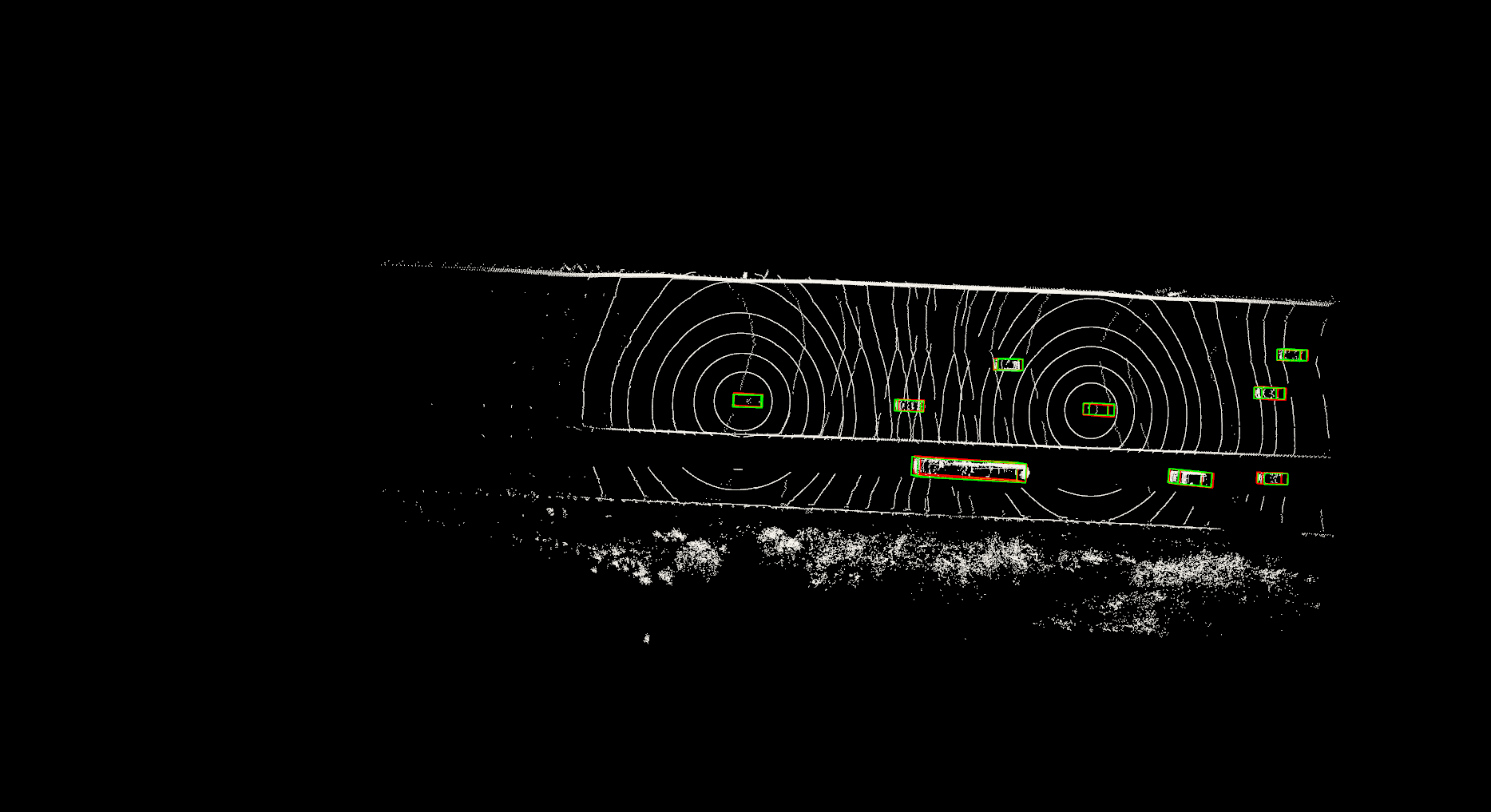}\\

\multirow[t]{1}{*}[\im_shift]{\begin{sideways}  V2X-ViT~\cite{xu2022v2xvit} \end{sideways}} &
 \includegraphics[ width=\xwidth\linewidth]{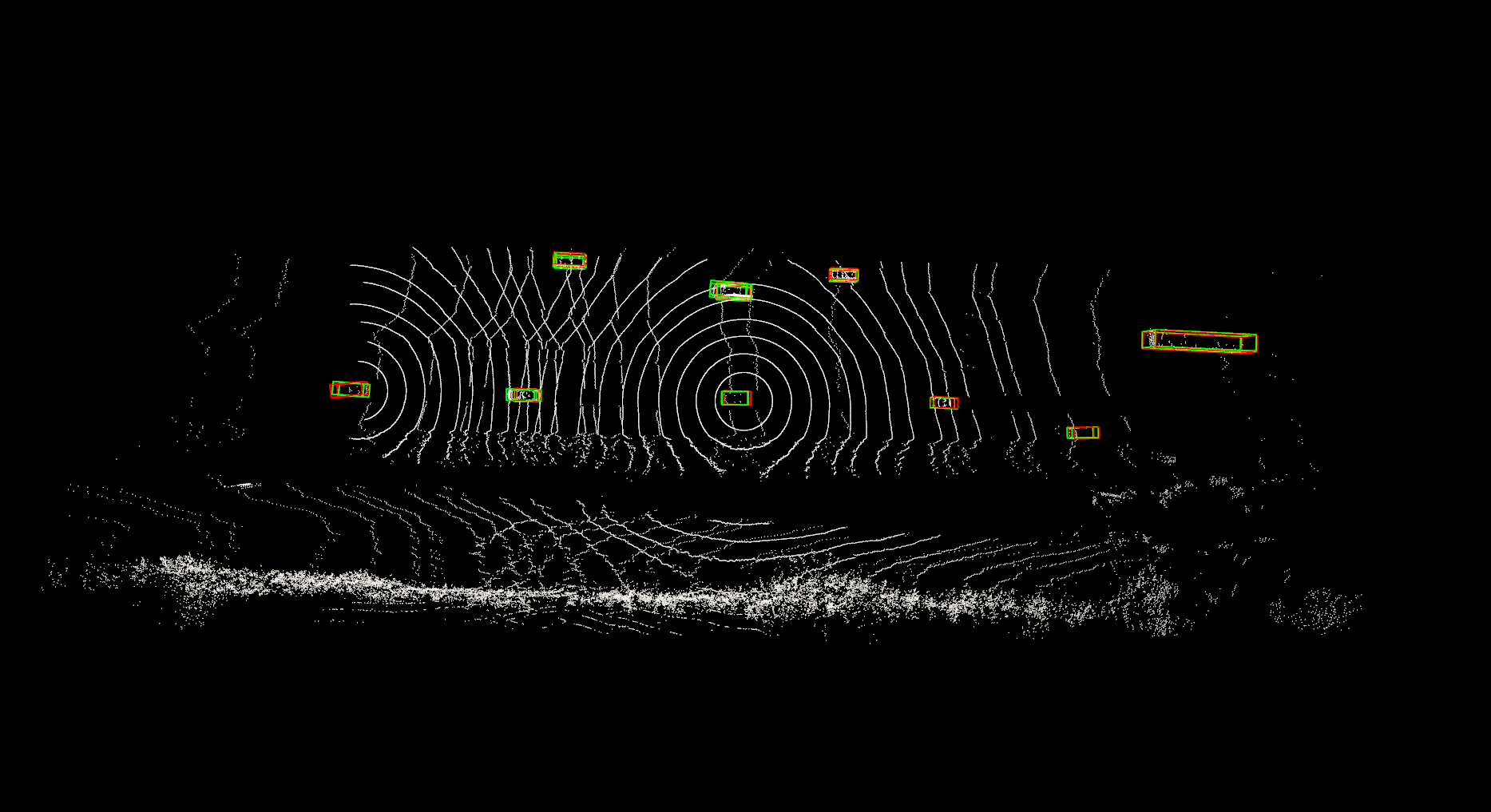}
& \includegraphics[ width=\xwidth\linewidth]{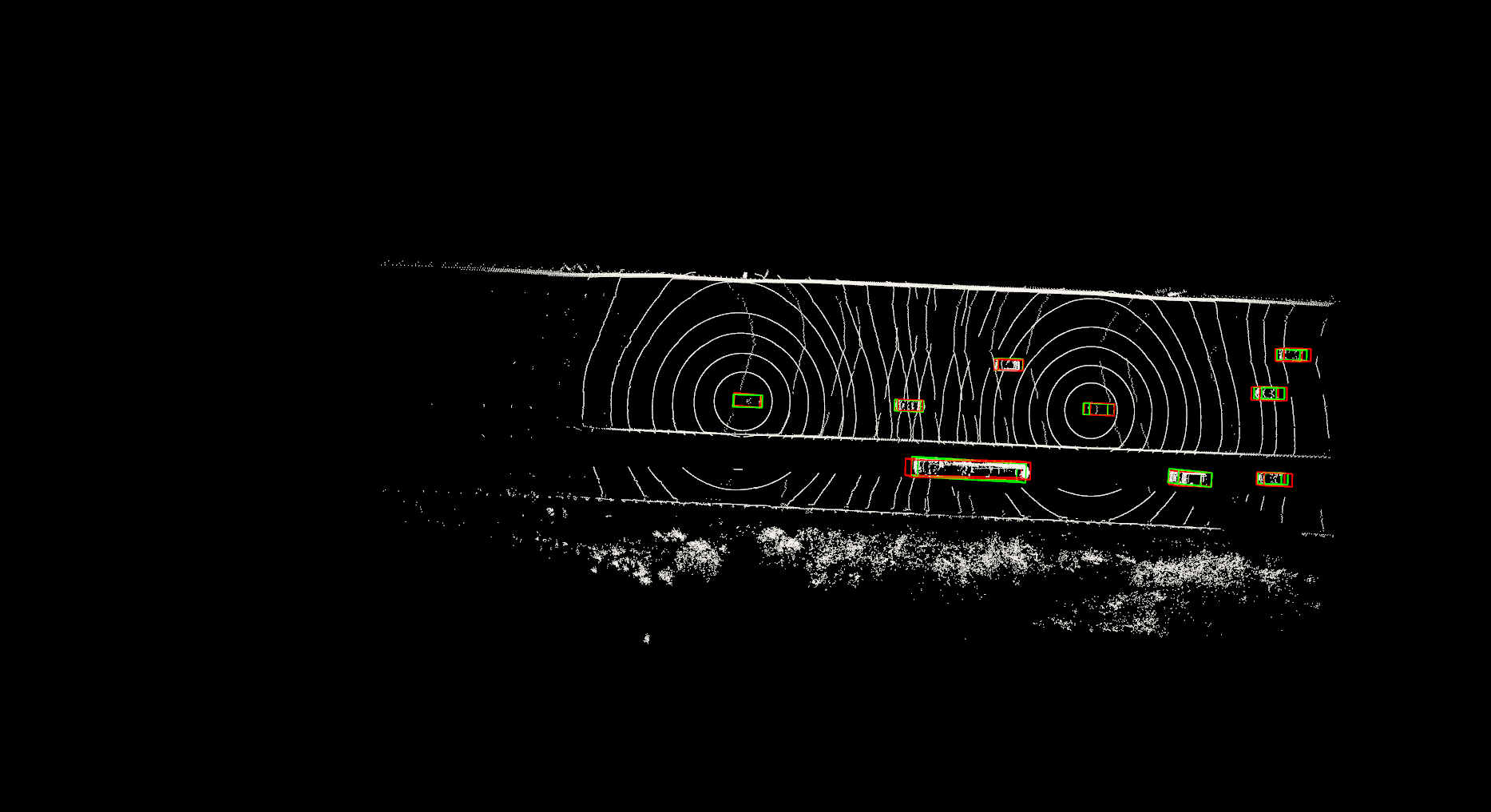}\\

\multirow[t]{1}{*}[\im_shift]{\begin{sideways}  CoBEVT~\cite{xu2022cobevt} \end{sideways}} &
 \includegraphics[width=\xwidth\linewidth]{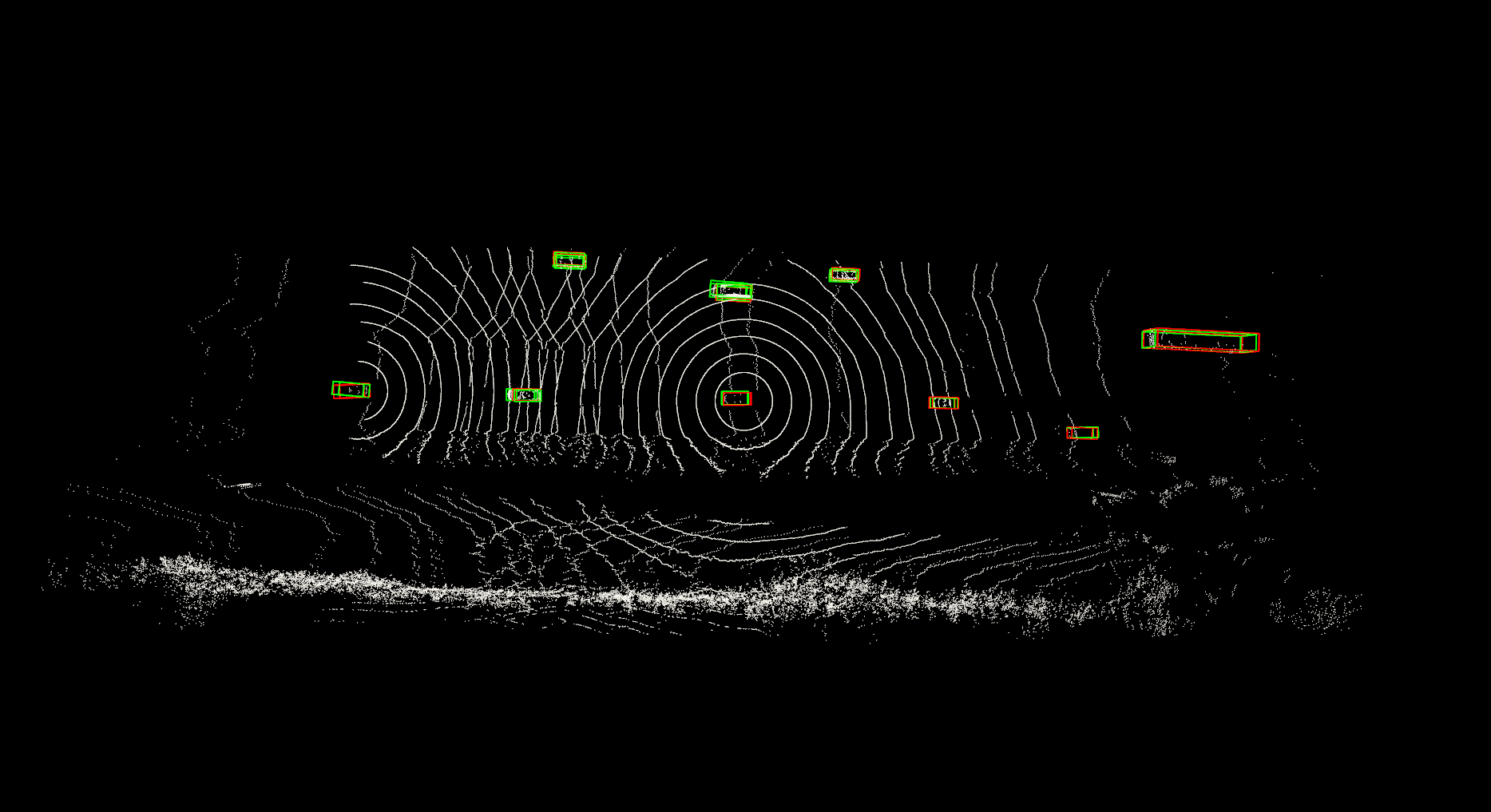}
 
& \includegraphics[width=\xwidth\linewidth]{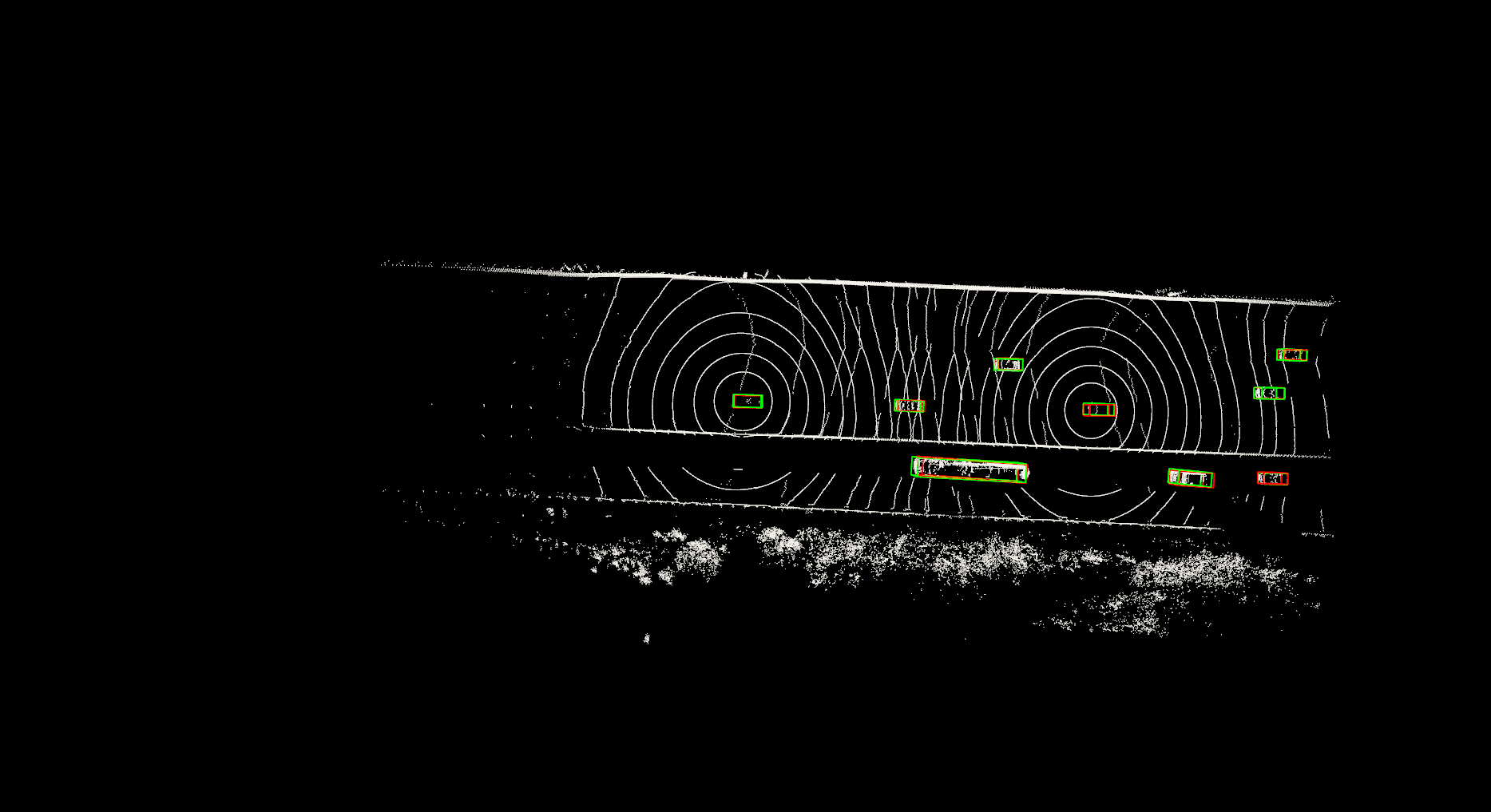}\\

\end{tabular}
\vspace{-3mm}
\caption{\textbf{Qualitative results of cooperative 3d object detection in two highway scenarios.} \textcolor{green}{Green} and \textcolor{red}{red} 3D bounding boxes represent the groundtruth and prediction, respectively.}
\label{fig:sup-qualitive4}
\end{figure*}

We demonstrate more qualitative results of the 3D detection comparisons in~\cref{fig:sup-qualitive3,fig:sup-qualitive4} under the \textit{Sync} setting. As shown in the urban scene in~\cref{fig:sup-qualitive3}, where traffic is heavier and crowded vehicles are causing severe occlusions, cooperative solutions yield significantly better detection results than No Fusion. Intermediate fusion methods also generate more accurate detection than early or late fusion within the medium radius ($<50$ m). It is obvious that among all the compared approaches, CoBEVT's prediction best aligns with the ground truth bounding boxes, which is consistent with the numerical results provided in the main paper. As for the highway scene in~\cref{fig:sup-qualitive4} where vehicles drive at higher speed but less crowdedness, all the cooperative methods can successfully predict the surrounding vehicles' bounding boxes, with some approaches (V2X-ViT, CoBEVT) slightly more accurate than others (V2VNet, Late Fusion, Earl Fusion).

\section{Domain Adaptation Results}
\label{sec:sup-da}

\begin{figure*}[!ht]
\centering
\footnotesize
\def\xwidth{0.42}
\def\yheight{0.18}
\def\xem{-2pt}
\def\im_shift{0.1\textwidth}
\setlength{\tabcolsep}{0.5pt}
\begin{tabular}{cccc}
 & Without Domain Adaption & With Domain Adaption\\
 \multirow[t]{1}{*}[\im_shift]{\begin{sideways} AttFuse~\cite{xu2022opv2v} \end{sideways}} &
\includegraphics[ width=\xwidth\linewidth]{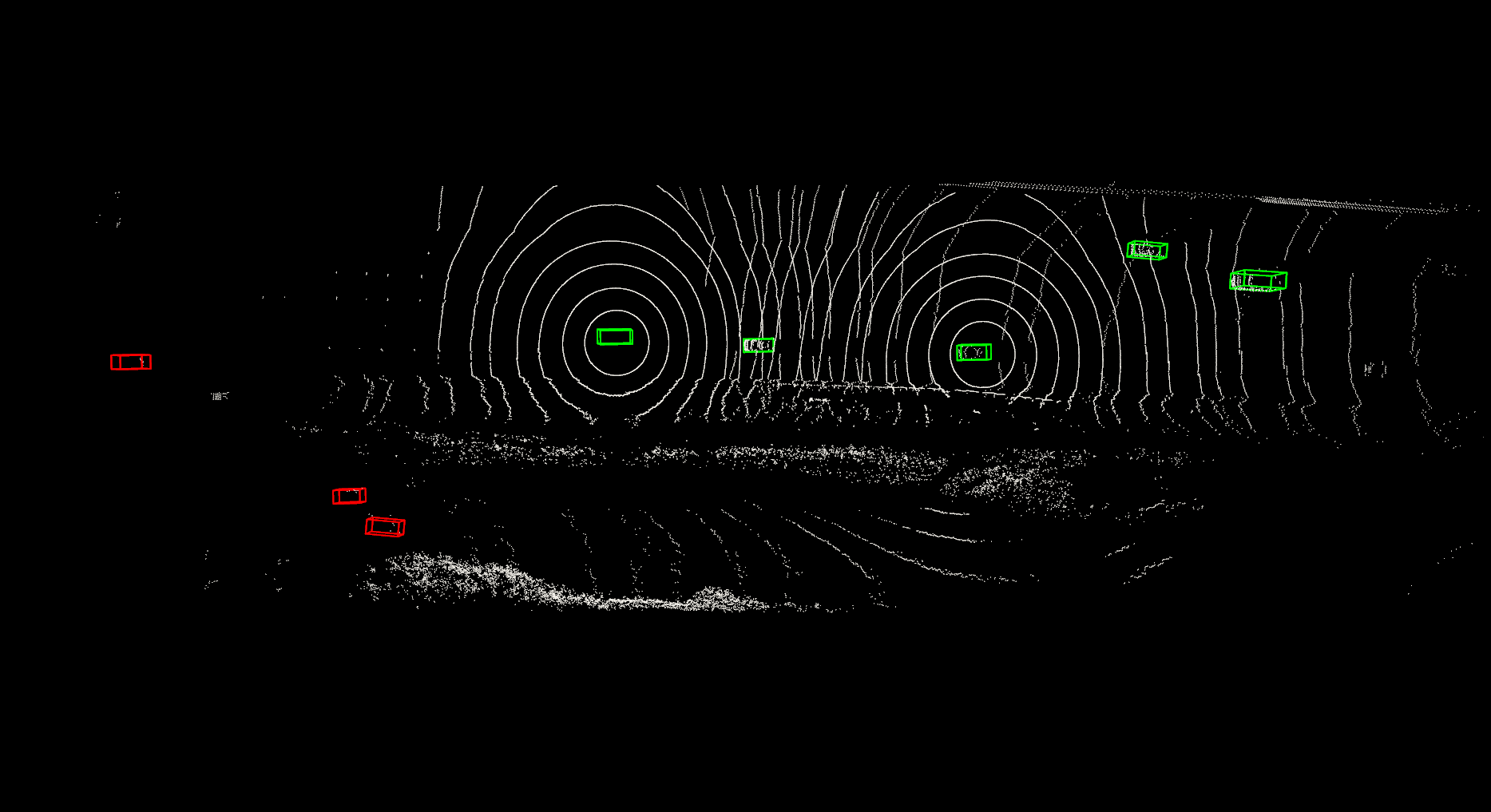}
& \includegraphics[ width=\xwidth\linewidth]{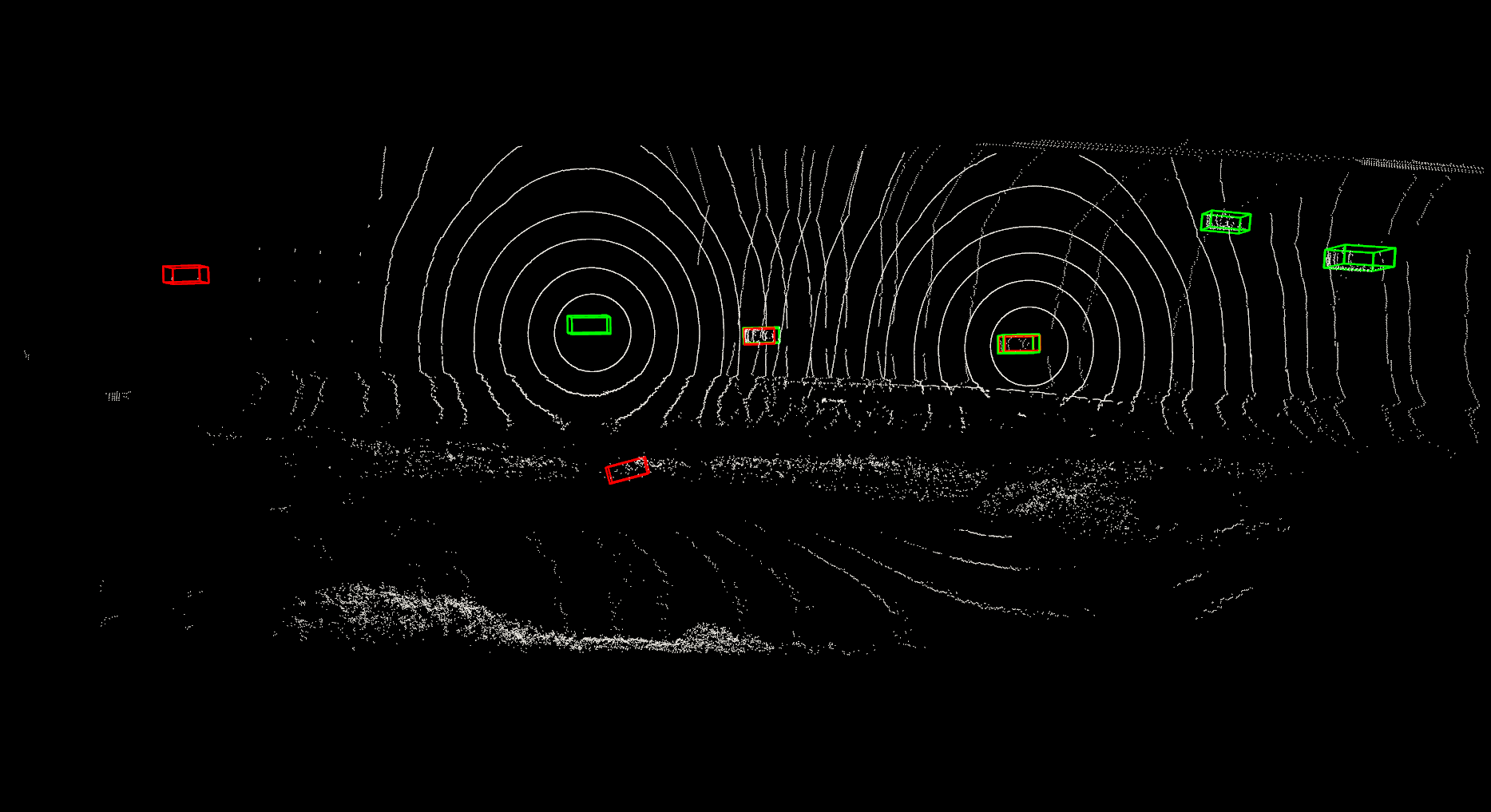}\\

\multirow[t]{1}{*}[\im_shift]{\begin{sideways} V2VNet~\cite{wang2020v2vnet} \end{sideways}}  &
 \includegraphics[ width=\xwidth\linewidth]{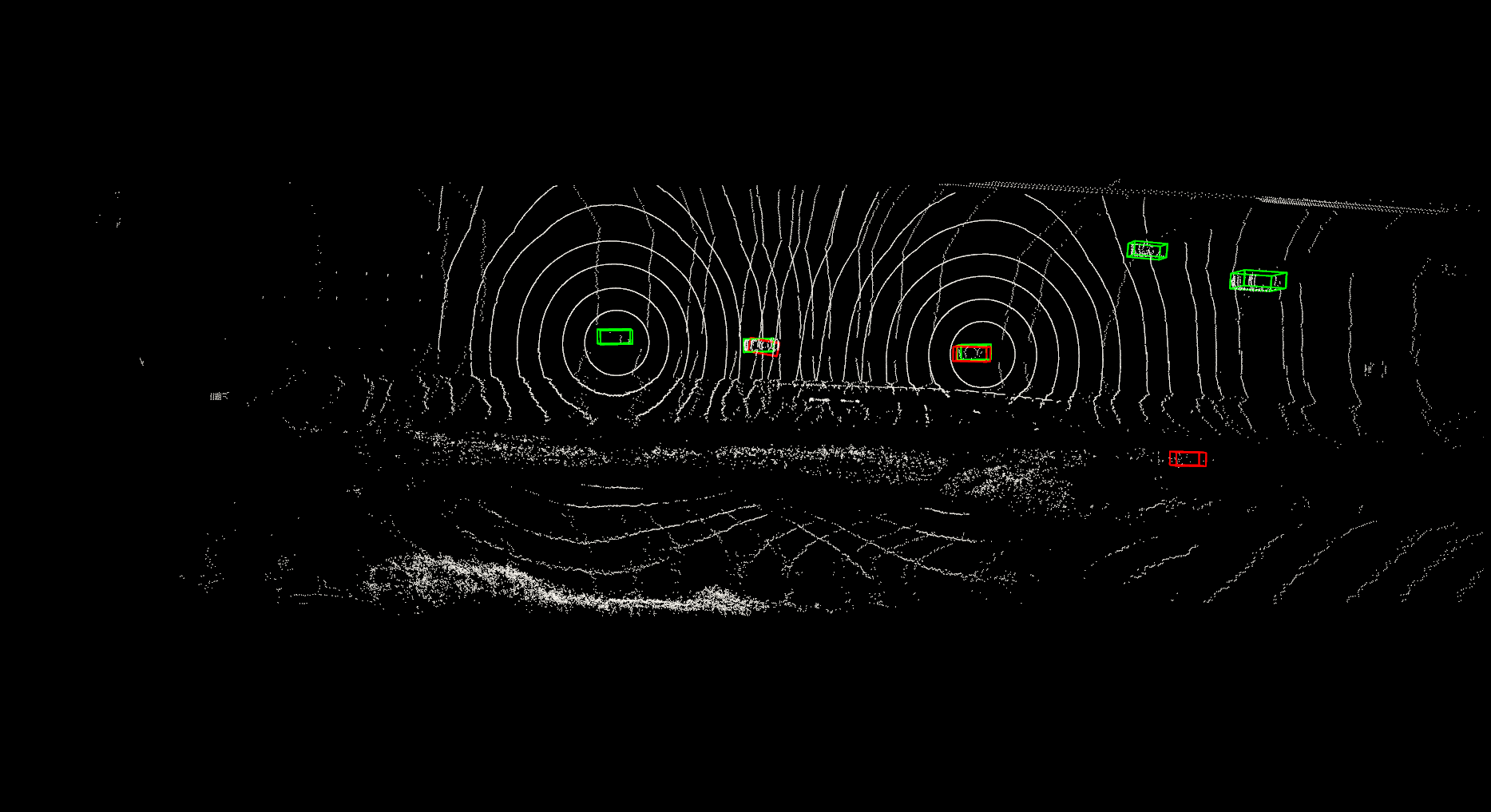}
& \includegraphics[ width=\xwidth\linewidth]{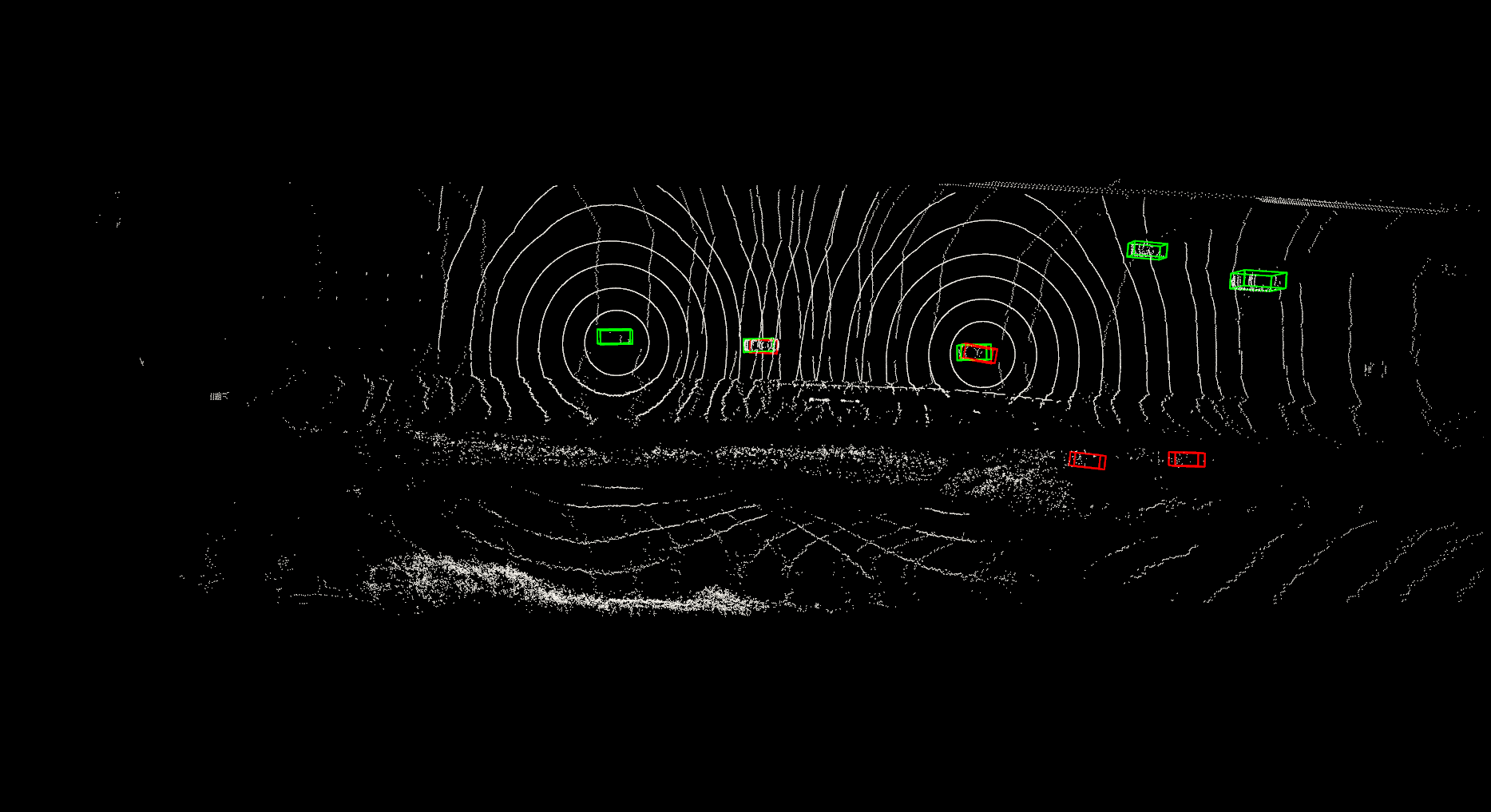}\\

\multirow[t]{1}{*}[\im_shift]{\begin{sideways}  F-Cooper~\cite{chen2019f}  \end{sideways}}  &
 \includegraphics[ width=\xwidth\linewidth]{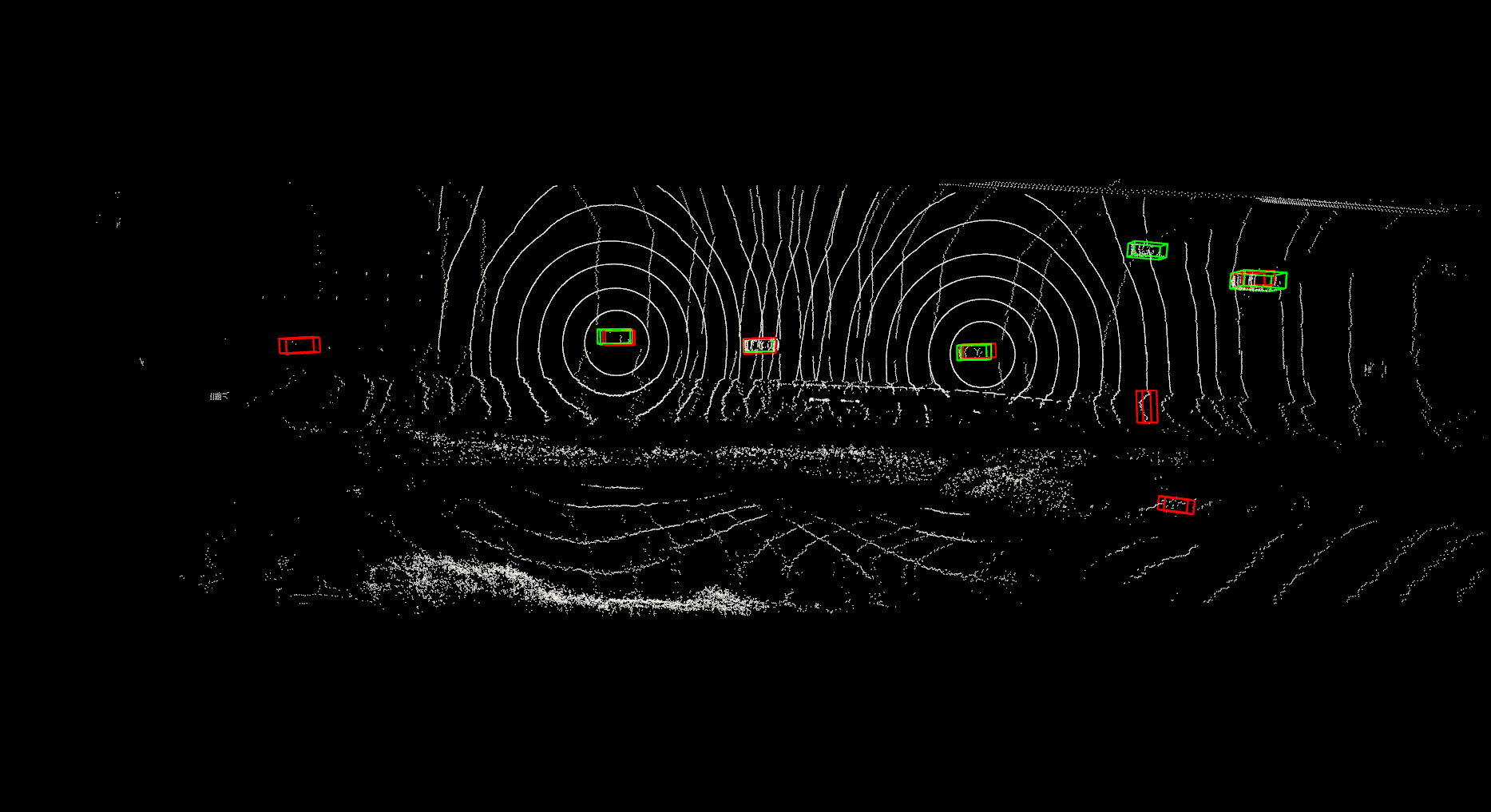}
& \includegraphics[ width=\xwidth\linewidth]{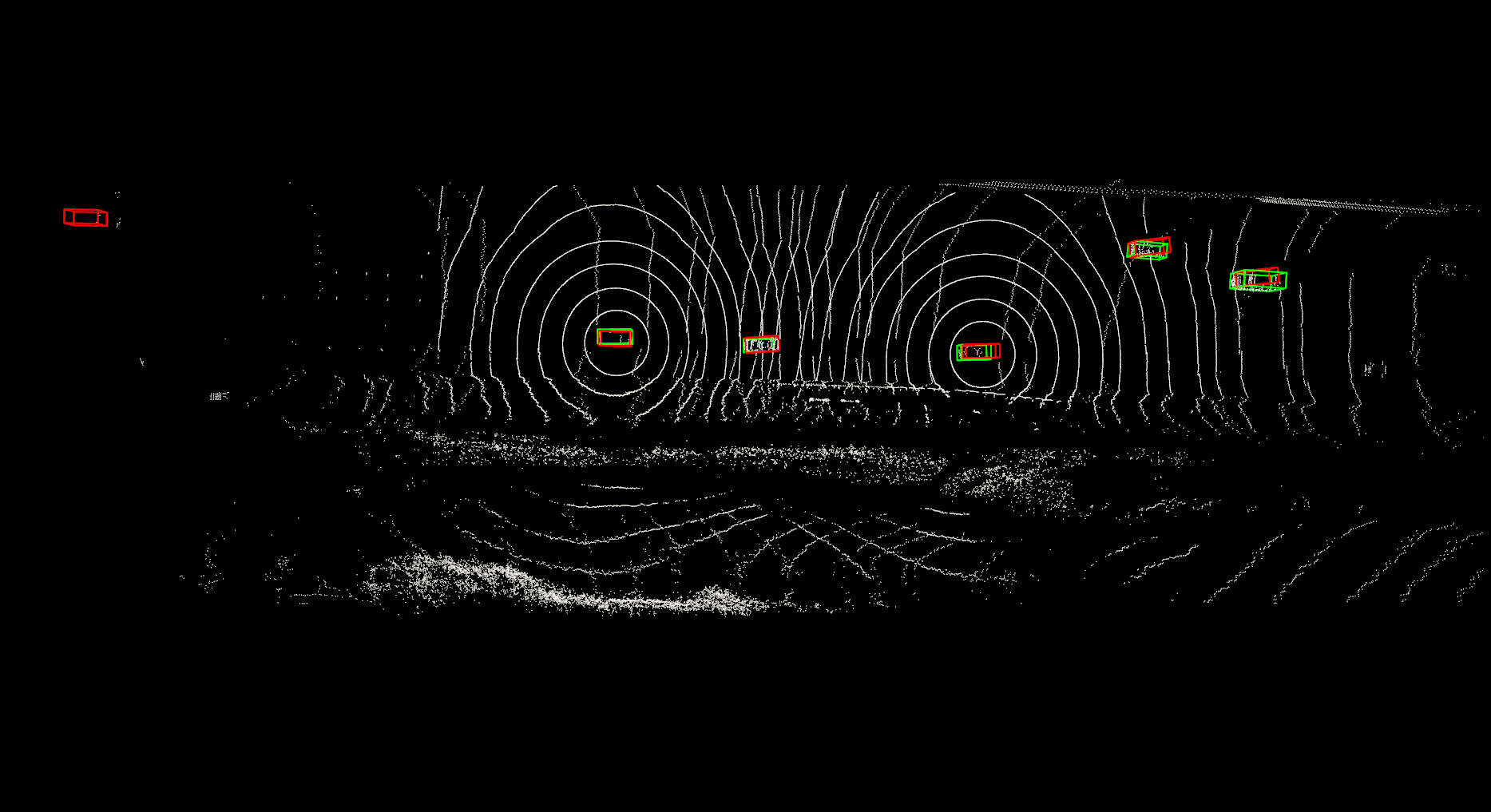}\\

\multirow[t]{1}{*}[\im_shift]{\begin{sideways}  V2X-ViT~\cite{xu2022v2xvit} \end{sideways}} &
 \includegraphics[ width=\xwidth\linewidth]{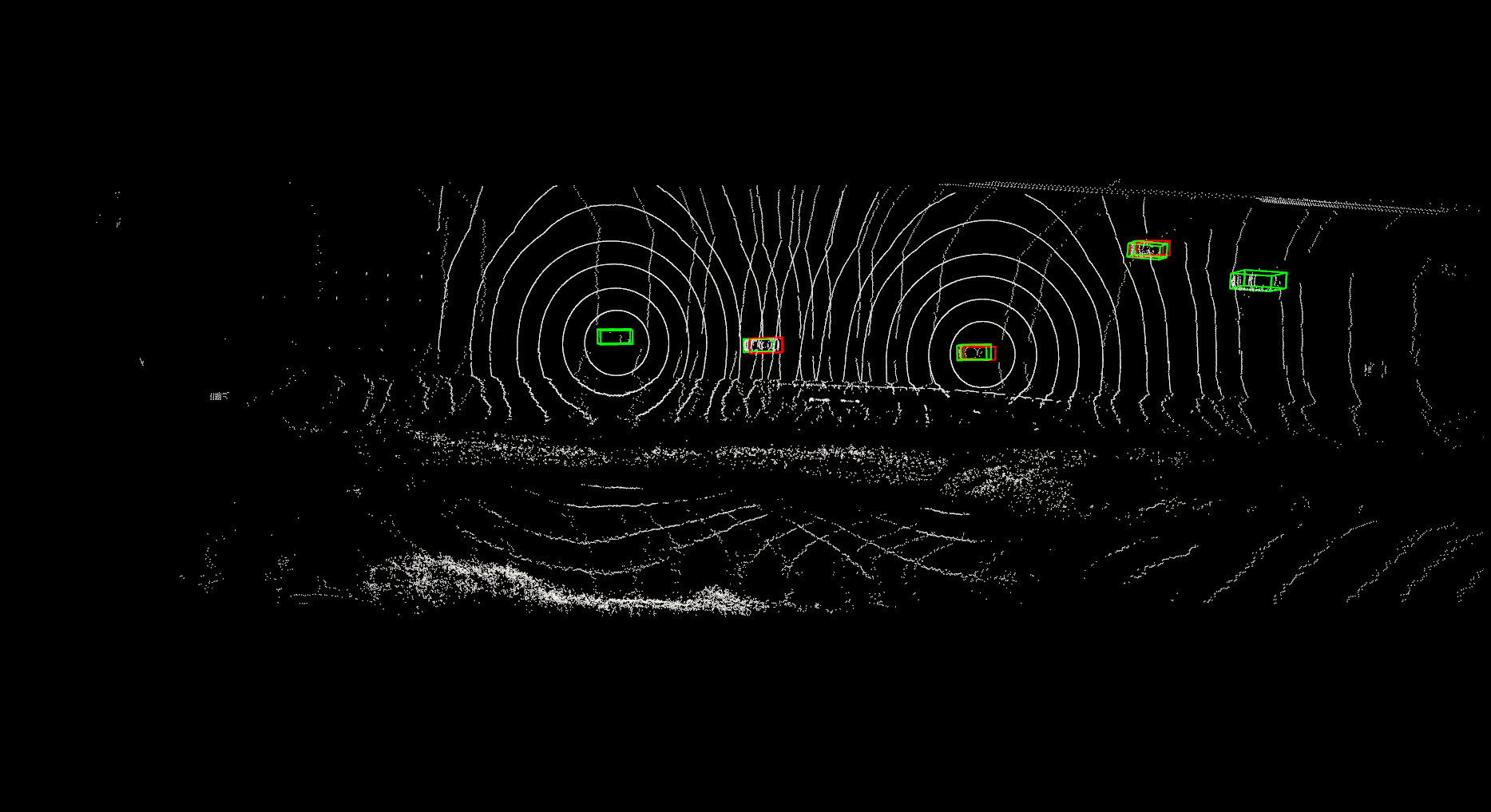}
& \includegraphics[ width=\xwidth\linewidth]{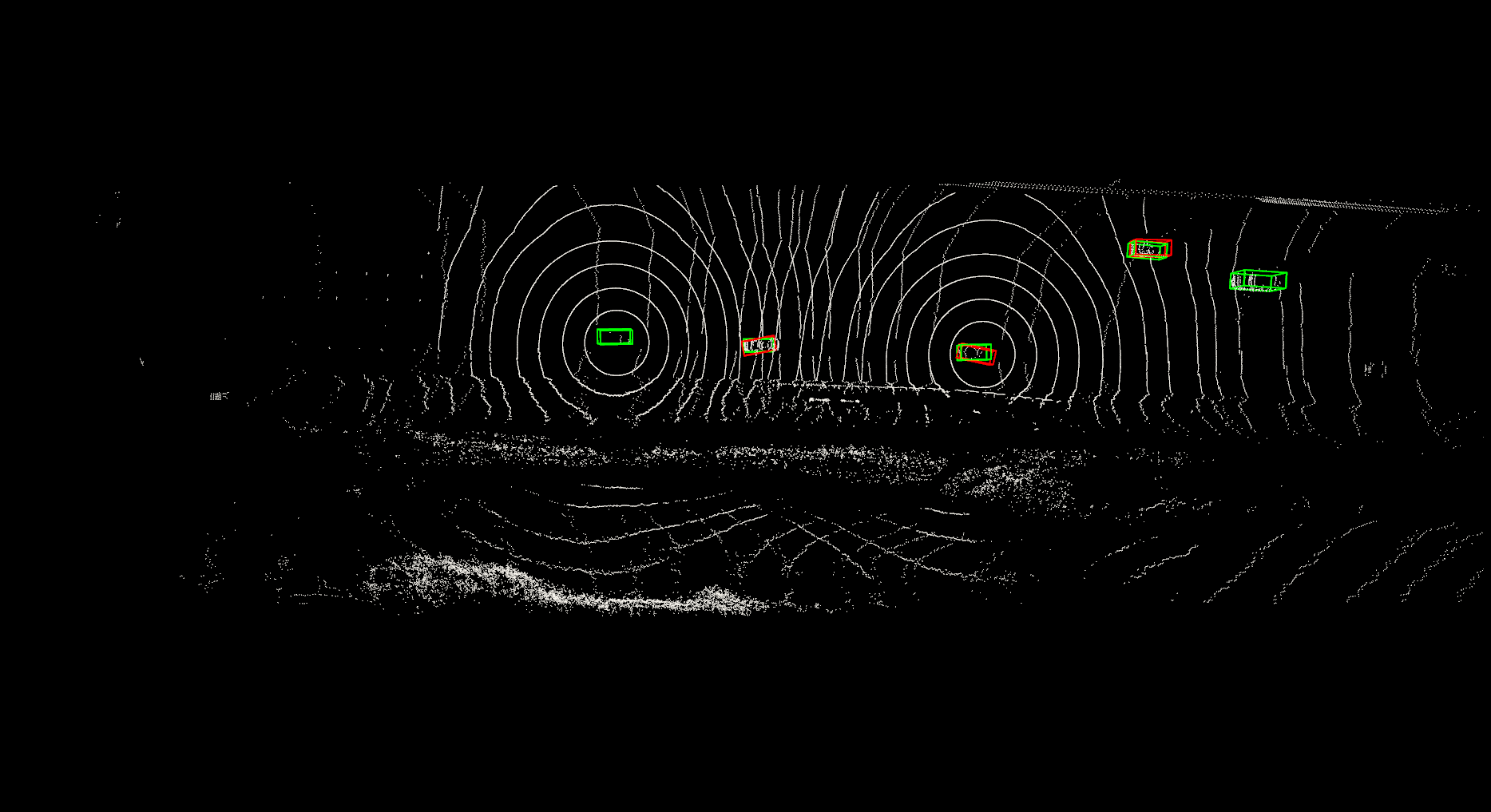}\\

\multirow[t]{1}{*}[\im_shift]{\begin{sideways}  CoBEVT~\cite{xu2022cobevt} \end{sideways}} &
 \includegraphics[ width=\xwidth\linewidth]{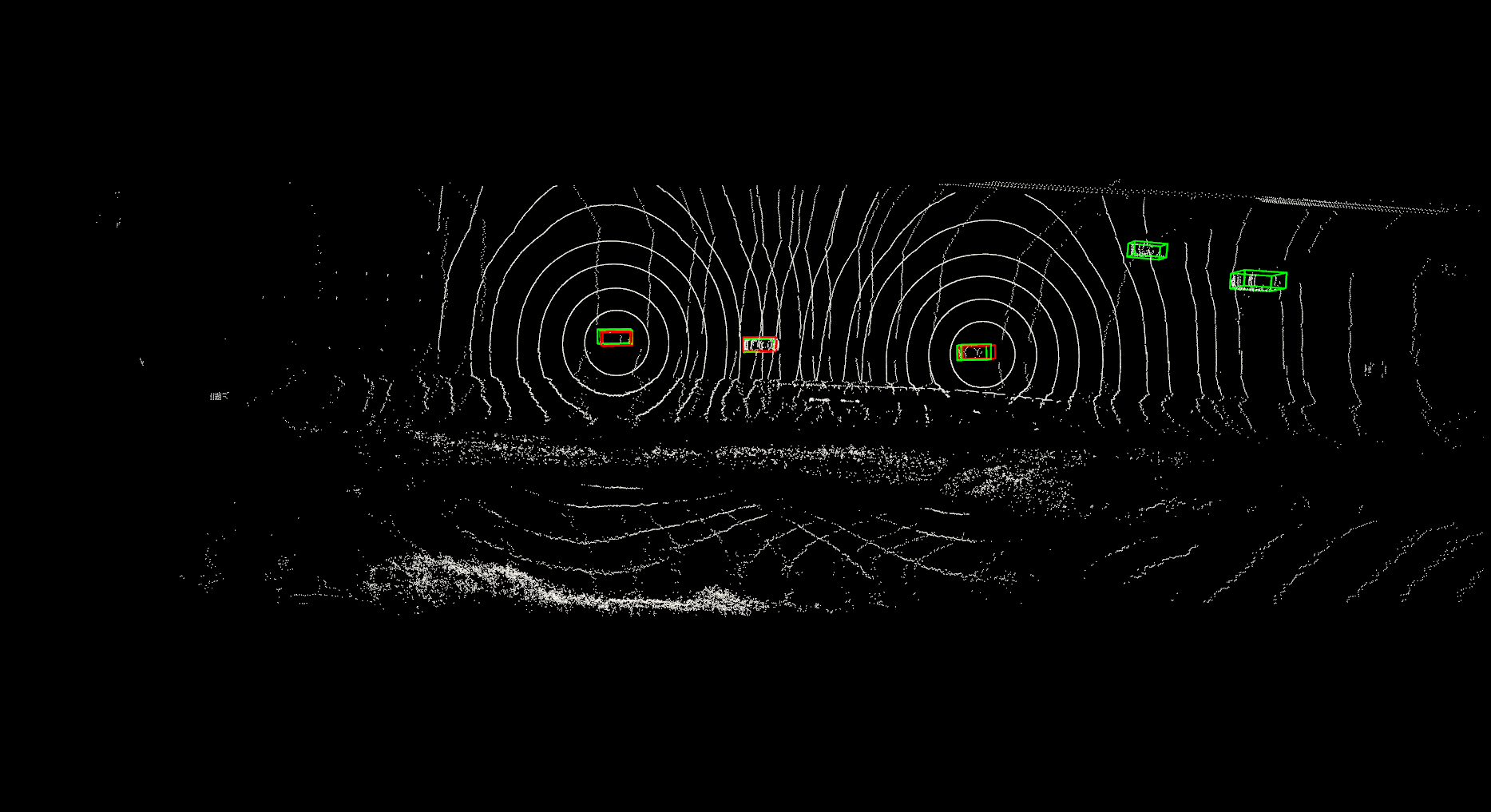}
& \includegraphics[ width=\xwidth\linewidth]{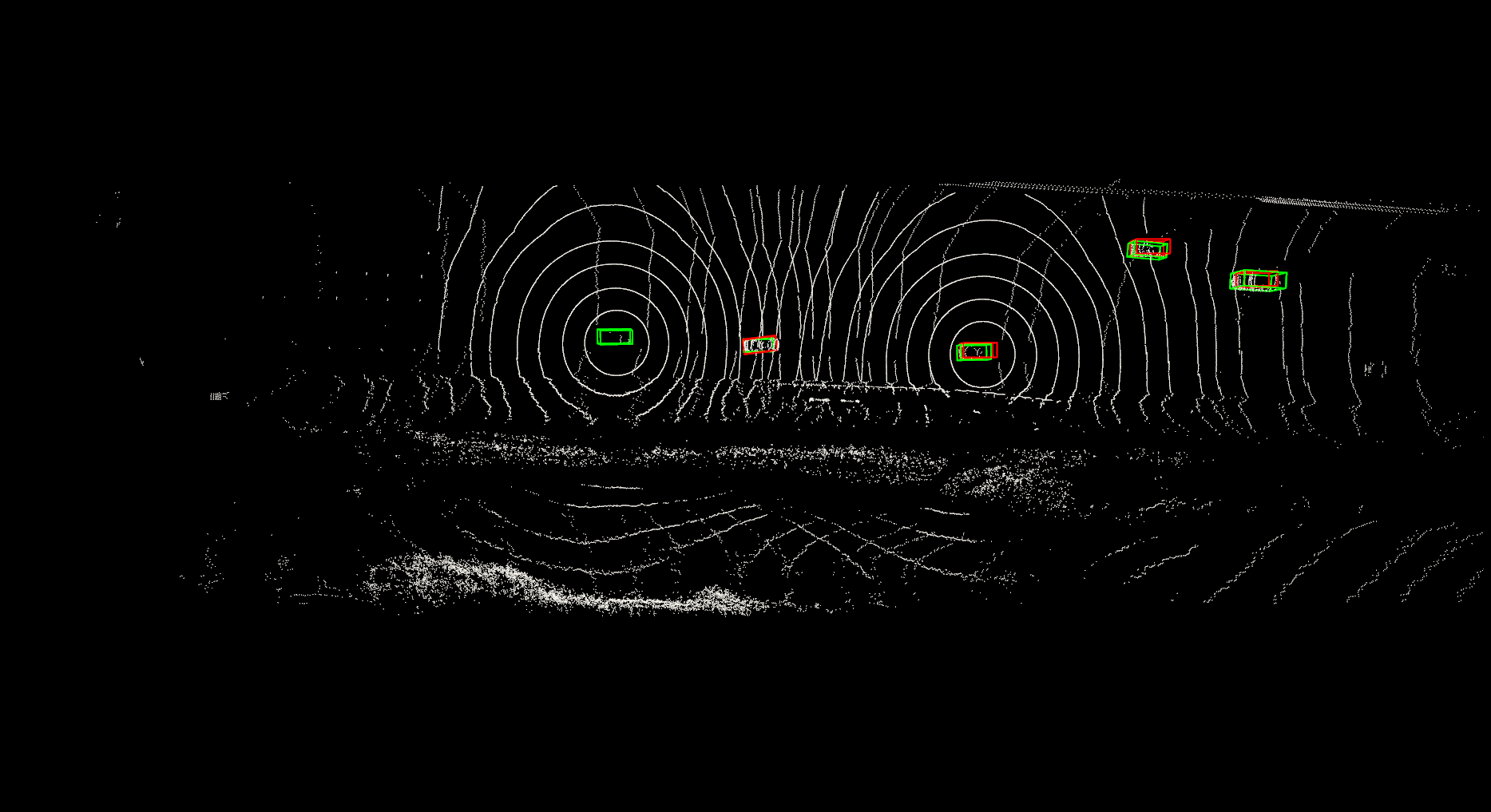}\\

\end{tabular}
\vspace{-3mm}
\caption{\textbf{Qualitative results of domain adaption in a highway scenario.} \textcolor{green}{Green} and \textcolor{red}{red} 3D bounding boxes represent the groundtruth and prediction, respectively.}
\label{fig:sup-da1}
\end{figure*}

\begin{figure*}[!ht]
\centering
\footnotesize
\def\xwidth{0.42}
\def\yheight{0.18}
\def\xem{-2pt}
\def\im_shift{0.1\textwidth}
\setlength{\tabcolsep}{0.5pt}
\begin{tabular}{cccc}
 & Without Domain Adaption & With Domain Adaption\\
 \multirow[t]{1}{*}[\im_shift]{\begin{sideways} AttFuse~\cite{xu2022opv2v} \end{sideways}} &
\includegraphics[ width=\xwidth\linewidth]{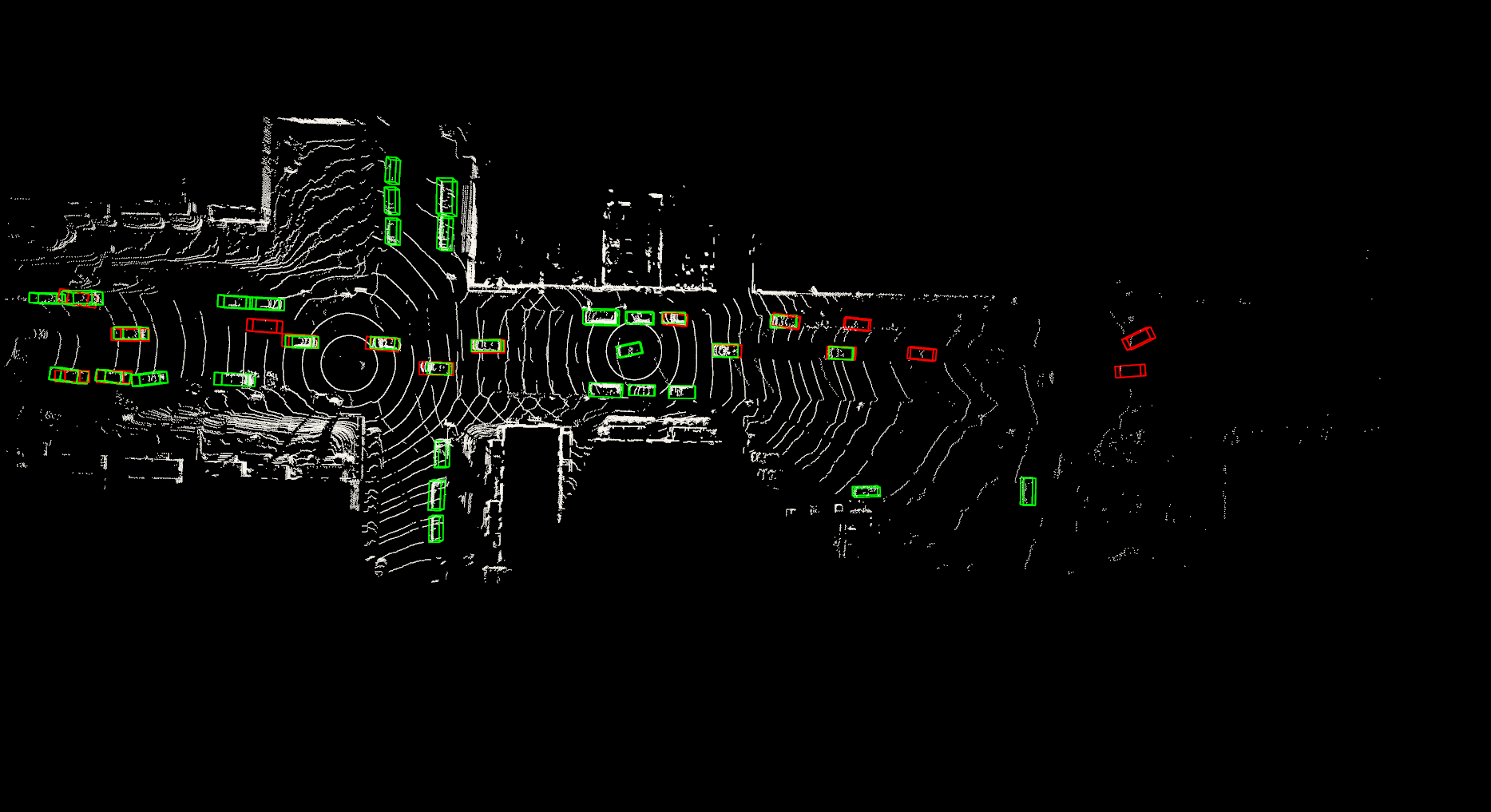}
& \includegraphics[ width=\xwidth\linewidth]{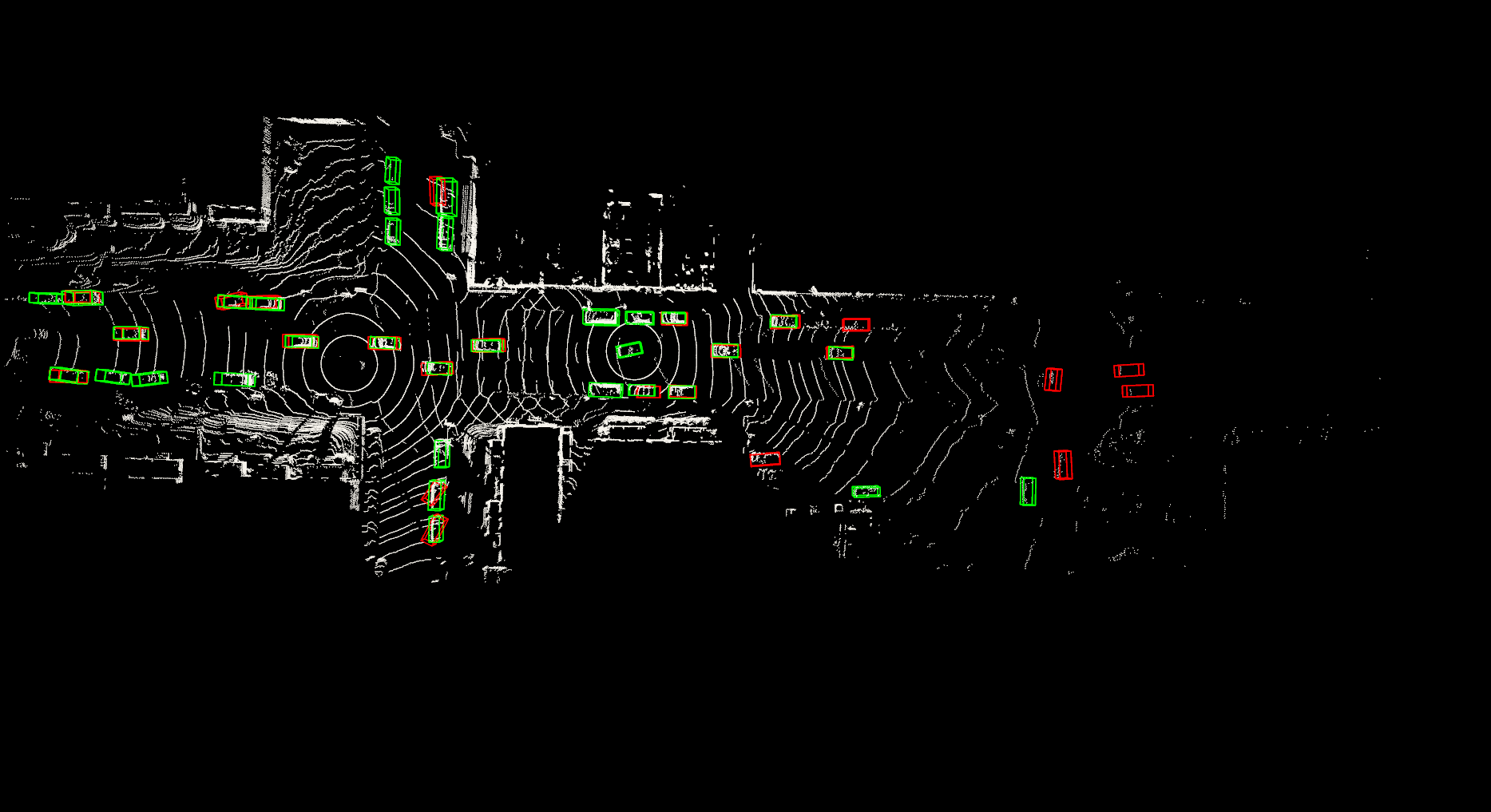}\\

\multirow[t]{1}{*}[\im_shift]{\begin{sideways} V2VNet~\cite{wang2020v2vnet} \end{sideways}}  &
 \includegraphics[ width=\xwidth\linewidth]{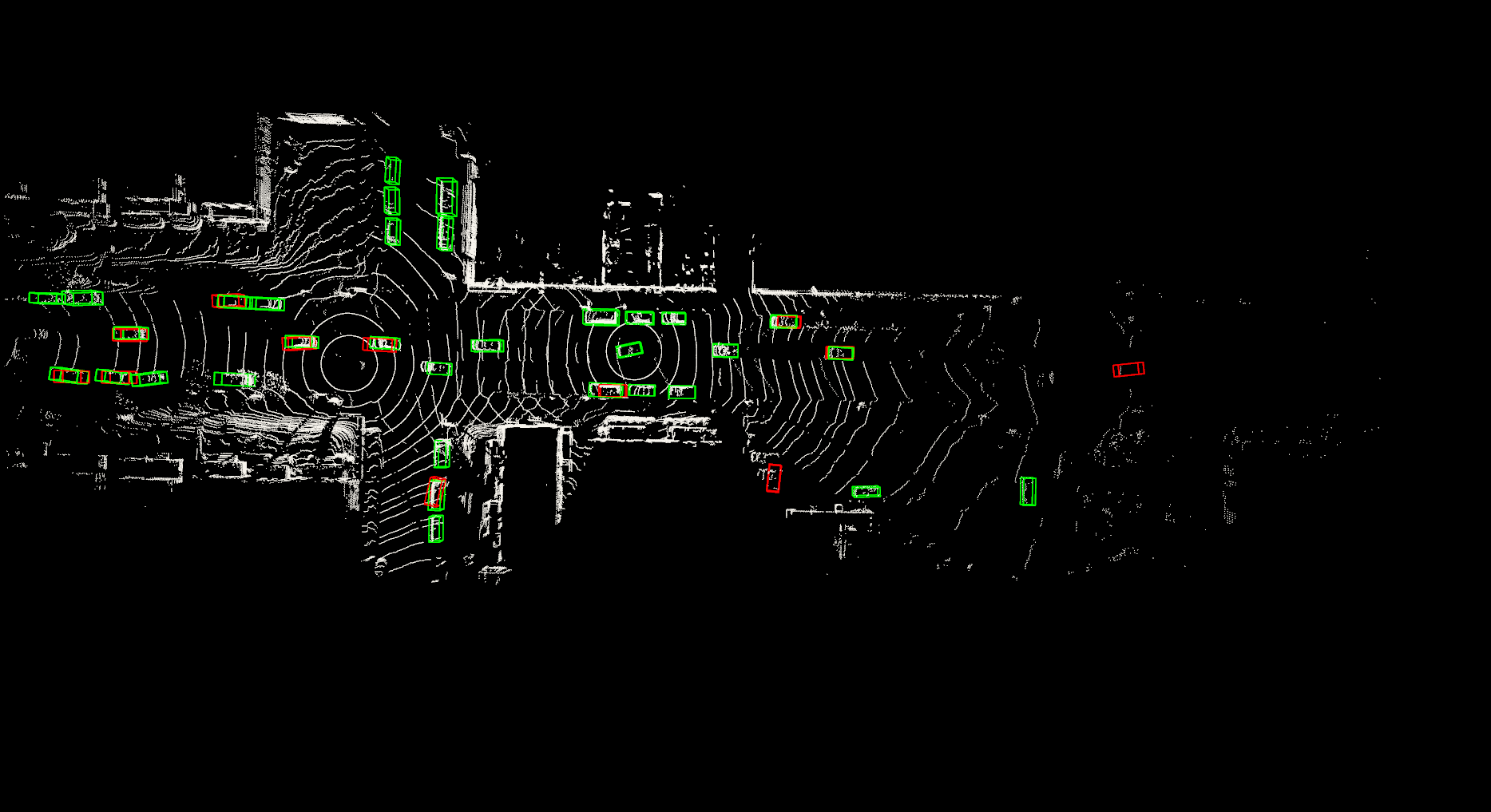}
& \includegraphics[ width=\xwidth\linewidth]{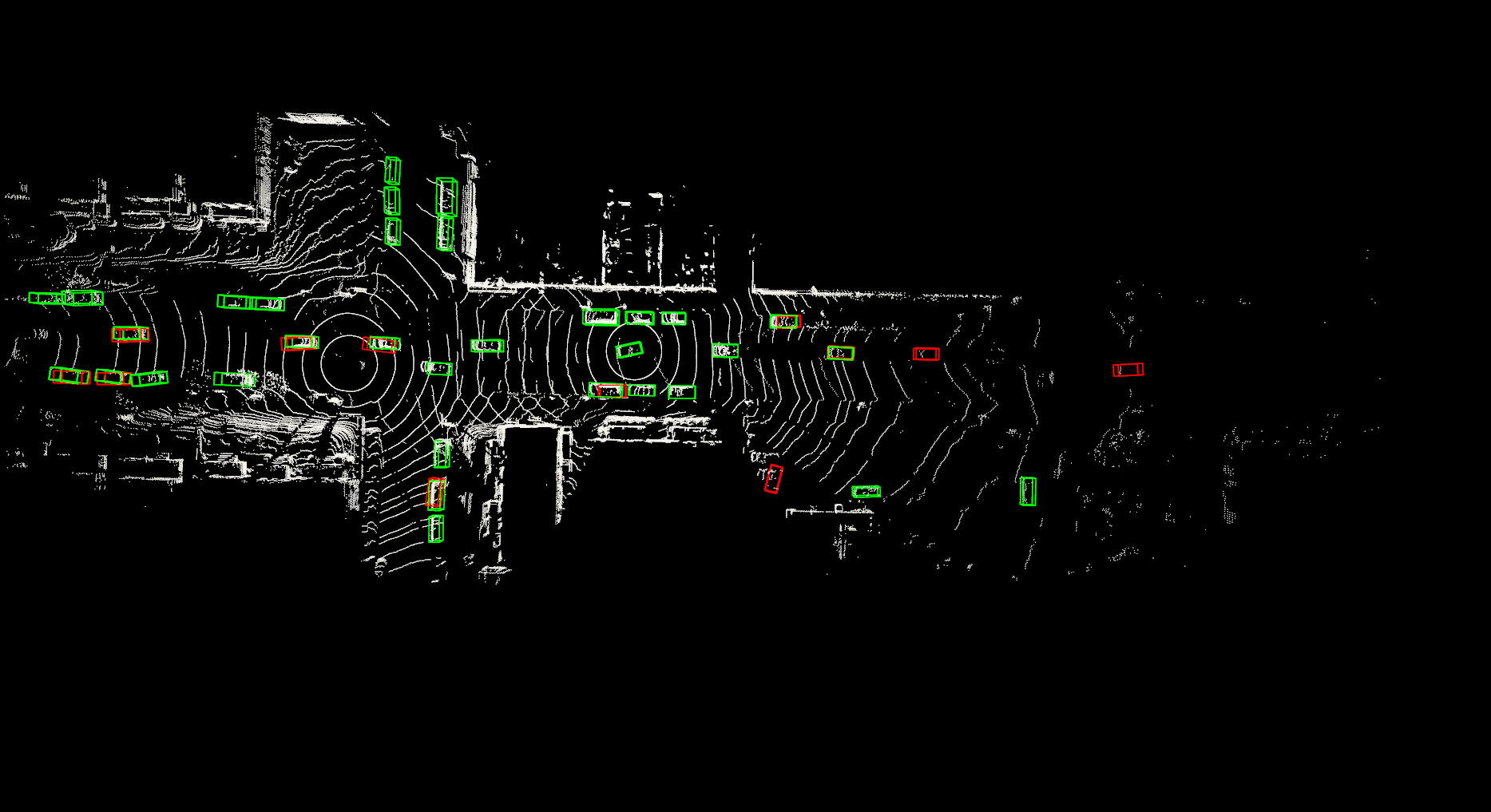}\\

\multirow[t]{1}{*}[\im_shift]{\begin{sideways}  F-Cooper~\cite{chen2019f}  \end{sideways}}  &
 \includegraphics[ width=\xwidth\linewidth]{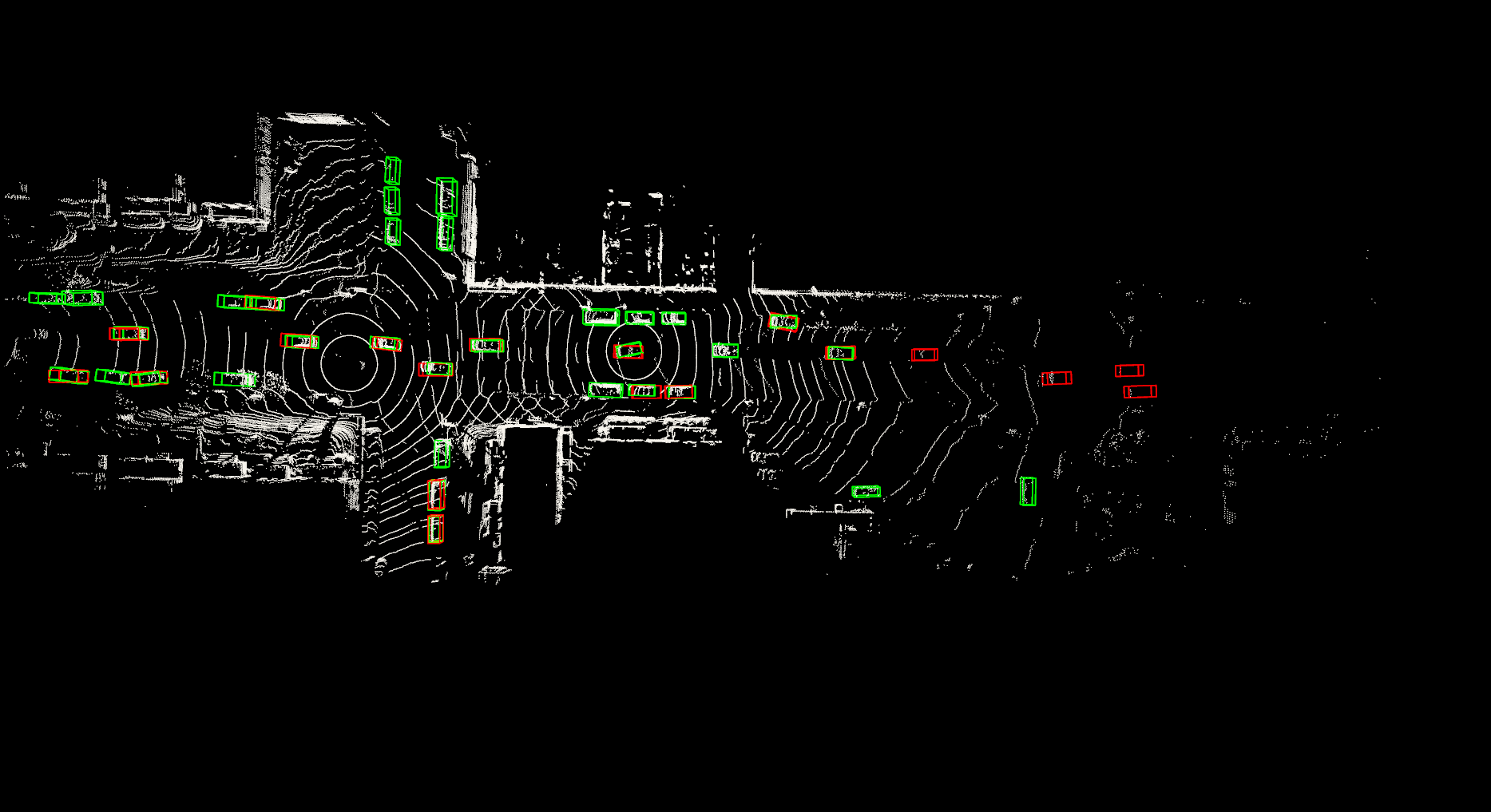}
& \includegraphics[ width=\xwidth\linewidth]{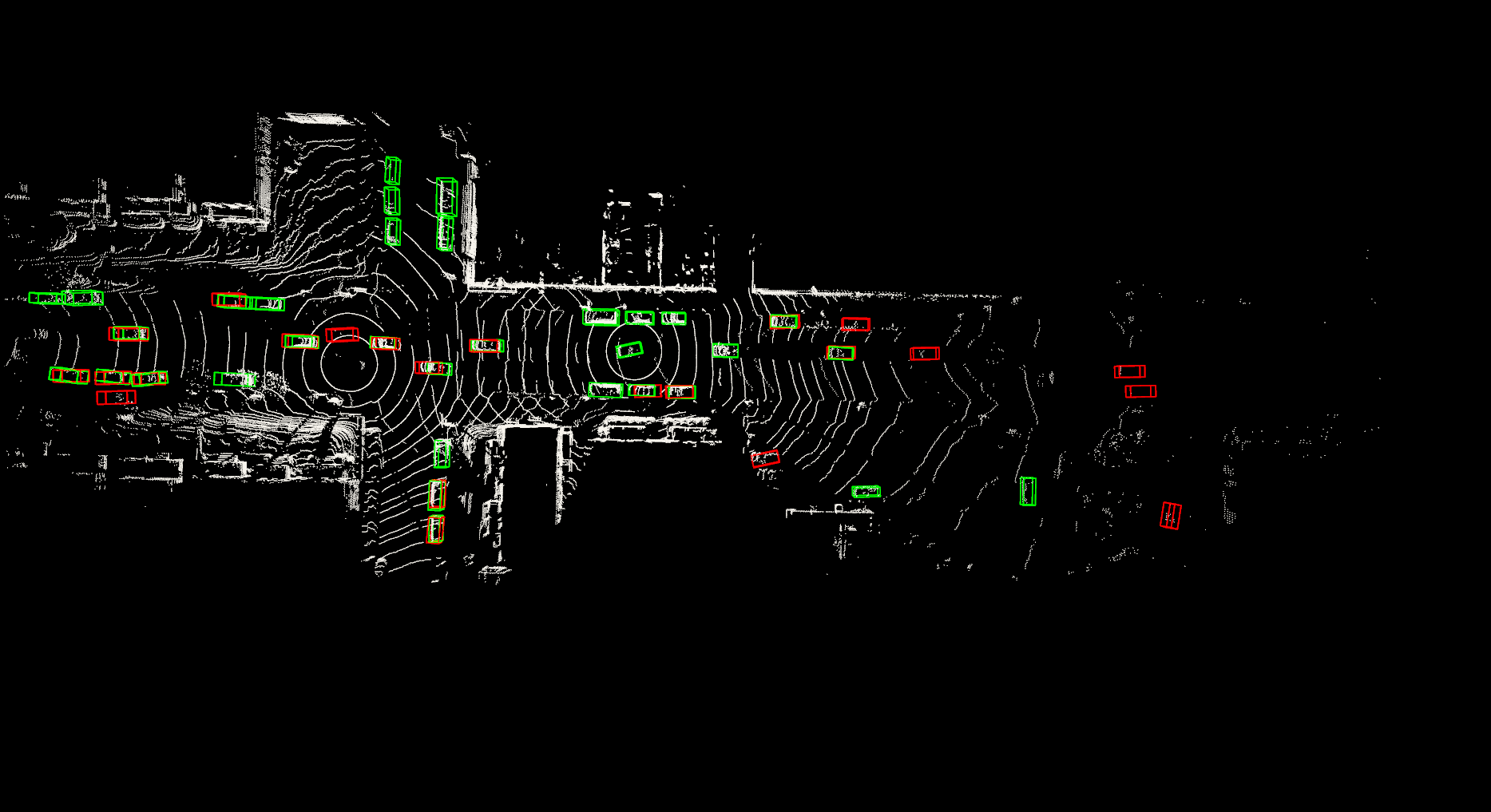}\\

\multirow[t]{1}{*}[\im_shift]{\begin{sideways}  V2X-ViT~\cite{xu2022v2xvit} \end{sideways}} &
 \includegraphics[ width=\xwidth\linewidth]{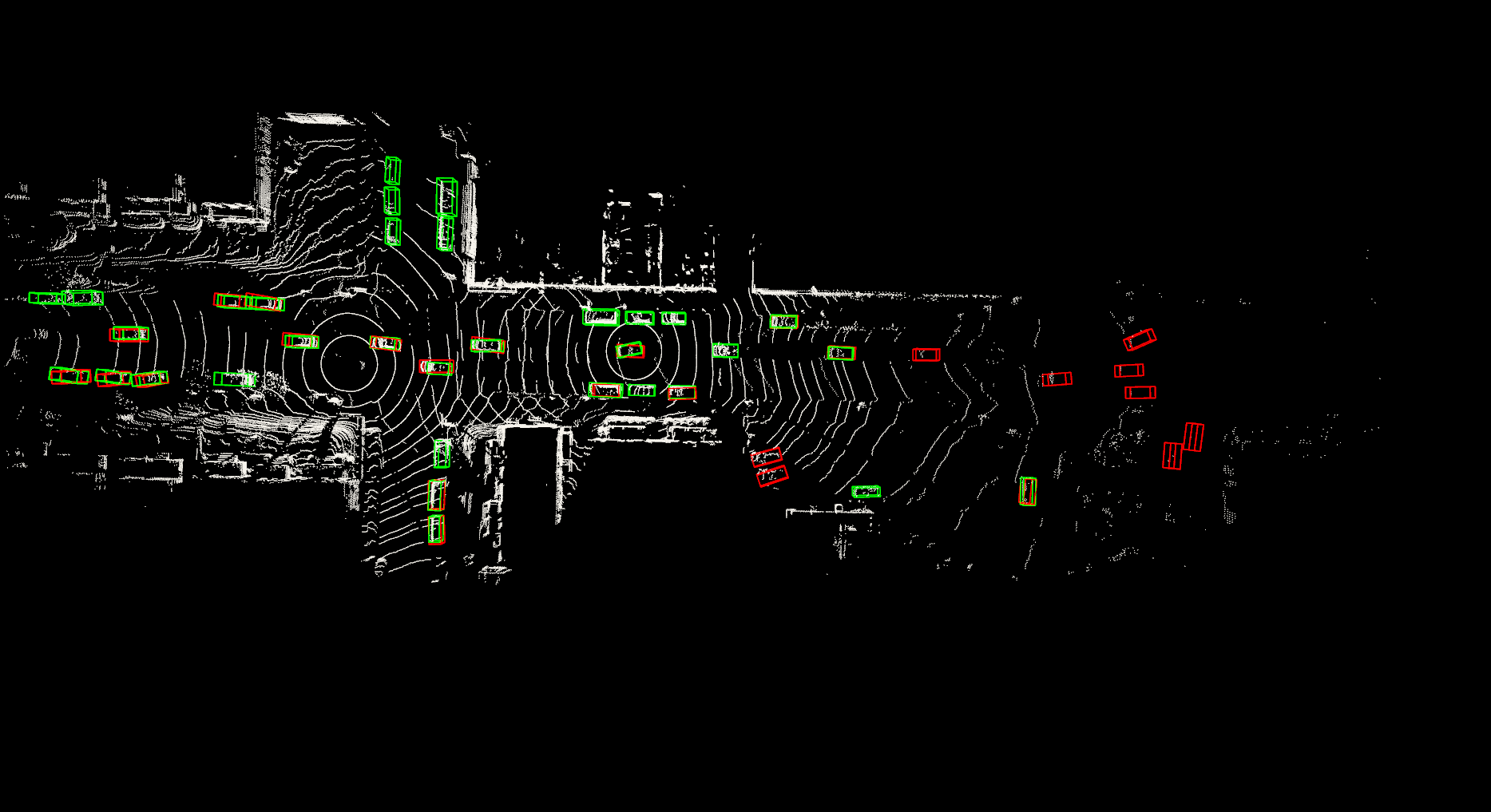}
& \includegraphics[ width=\xwidth\linewidth]{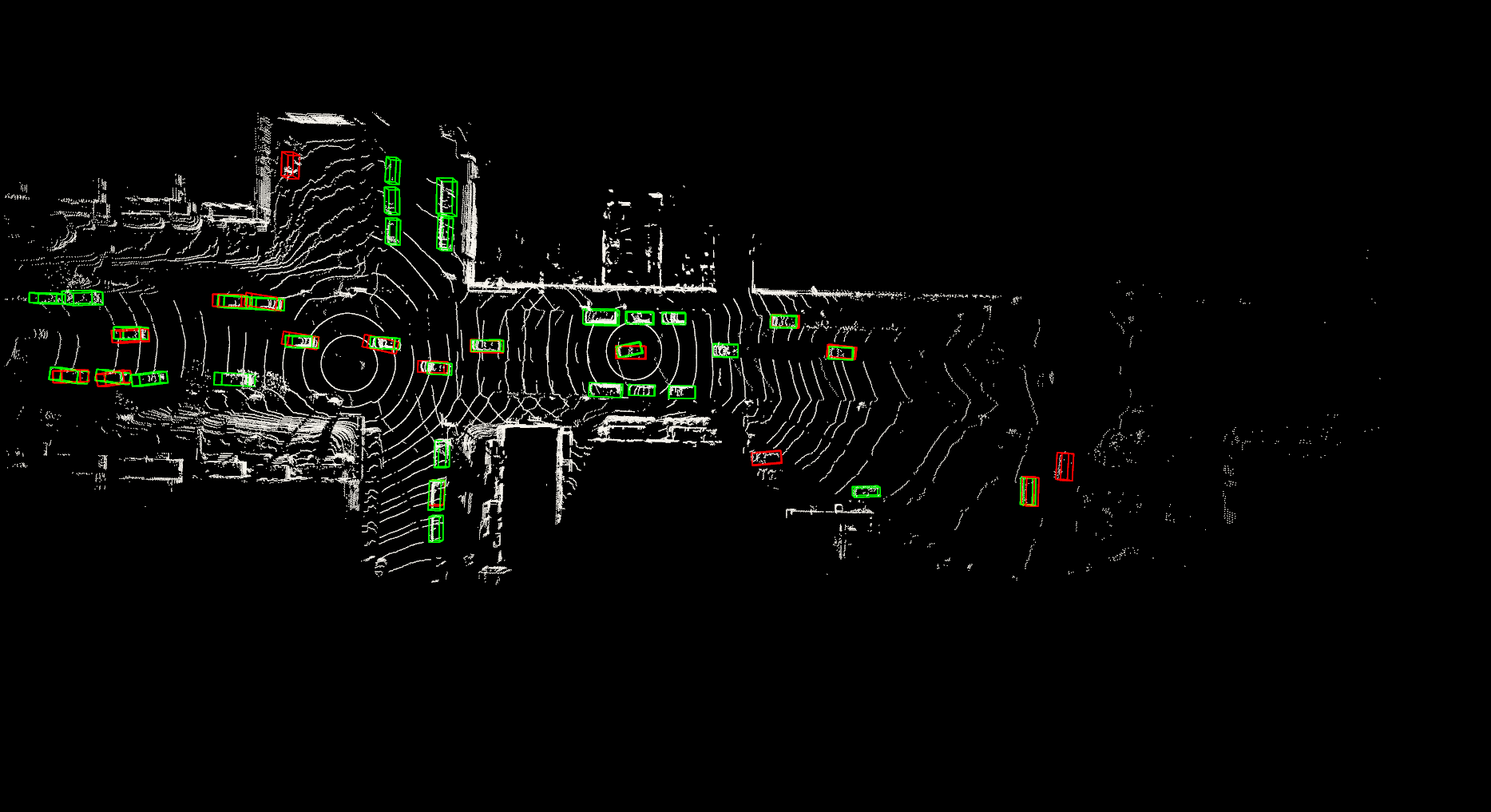}\\

\multirow[t]{1}{*}[\im_shift]{\begin{sideways}  CoBEVT~\cite{xu2022cobevt} \end{sideways}} &
 \includegraphics[ width=\xwidth\linewidth]{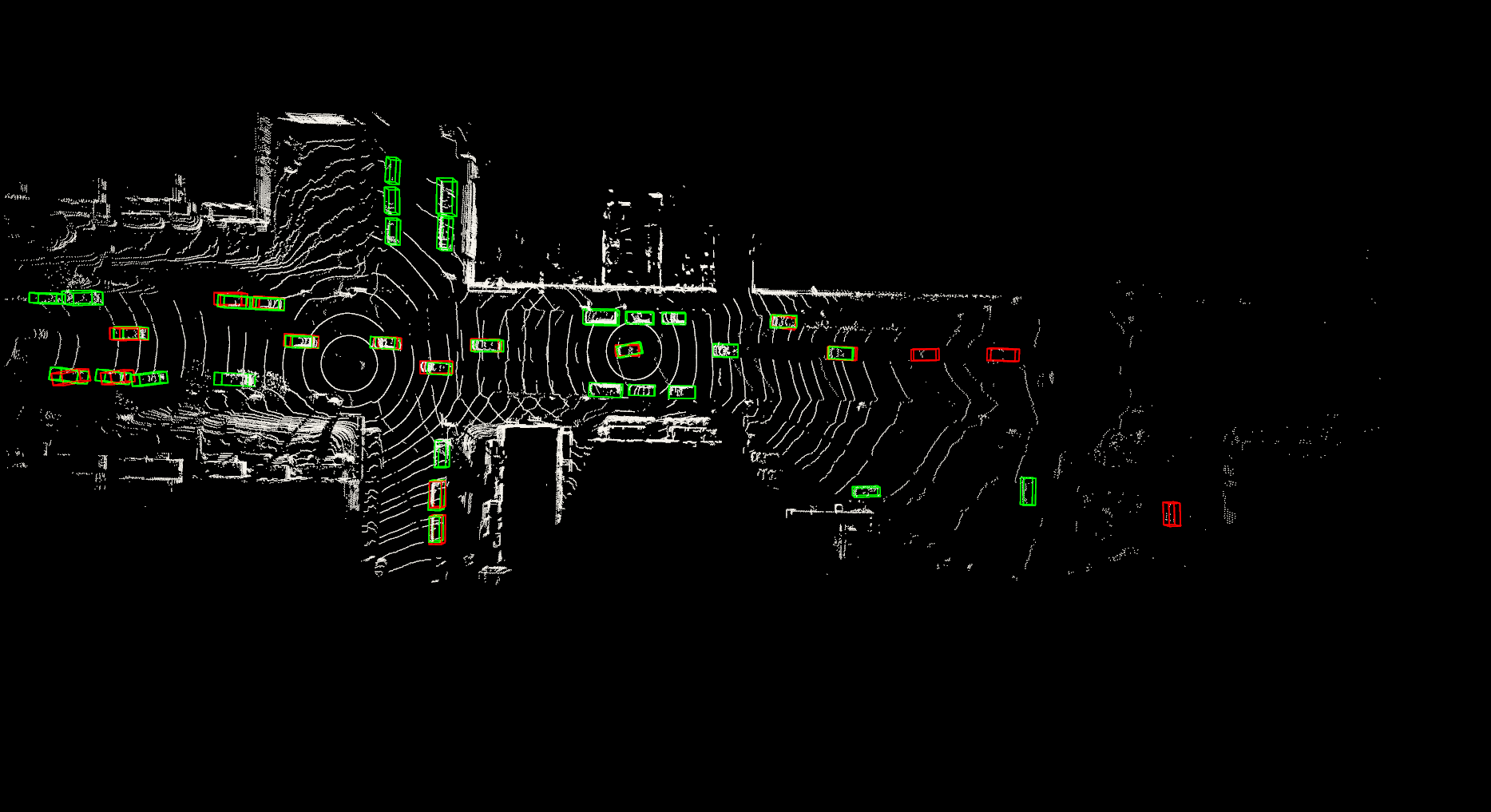}
& \includegraphics[ width=\xwidth\linewidth]{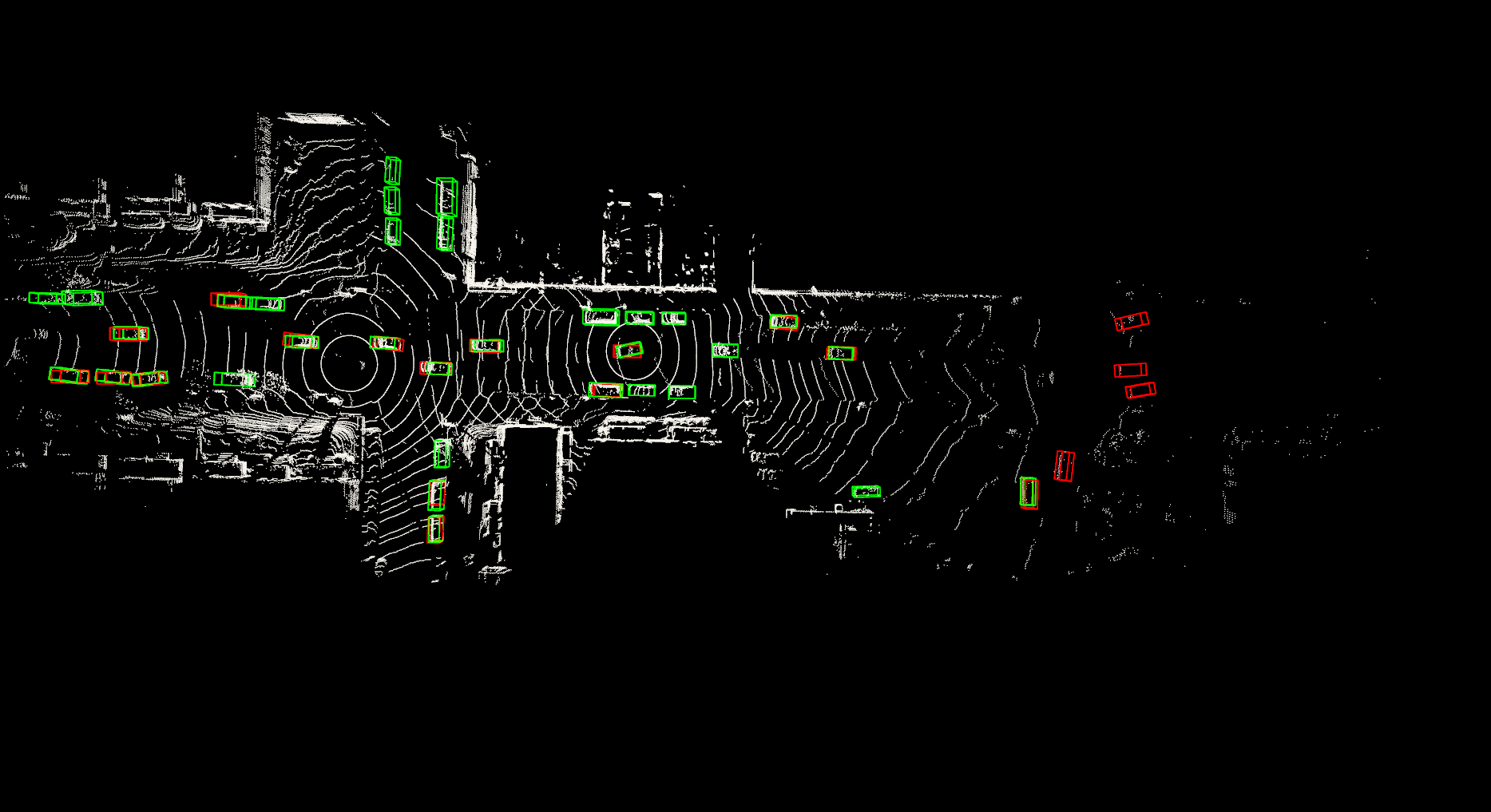}\\

\end{tabular}
\vspace{-3mm}
\caption{\textbf{Qualitative results of domain adaption in an intersection scenario.} \textcolor{green}{Green} and \textcolor{red}{red} 3D bounding boxes represent the groundtruth and prediction, respectively.}
\label{fig:sup-da2}
\end{figure*}

\cref{fig:sup-da1,fig:sup-da2} show the qualitative results of the cooperative domain adaptation. It may be observed in the highway scenario (\cref{fig:sup-da1}) that all the models benefited from applying domain adaptation strategies, with AttFuse and F-Cooper gaining the most, observing from the huge performance difference between the detection results without and without domain adaptation. In a more crowded intersection scene (\cref{fig:sup-da2}), we may see that F-Cooper, V2X-ViT, and CoBEVT are top performers among the compared methods. However, F-Cooper has been observed to produce more false positives while less so for V2X-ViT, after domain adaptation is applied.


\end{document}